# Foundation Models in Medical Image Analysis: A Systematic Review and Quantitative Analysis

Praveenbalaji Rajendran[1], Mojtaba Safari[2], Wenfeng He[3], Mingzhe Hu[3], Keyur Shah[1], Shansong Wang[2], Jun Zhou[1] and Xiaofeng Yang[1,2,3*]

[1]Department of Radiation Oncology and Winship Cancer Institute, Emory University, Atlanta, GA 30322
[2]Department of Radiation and Cellular Oncology, The University of Chicago, IL 60637

[3]Department of Computer Science and Informatics, Emory University, Atlanta, GA 30322

## ABSTRACT

Recent advancements in foundation models (FMs) have catalyzed a paradigm shift in medical image analysis. Unlike traditional task-specific artificial intelligence (AI) models, FMs leverage large-scale datasets to learn generalized representations that can be adapted to downstream clinical applications. Despite the rapid proliferation of FM research in medical imaging, there is a lack of unified synthesis that systematically maps the evolution of architectures, training paradigms, and clinical applications across modalities. To address this gap, this review provides a comprehensive and structured synthesis of FMs in medical image analysis by systematically organizing studies into two primary categories: vision-only foundation models (VFMs) and vision-language foundation models (VLFMs), based on their architectural foundations, training strategies, and downstream clinical tasks. A quantitative analysis was conducted on both VFMs and VLFMs to characterize temporal trends in dataset utilization and application domains, along with pooled performance and subgroup analyses. We also critically discuss persistent challenges, including cross-domain generalization, computational scalability, FM evaluation, fairness, and deployment. Finally, we identify key future research directions aimed at enhancing the robustness, interpretability, and clinical integration of FMs, thereby accelerating their translation into real-world medical practice.

## 1. INTRODUCTION

In the 21st century, medical imaging has emerged as a cornerstone of modern healthcare. Modalities, such as X-ray, computed tomography (CT), magnetic resonance imaging (MRI), ultrasound (US) imaging, and microscopy, have become indispensable tools for the diagnosis and treatment of various human diseases [1]. Traditionally, the interpretation of these medical images has relied heavily on the expertise of medical professionals, such as radiologists, clinicians, and medical physicists. However, as modern medical imaging becomes more complex and multidimensional, manual interpretation has become more time-consuming, labor-intensive, and prone to inter-observer variability, posing challenges to consistency and efficiency in clinical workflows.

Over the past decade, advancements in high-performance computing and data availability have ushered in a new paradigm in artificial intelligence (AI), known as deep learning (DL). A field that has significantly transformed medical image analysis by replacing traditional manual analysis with hierarchical representation-based learning from raw image data. Over the past decade, DL has achieved superior performance across a wide range of medical imaging tasks, including image classification, segmentation, detection, reconstruction, registration, and computer-aided diagnosis (CAD) [2]. However, conventional DL approaches often suffer from significant limitations, such as their dependence on large-scale annotated datasets for training, a process that is time-consuming and labor-intensive. Moreover, these conventional DL models are often constrained to specific tasks, requiring extensive fine-tuning to adapt to new applications, thereby limiting their scalability and generalizability [3-6].

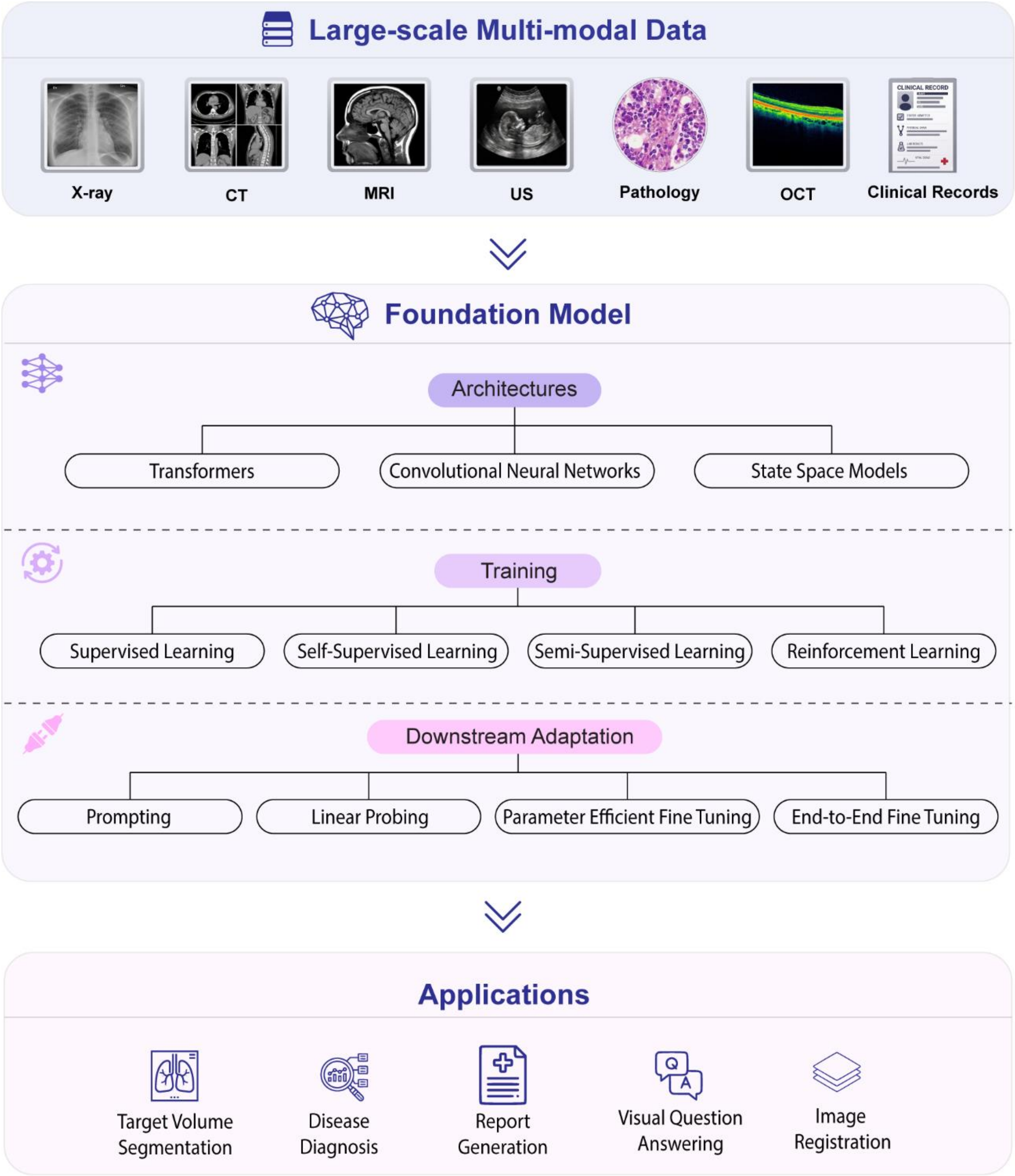

**Fig. 1.** General overview of the FM pipeline in medical image analysis.

In recent years, the AI landscape has been transformed by the development of more generalizable large-scale pretrained frameworks termed foundation models (FMs). Trained on vast and diverse labeled or unlabeled datasets, these FMs are typically characterized by large-scale architectures that can generalize across a wide range of downstream tasks with zero-shot or few-shot adaptation [7]. The conceptual genesis of FMs can be attributed to natural language processing (NLP), where transformer-based large language models (LLMs) such as BERT [8] and GPT [9] have demonstrated unprecedented capabilities in understanding and generating human language. Subsequently, advanced LLMs, such as GPT-5 [10], PaLM [11], Llama [12], and Gemini [13], have further substantiated the potential of LLMs by enabling significant advances in coherent text generation, complex reasoning, multilingual processing, and factual grounding.

Following their success in NLP, FMs have been increasingly adapted to computer vision with the advent of Vision Transformers (ViTs) [14] and subsequently extended into multimodal contexts. Models such as Contrastive Language–Image Pre-training (CLIP) [15], Flamingo [16], and DALL·E [17] epitomize this by utilizing contrastive pretraining [18] to align visual and textual data in a shared embedding space, thereby enabling robust performance across diverse tasks, including cross-modal retrieval, image classification, and visual question answering (VQA). Recognizing this potential,

FMs have been increasingly adapted to medical image analysis, where data is inherently multimodal, originating from diverse sources such as radiological images, pathological slides, clinical notes, and electronic health records. In particular, vision-only foundation models (VFMs) and vision–language foundation models (VLFMs) are well-suited for addressing these complexities. VFMs, which operate exclusively on image data, have demonstrated strong performance over a range of medical imaging tasks, including anatomical segmentation [19, 20], disease classification [21, 22], and lesion detection [23, 24]. Conversely, VLFMs extend the capabilities of VFMs by jointly embedding visual and textual modalities, enabling a comprehensive understanding of the clinical context for complex decision-making. Consequently, VLFMs have demonstrated strong performance in cross-modal applications, including automated report generation [25, 26], multimodal retrieval [27, 28], and VQA [29, 30]. An overview of the general FM workflow, illustrating the pretraining, fine-tuning, and downstream deployment stages, is shown in Fig. 1.

While several recent reviews have explored FMs in medical imaging, they emphasize different aspects of the field. For instance, Azad et al. [31] performed one of the earliest comprehensive surveys categorizing ~40 studies by their training strategies; however, this study preceded the rapid expansion of VLFM research. Jiao et al. [32] conducted a PRISMA-style systematic review of medical FMs, stratified by modality and architecture type, providing broad methodological coverage excluding the quantitative synthesis of performance outcomes. In the vision-only domain, Veldhuizen et al. [33] conducted the most extensive coverage to date by reviewing over 150 VFM studies categorized by their imaging modality, excluding VLFMs. Conversely, Liu et al. [34] and Ryu et al. [35] focused primarily on VLFMs and multimodal FMs, excluding VFMs. Similarly, Sun et al. [36] provided a systematic review and meta-analysis with quantitative performance synthesis limited to VLFMs, lacking a corresponding VFM analysis or cross-paradigm comparison. Niu et al. [37] synthesized the medical imaging FM landscape offering extensive coverage of the architectures, training workflows and efficiency techniques. However, they adopted a qualitative approach without a structured VFM-VLFM taxonomy and quantitative analysis.

Despite these extensive contributions, a comprehensive synthesis encompassing both VFM and VLFM paradigms is still lacking. In particular, an integrated analysis is essential for understanding their complementary roles and emerging areas of convergence. Examining both paradigms within a common framework enables cross-paradigm comparison of their training strategies, downstream tasks, and sources of performance heterogeneity, thereby revealing shared methodological trends and paradigm-specific strengths and limitations, supporting a task-driven approach to model selection. Thus, to address this gap, this review integrates a structured taxonomy with quantitative synthesis across both paradigms. Within the VFM paradigm, we present a detailed taxonomy organized into Segment Anything Model (SAM)-based adaptations and generalist VFMs, encompassing both task-agnostic and task-specific frameworks. Within the VLFMs, we systematically categorize the studies into contrastive learning-based and generative frameworks, with the latter including both generalist and domain-specific generative models. Furthermore, we conducted a quantitative analysis across both paradigms encompassing pooled performance synthesis and subgroup analyses, along with an examination of temporal and modality-specific trends in dataset utilization and downstream tasks. Finally, we discuss the real-world clinical impact of FMs, outline the existing limitations, and propose future research directions. This review is intended for the biomedical imaging community, with a focus on methodological trends along the imaging pipeline and pragmatic deployment considerations, rather than exhaustive clinical validation.

## 2. LITERATURE SEARCH AND TAXONOMY

To capture the extent of FM research in medical image analysis, we performed systematic searches across multiple databases, including PubMed, IEEE, ArXiv, and Google Scholar. The search queries incorporated key terms such as "foundation model", "vision foundation model", "visual foundation model", "vision-language model", "visual-language model", "medical foundation model", "self-supervised", "contrastive learning", "masked autoencoder", "CLIP", "SAM", and "Segment Anything" along with medical imaging domains and modalities such as "radiology", "CT", "MRI", "ultrasound", "pathology", "ophthalmology". The literature search covered English-language records published from 01 January 2022 to 28 February 2026. We followed a PRISMA-style screening process, as illustrated in Fig. 2. The initial database search yielded PubMed (n = 401), IEEE Xplore (n = 315), ArXiv (n = 558), and Google Scholar (n = 176), yielding 1450 records, from which 312 duplicates were removed. The remaining 1,138 records underwent title and abstract screening by two independent reviewers, resulting in the exclusion of 724 records. Records were excluded if they (i) did not involve medical image analysis, (ii) did not utilize a FM, (iii) did not report evaluations of downstream medical imaging tasks, (iv) focused solely on textual or clinical report analysis, or (v) were non original article types. Disagreements between reviewers were resolved through discussion and consensus. The remaining 414 full-text articles were retrieved and assessed for eligibility. During the full-text review, 177 articles were excluded based on the following criteria: not foundation models, review articles, studies outside the scope of medical imaging, full-text unavailable, or overlapping ArXiv preprints. This resulted in 237 studies (VFMs = 137, VLFMs = 100) included in the qualitative synthesis, of which 224 were included in the final quantitative analysis (VFMs = 135, VLFMs = 89).

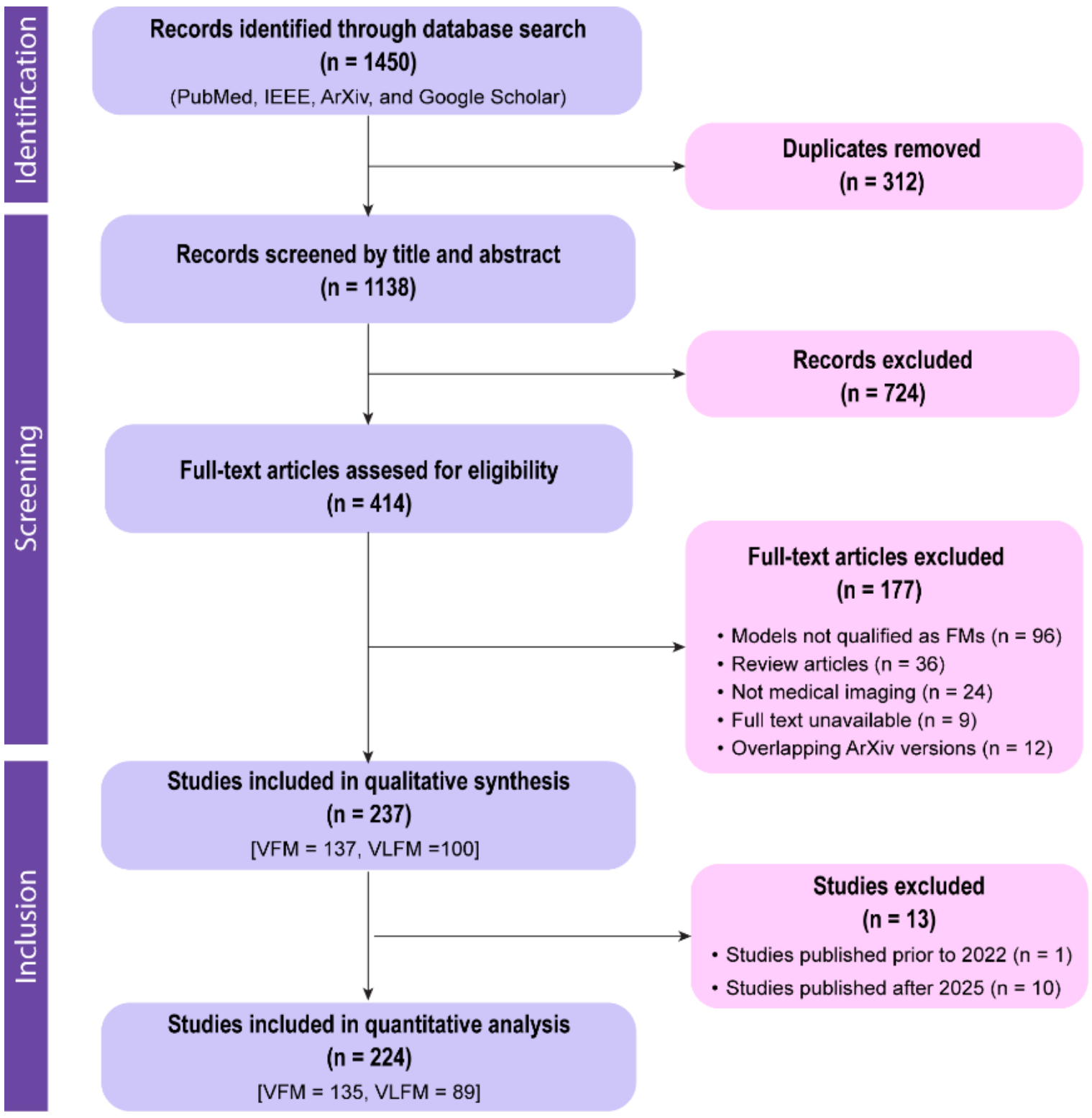

**Fig. 2.** Flow diagram illustrating the PRISMA-style literature selection process.

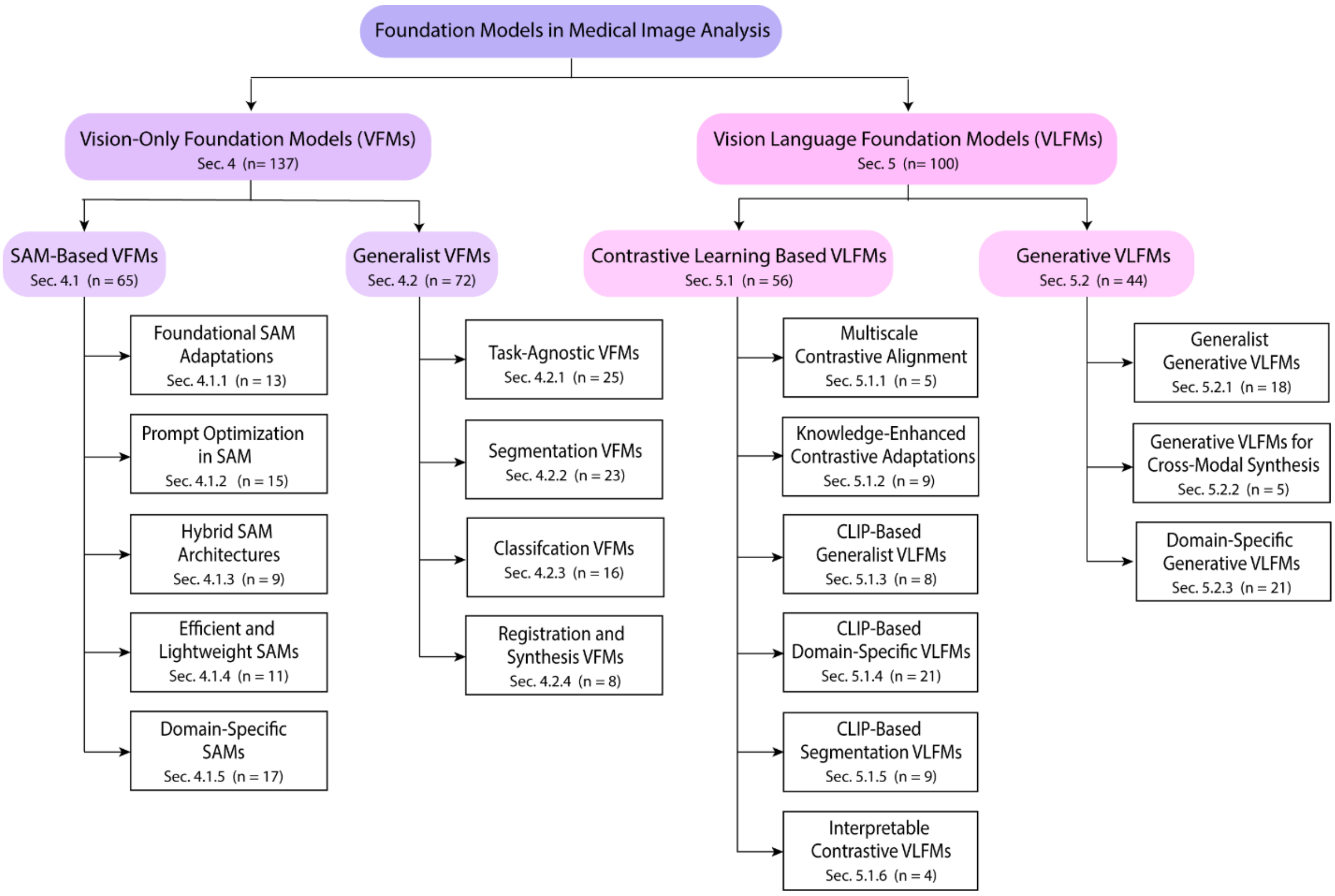

**Fig. 3.** Taxonomy of foundation models in medical image analysis.

To provide a structured synthesis of the survey, we present a hierarchical taxonomy of FMs in medical image analysis, as illustrated in Fig. 3. The taxonomy organizes the 237 studies into two primary categories: VFMs and VLFMs. Within VFMs, 137 studies are further organized based on their architectural origins into SAM-based adaptations spanning foundational adaptations, prompt optimization, hybrid architectures, and lightweight variants and generalist VFMs covering task-agnostic, segmentation, classification, and registration frameworks. Within VLFMs, 100 studies are categorized based on their pretraining paradigm into contrastive learning-based frameworks comprising multiscale contrastive alignment, knowledge-enhanced adaptations, CLIP-based generalist, domain-specific, and segmentation models and generative frameworks, including generalist generative, cross-modal synthesis, and domain-specific generative models. This taxonomy serves as an organizational guide for the subsequent sections of this review.

# 3. FOUNDATIONS AND TERMINOLOGY

## 3.1 Architectural Backbones of FMs

In FMs, the architectural backbone plays a pivotal role in extracting high-level semantic representations from data and converting them into generalizable embeddings utilized for a wide range of downstream tasks. Over the years, these backbones have evolved from convolution-based to more flexible and scalable transformer- and self-attention-based approaches. Thus, the architectural paradigms of FMs can be broadly classified into three main categories: convolutional neural networks (CNNs), transformer-based architectures, and state space models. However, these architectures are not mutually exclusive, and many applications often adopt hybrid architectures by combining these paradigms to leverage their strengths in handling diverse modalities. This section provides an overview of these foundational architectural classes and discusses key innovations, including encoder–decoder frameworks, hierarchical processing strategies, and promptable segmentation models.

### 3.1.1 Convolutional Neural Networks

Over the last decade, CNNs have been the most widely used architecture in medical image analysis. Typically, CNNs leverage convolutional layers along with pooling and activation layers to learn hierarchical features from images. In particular, convolutional layers employ learnable kernels to generate feature maps, whereas pooling layers reduce spatial dimensions, and activation layers introduce nonlinearity, thereby enabling the network to learn complex representations. The evolution of CNN architectures toward deeper networks began with the VGGNet [38] demonstrating significant performance improvements by increasing the network depth through the stacking of convolutional layers. However, as network depth increased, CNN began to suffer from the vanishing gradient problem. ResNet [39] addressed this issue by introducing residual connections, which enabled effective training of very deep networks. Similarly, DenseNet [40] promoted feature reuse by densely connecting each layer to all its preceding layers, enhancing gradient flow and reducing redundancy.

Following these advancements, task-specific architectures have been developed to address the unique challenges of medical image analysis. A foundational example is the U-Net architecture [41], which introduces an encoder-decoder structure with skip connections to preserve high-resolution spatial features during image segmentation. As one of the most widely used architectures in medical imaging, U-Net employs an encoder to capture semantic features through progressive downsampling and learns coarse-grained representations. The decoder then upsamples these features to recover spatial resolution and enable precise localization. Skip connections act as a bridge between the encoder and decoder to enhance segmentation accuracy by preserving spatial information. V-Net [42] adapted the U-Net architecture for 3D volumetric information by replacing the 2D convolutions with 3D convolutions. Another notable adaptation is the feature pyramid network [43], which enhances standard CNN by introducing a top-down pathway and lateral connections to fuse multiscale feature maps, exemplifying robust performance across detection and segmentation tasks. Despite these advances, CNNs are inherently limited in modeling long-range dependencies due to the localized nature of convolutional filters. Furthermore, their scalability to large and heterogeneous datasets remains limited.

### 3.1.2 Transformers

Originally introduced for NLP, transformers have emerged as the intrinsic component of modern FMs due to their ability to capture long-range dependencies through self-attention mechanisms [44]. Unlike recurrent architectures, transformers rely on self-attention to model global interactions in sequential data, enabling superior scalability across heterogeneous datasets. Typically, a transformer layer comprising a multi-head self-attention block followed by a feed-forward neural network is utilized to learn diverse feature representations by combining token-to-token interactions with nonlinear position-wise transformations. This architectural design endows transformers with exceptional representational power across a variety of sequential modeling tasks.

Extending the transformer architecture to the visual domain, ViTs signify a paradigm shift in computer vision by treating images as sequences of fixed-size, non-overlapping patches, analogous to word tokens [14]. In ViTs, each image patch is linearly embedded and augmented with positional encodings before being processed through successive transformer layers. This patch-based representation enables ViTs to leverage self-attention mechanisms to model both global and local contextual relationships across the entire image, in contrast to CNN operating with a limited receptive field. Moreover, the use of dynamic data-driven attention weights instead of static convolutional kernels allows ViTs to flexibly adjust their receptive fields based on the input data, making them adept at processing high-resolution and complex visual information. Over the years, ViTs and their hierarchical variants, such as Swin transformers [45] and UNETR [46], have demonstrated superior performance across a wide range of medical imaging tasks, including anatomical segmentation, disease classification, and lesion detection.

The scalability of ViTs has catalyzed the development of several promptable VFMs. A notable example is SAM [47], which utilizes diverse prompts, such as bounding boxes or points, to generate zero-shot segmentation maps without retraining. ViTs are also the foundation for universal generalist models, such as TotalSegmentator [19] and SegVol [48], which aim to segment multiple anatomical structures or pathologies from diverse imaging modalities using a single pretrained backbone. ViTs have also been integrated with CNNs to develop powerful hybrid architectures that combine efficient local feature extraction with global context modeling. An example is MedFormer [49], which combines a CNN backbone with ViTs for localized attention, achieving improved accuracy and efficiency across complex medical imaging tasks.

#### 3.1.3 State Space Models

State space models (SSMs) represent a distinct class of computationally efficient architectures for capturing long-range dependencies in sequential data. Rooted in classical control theory, SSMs model the dynamics of a system through linear state transitions, enabling sequential processing without the memory overhead of self-attention. Although primarily implemented for NLP, Mamba [50] enables SSM adaptation to vision applications through its selective scan mechanism, wherein state transitions adapt to specific inputs, allowing the model to selectively retain pertinent contextual features while suppressing irrelevant information along the sequence. Vision Mamba (Vim) [51] exemplifies this adaptation to visual data by extending the Mamba architecture to 2D images through bidirectional scanning of image patch sequences with positional embeddings, thereby significantly reducing the computational overhead of ViTs. VMamba [52] further enhances this spatial modeling by introducing a cross-scan module to traverse 2D feature maps along four spatial directions, providing direction-aware global context at linear complexity. Building on these advancements, OpenUS [53] and 3D-SCUMamba [54] pioneered the adaptation of SSMs in medical imaging by leveraging Vim backbones for classification and segmentation. However, the adaptations of SSM as FM architectures are still nascent relative to transformers and CNN.

### 3.2 Learning Paradigms in FM Training

In contrast to conventional DL approaches that rely on task-specific labeled datasets for training, FMs are pretrained on large corpora of labeled or unlabeled data to learn generalizable representations. [7]. This is typically achieved through a variety of learning paradigms, including supervised learning (SL), semi-supervised learning (semi-SL), self-supervised learning (SSL), and reinforcement learning (RL) [7, 55]. This section summarizes these paradigms and highlights their applications in training VFMs and VLFMs for medical image processing.

#### 3.2.1 Supervised Learning

SL is a classical learning paradigm where a DL model is optimized utilizing pairs of input and ground truth labels to learn the underlying relationships between the input attributes and target predictions [56]. In SL, the model parameters are iteratively updated to minimize task-dependent objectives via gradient-based optimization. Typically, in medical imaging, SL is predominantly used for classification, detection, and segmentation tasks, and it has demonstrated expert-level performance when high-quality annotations are available [34, 57]. However, optimizing FMs via SL is often constrained by the scale and fidelity of the labeled data, as expert annotations are expensive, time-consuming, and prone to inter-observer variability, acquisition differences, and diagnostic errors. Consequently, many FM pipelines rely on SSL or semi-SL to learn transferable representations and utilize SL primarily for downstream fine-tuning or training task-specific heads. However, in scenarios where large-scale annotated datasets exist, SL can lead to foundational capabilities [58, 59].

#### 3.2.2 Self-Supervised Learning

SSL is a learning paradigm that enables FMs to learn from large-scale unlabeled data by generating supervisory signals (pseudo-labels) from the data itself. In particular, SSL employs pretext tasks to train models by extracting meaningful features from raw inputs [55]. Generally, the SSL paradigm can be categorized into predictive, generative, and discriminative techniques [55, 60].

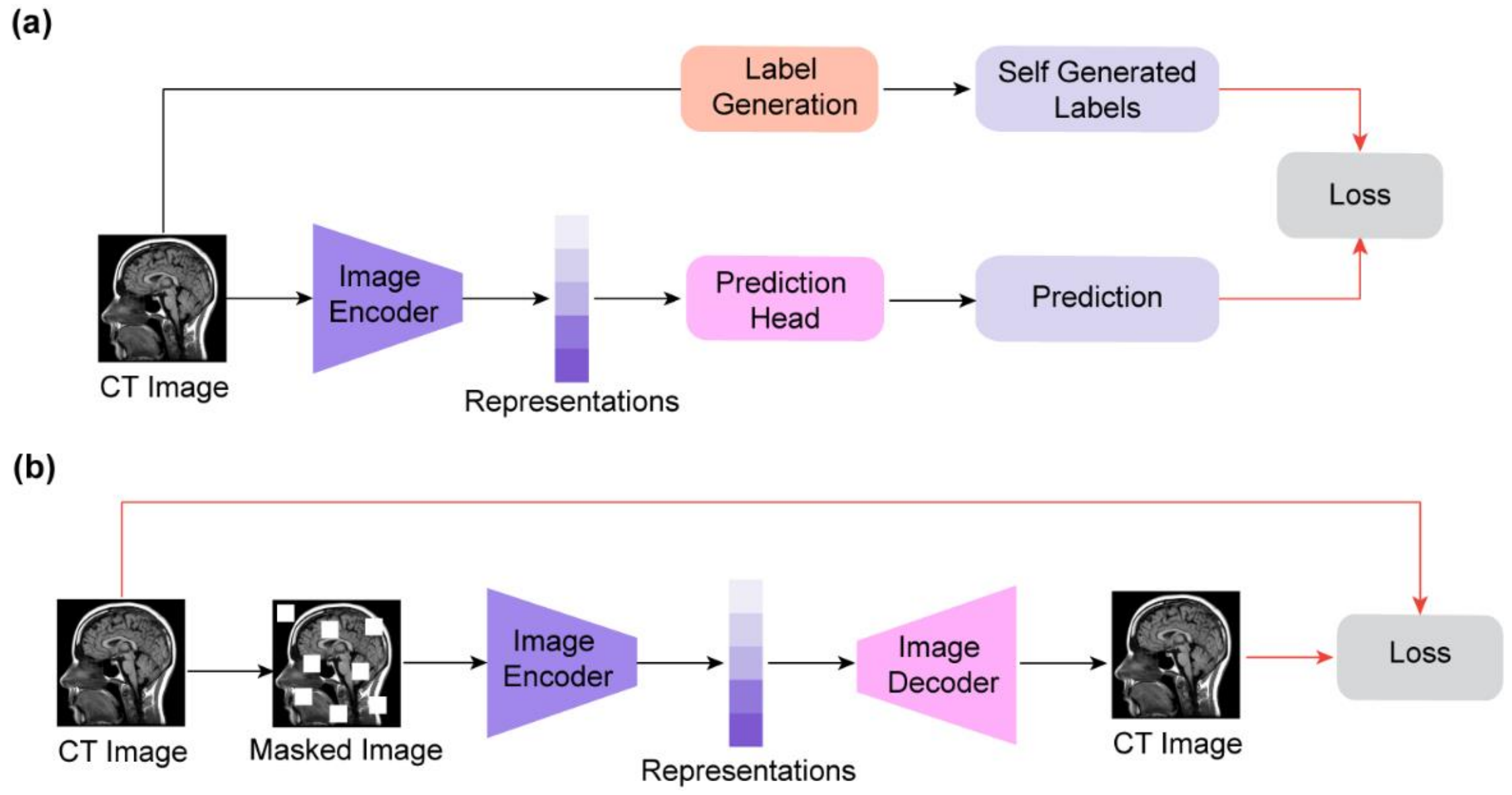

**Fig. 4.** Schematic representation of (a) predictive and (b) generative SSL paradigms in medical imaging.

**Predictive Learning:**

Predictive SSL involves the generation of supervisory signals through pretext tasks that require the model to predict masked or transformed aspects of data. As illustrated in Fig. 4(a), these predictive paradigms enable the network to learn semantic and spatial relationships by reconstructing or identifying transformed regions. For instance, Doersch et al. [61] proposed a relative patch prediction task, where an image is divided into patches, and the model is trained to predict the relative position of one patch with respect to another by encouraging the model to learn spatial relationships within the image. Similarly, Noroozi et al. [62] introduced a jigsaw puzzle task, where image patches are shuffled, and the model learns to predict the correct permutation, capturing both semantic and spatial dependencies. Gidaris et al. [63]. proposed a rotation prediction task, wherein a CNN learns to classify the transformations applied to an image (0°, 90°, 180°, or 270°) for efficiently learning the semantic features. However, in medical imaging, absolute orientation is not a reliable semantic cue owing to variability in acquisition protocols and patient positioning. Thus, rotation-based pretext tasks may be less reliable and can encourage learning acquisition and interpolation-induced cues rather than anatomy-driven representations.

Beyond pixel-level prediction, the joint embedding predictive architecture (JEPA) [64] extends the predictive paradigm by reformulating SSL to predict abstract semantic features in the latent space rather than reconstructing raw pixel values. By predicting these feature representations, JEPA aligns with the concept of a world model by encouraging the model to capture high-level semantic features while filtering out high-frequency noise and modality-specific artifacts [64]. A prominent instantiation of JEPA is image-based JEPA (I-JEPA) [65], which masks large contiguous image regions and trains a predictor network to forecast the target representations without relying on handcrafted data augmentations or negative pairs. Similarly, data2vec [66] and VL-JEPA [67] extend JEPA across modalities. Specifically, VL-JEPA predicts continuous semantic embeddings of target texts conditioned on visual inputs rather than autoregressively generating discrete tokens to capture task-specific features and discard task-irrelevant surface-level linguistic variability. In contrast, data2vec predicts latent representations of speech, vision, and language inputs using teacher-student distillation, demonstrating that latent-space prediction can serve as a unifying objective across modalities. In medical imaging, JEPA-based approaches hold great potential for modalities where augmentation-free training is desirable to preserve clinically meaningful features.

**Generative Learning:**

Generative SSL utilizes the input data itself as supervisory signals to capture the underlying probability distribution of data [68]. A classic example of a generative task is the autoencoder, where an encoder transforms high-dimensional input data into a lower-dimensional latent representation, and the decoder reconstructs it from this representation, thereby encouraging the encoder to learn compact, informative representations during training[68].

Inspired by masked language modeling (MLM), masked image modeling (MIM) has emerged as a dominant generative paradigm in computer vision [Fig. 4(b)]. In MIM, parts of the input are masked, and the model is trained to reconstruct the missing information. A prominent example is the Masked Autoencoder (MAE) [69], which employ an encoder–decoder architecture to reconstruct randomly masked image regions. Similarly, Pathak et al. [70] proposed context encoders which learns image inpainting by predicting missing regions from the surrounding context. An extension of MAE in the medical imaging domain is VIS-MAE [71], which has demonstrated strong performance in classification and segmentation tasks on volumetric medical data.

In VLFMs, MIM extends to the paradigm of masked representation modeling (MRM), where masked image and text tokens are jointly utilized to train on multimodal data [72]. For instance, RadFM [73] applies generative modeling to produce radiology reports conditioned on imaging inputs, such as X-ray or MRI. In addition to masked modeling, diffusion models have gained prominence as powerful generative frameworks in medical imaging. In diffusion modeling, noise is gradually added to the data, and the model learns to reverse this process, enabling the generation of high-fidelity medical images. For example, MedDiff-FM [74] utilizes diffusion-based modeling for a diverse range of downstream tasks, such as denoising, anomaly detection, and image synthesis, without minimal or no task-specific fine-tuning.

**Discriminative Learning:**

In discriminative learning, contrastive SSL represents a prominent subclass, where models are trained to map positive pairs closer in the embedding space while pushing negative pairs apart. As illustrated in Fig. 5 (a), this process encourages the model to learn invariant and discriminative representations by contrasting similar and dissimilar samples. For example. Chen et al [18] proposed SimCLR, a contrastive learning framework, where positive pairs are generated by applying different stochastic augmentations to the same image, while other images acts as negatives. SimCLR demonstrated that the use of a learnable nonlinear projection head with strong data augmentations has significantly improved the representation quality. However, a primary limitation is the requirement of very large batch sizes to ensure sufficient negative samples, which makes training computationally expensive and memory intensive.

To overcome this limitation, He et al. [75] introduced MoCo, which uses a dynamic queue to store negative samples across training batches to reduce memory requirements while maintaining performance. Similarly, Caron et al. [76] proposed SwAV, which avoids explicit pairwise comparisons and large memory requirements by leveraging a clustering-based objective and a swapped prediction mechanism integrated with a multicrop augmentation strategy. Recent advancements further reduce the dependence on negative pairs while learning view-invariant discriminative representations using techniques such as stop-gradient and self-distillation. For instance, DINO [77] is a negative-free self-distillation approach where a student learns from a momentum-updated teacher network, yielding semantically rich features without labels. DINOv2 [78] further refines DINO by incorporating robust data augmentations and large-scale pretraining to enhance cross-domain generalization. DINOv3 [79] extends DINO and DINOv2 by integrating gram anchoring to stabilize and prevent the degradation of dense feature maps during pretraining, enhancing generalization across diverse tasks without fine-tuning.

Among VLFMs, contrastive learning plays an important role in aligning image and text embedding into a shared latent space [Fig. 5 (b)]. A notable example is CLIP by Radford et al. [15], where a vision encoder and a text encoder are jointly trained with contrastive loss to maximize the similarity of paired image-text representations. In medical imaging, frameworks such as MedCLIP, PMC-CLIP, CXR-CLIP, and BioViL [80-83] have adapted the CLIP architecture by pretraining on large-scale medical image-text datasets, enabling zero-shot classification, cross-modal retrieval, and text-guided diagnosis.

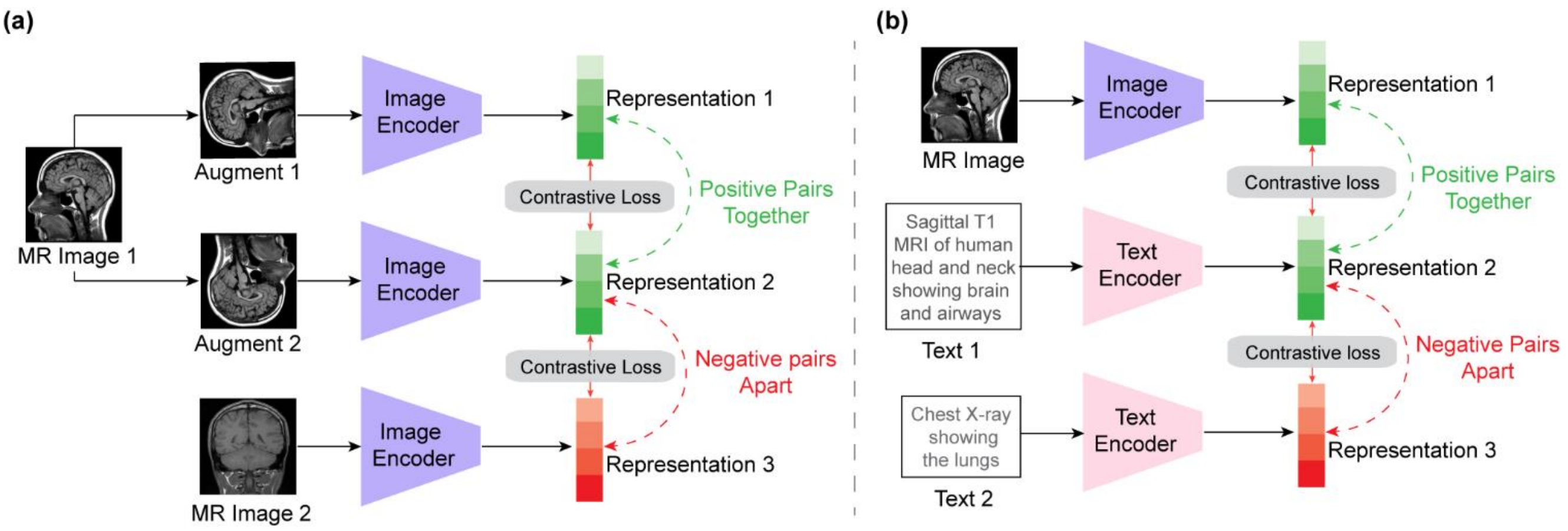

**Fig. 5.** Schematic representation of (a) image–image contrastive SSL framework and (b) vision–language contrastive SSL framework.

### 3.2.3 Semi-Supervised Learning

Semi-SL bridges the gap between supervised and unsupervised learning by utilizing a small set of labeled data alongside large corpora of unlabeled data for training FMs. In semi-SL, a widely adopted strategy includes consistency regularization and pseudo-labeling. For instance, FixMatch [84] integrates confidence-thresholded pseudo-labeling with consistency regularization to generate pseudo-labels from weakly augmented unlabeled data while enforcing prediction consistency under strong augmentations. Similarly, EMA [85] employs a teacher–student framework, where the teacher network is updated as an exponential moving average of the student to produce pseudo-labels for guiding unlabeled data training. Extensions of these frameworks often adopt knowledge distillation to transfer supervisory signals from teachers to student to improve representation learning in low-label settings[86].

In recent years, semi-SL approaches have been integrated with contrastive and distillation-based objectives in VITs, enabling more robust and discriminative representation learning from limited annotations [87]. By combining pseudo-labeling, consistency regularization, and contrastive objectives, these frameworks improve inter-class separability in the embedding space and enhance generalization across heterogeneous datasets. In medical imaging, semi-SL is more suitable for segmentation and classification tasks, where annotated data is scarce but large volumes of unlabeled CT, MRI, or X-ray scans are readily available.

### 3.2.4 Reinforcement Learning

RL represents a distinct learning paradigm, where an FM is trained through interactions guided by scalar reward signals rather than explicit input-output pairs [88]. In VLFMs, given the medical image and context, the actions of the FMs are evaluated to generate the reward signal. The model parameters are then updated to maximize this cumulative reward, enabling optimization of complex non-quantifiable objectives, such as clinical accuracy, interpretability, and factual consistency. Unlike conventional SL and SSL paradigms, RL enables optimization utilizing a human-in-the-loop feedback mechanism. Notably, reinforcement learning from human feedback (RLHF) [88] adopts the RL framework comprising three stages: supervised fine-tuning on task-specific data, training a reward model based on human rankings of model outputs, and FM policy optimization using proximal policy optimization [89] [Fig. 6] with a Kullback–Leibler divergence penalty to prevent deviation from the reference policy. To mitigate the computational overhead, group relative policy optimization (GRPO) [90] eliminates the need to train separate rewards model by computing the relative advantages among multiple sampled outputs for each prompt. Furthermore, RL has been widely adopted to instill structured reasoning capabilities in VLFMs to overcome hallucinations in diagnostic outputs [91] [92]. Consequently, several VLFMs including MINIM [91], EditGRPO [92], ChestX-Reasoner [93], PathReasoner-R1 [94], and Patho-AgenticRAG [95], have leveraged RL for tasks ranging from clinical image synthesis and report generation to structured diagnostic reasoning.

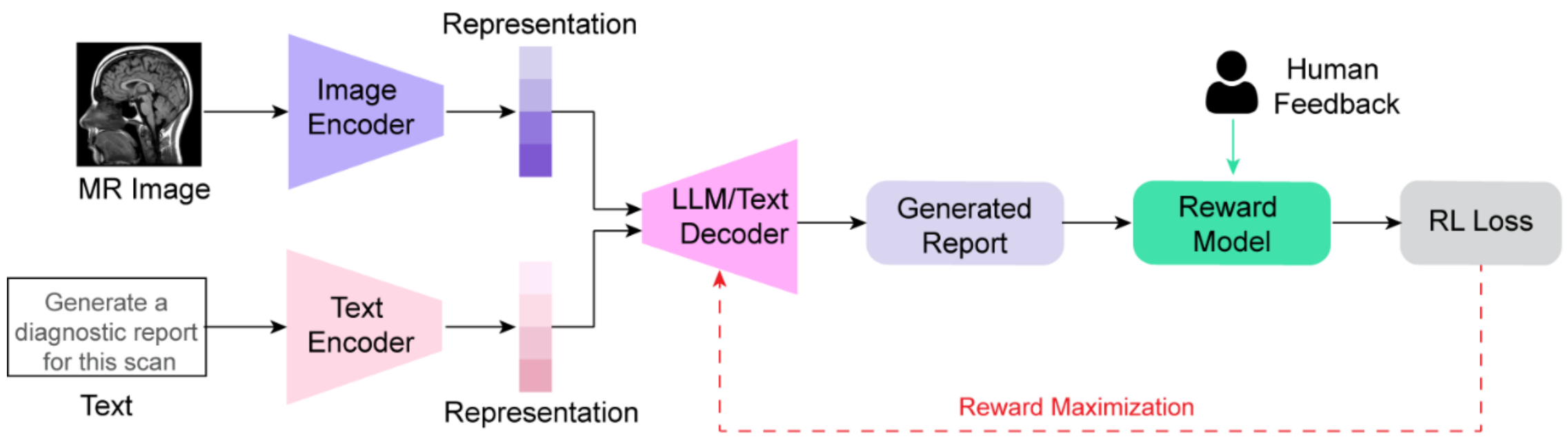

**Fig. 6.** Schematic of RLHF for medical report generation

### 3.3 Downstream Adaptation

Although pretrained FM backbones encode transferable representations for diverse downstream tasks, achieving robust task-specific performance typically requires task- and domain-specific adaptation. In FMs, these adaptation strategies can be categorized into prompt-based adaptation, lightweight supervised adaptation, parameter-efficient fine-tuning (PEFT), and full end-to-end fine-tuning.

#### 3.3.1 Prompt-Based Adaptation

Prompting is a lightweight adaptation technique that steers FM behavior using carefully designed inputs without updating the model parameters. In practice, prompting can be broadly classified into zero-shot and few-shot prompting. In zero-shot prompting, FMs are provided with task-specific instructions to generate the output. In few-shot prompting, task-specific instructions are accompanied by examples to elicit the desired output. A notable implementation of few-shot prompting is in-context learning (ICL), where demonstrations in the prompt serve as contextual conditioning signals to guide the model toward the desired response without any parameter updates. While prompting in VLFMs typically involves textual queries for multimodal reasoning, spatial prompts such as points, bounding boxes and masks are also utilized in promptable VFMs to specify target regions for high-fidelity segmentation [96-98].

#### 3.3.2 Lightweight Supervised Adaptation

In lightweight supervised adaptation, a pretrained frozen FM backbone is adapted to downstream tasks by training a task-specific predictor, such as linear probes or small task-specific heads (e.g., MLPs or CNNs), on labeled data. In particular, linear probes are utilized for classification and regression tasks where pretrained embeddings are linearly separable. On the other hand, task-specific heads are employed for detection and segmentation, where structured outputs require more expressive decoders. Additionally, generative heads are used when downstream objectives involve synthesis, such as report generation or image synthesis. Although adapting lightweight heads minimizes computational costs, they are often insufficient for complex tasks that require deep, domain-specific feature reorganization [98].

#### 3.3.3 Parameter Efficient Fine-Tuning

PEFT adapts FMs to downstream tasks by fine-tuning only a subset of model parameters, thereby reducing computational and memory requirements [99, 100]. In PEFT, one of the most widely used techniques is low-rank adaptation (LoRA), where trainable low-rank decomposition matrices are injected into pretrained layers, while freezing the original pretrained weights [101, 102]. Adapter tuning is another popular PEFT approach, where pretrained adapter modules are inserted into the pretrained FM backbone to facilitate efficient task specialization and multitask deployment by swapping adapter sets [100]. PEFT has also been extended to prompt tuning, where a small set of soft prompt embeddings is optimized to condition the FM for the target task [100].

#### 3.3.4 Full End-to-End Fine-Tuning

In full end-to-end fine-tuning, all FM parameters are optimized for a specific downstream task using labeled data. Although this approach yields the highest performance, it requires substantial computational resources and labeled data to mitigate overfitting. Beyond standard supervised fine-tuning, specialized fine-tuning variants have also been proposed, including instruction tuning, visual instruction tuning, and preference-based alignment. In instruction tuning, training is conditioned on demonstration-style prompts to improve task-specific performance [103, 104]. Visual instruction tuning extends instruction tuning to multimodal context by jointly conditioning on both visual and text pairs [105]. Preference-based alignment further refines model behavior using human feedback, most notably through RLHF [88], which optimizes FMs against a learned preference signal, as discussed in Section 3.2.4.

# 4. VISION-ONLY FOUNDATION MODELS

This section systematically reviews VFMs in medical imaging by categorizing them into two primary groups: SAM-based VFMs and generalist VFMs, where SAM-based VFMs encompass models that adapt the SAM [27] framework for medical imaging through architectural modifications, prompt optimization, and domain-specific fine-tuning. In contrast, generalist VFMs encompass architecturally diverse models stratified by their downstream tasks into task-agnostic, segmentation, classification, registration, and synthesis frameworks. A comprehensive summary of all VFM studies included in qualitative synthesis is provided in Supplementary Table 1. Following the qualitative synthesis, a quantitative analysis of the VFMs is then conducted to evaluate the overall performance and dataset utilization trends.

## 4.1 SAM-Based VFMs

Introduced by the Segment Anything (SA) project in 2023, SAM [27] marked a new era of universal segmentation in VFMs by reframing segmentation as a prompt-conditioned prediction task. SAM achieves this flexibility through a modular architecture comprising a ViT based image encoder that converts the input image into a high-dimensional embedding, a prompt encoder that maps user inputs (points, bounding boxes, and masks) into conditional embeddings, and a lightweight mask decoder that fuses image and prompt embeddings using a two-way transformer to predict segmentation masks. During training, SAM simulates interactive segmentation by sampling prompts over multiple rounds per ground-truth mask and optimizing mask prediction with standard segmentation losses. To address the scarcity of large-scale annotations, the SA project introduced a data engine that couples model training with annotation, progressing from assisted-manual labeling to fully automatic mask generation. This process culminated in the SA-1B dataset, containing more than 1.1B masks from 11M diverse images. Trained on the SA-1B dataset, SAM enables zero-shot segmentation without task-specific fine-tuning and demonstrates strong generalization across a variety of segmentation tasks. Recognizing this potential, SAM has been increasingly adapted for medical imaging applications; the following sections provide a categorized overview of these adaptations, highlighting key extensions and domain-specific adaptations.

### 4.1.1 Foundational SAM Adaptations

The introduction of SAM as a universal segmenter in computer vision has garnered significant attention in medical image analysis over recent years. Early studies investigated the potential of SAM's zero-shot capabilities for medical image segmentation without retraining [106, 107]. However, despite its superior performance in the natural imaging domain, the direct adaptation of SAM to medical imaging is limited by a substantial domain gap arising from differences in texture, contrast, modality-specific features, and complex anatomical structures compared to natural images [108]. To address this limitation, numerous foundational adaptations of SAM have been proposed to bridge this gap by fine-tuning the original SAM on large-scale medical datasets.

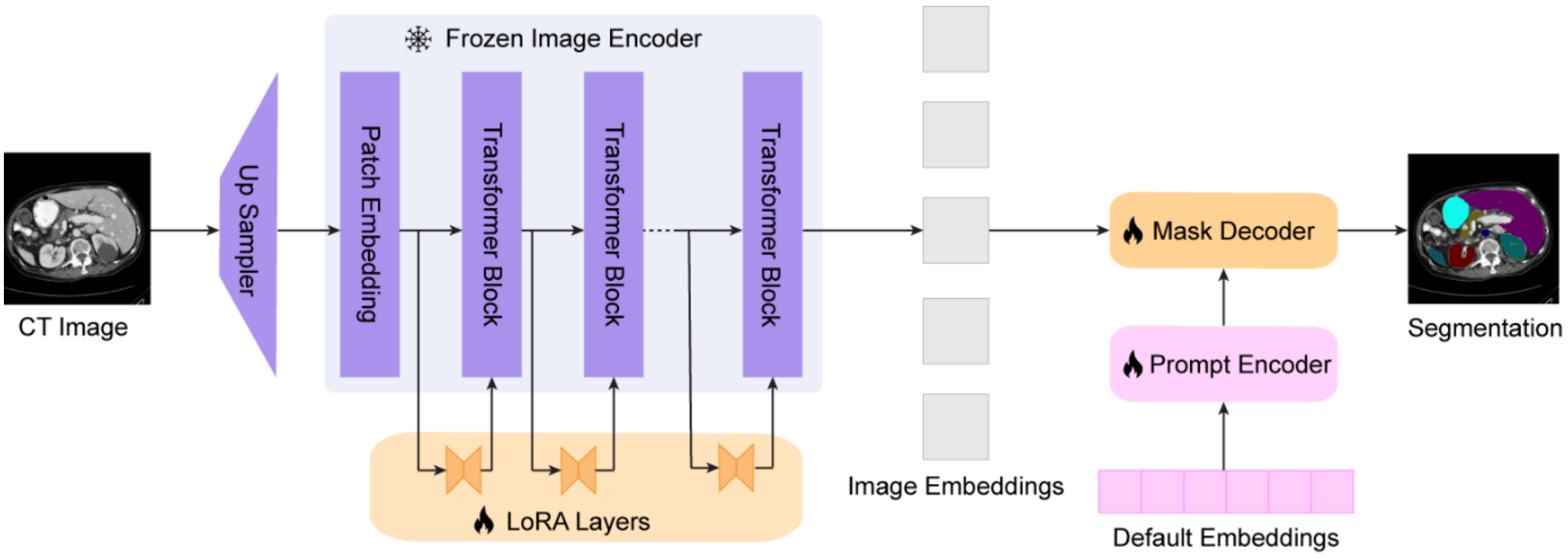

**Fig. 7.** Overview of SAMed [109] architecture

MedSAM, proposed by Ma et al.[59], is one of the earliest foundational adaptations that fine-tunes the original SAM architecture on a large-scale curated medical image dataset comprising over 1M image–mask pairs across 10 diverse imaging modalities, including CT, MRI, and US. Similarly, SAM-Med2D [110] extended this approach by fine-tuning SAM on a dataset of ~4.6M 2D medical images, enabling the model to adapt to diverse prompts, such as points, bounding boxes, and masks, outperforming SAM. Conversely, PEFT strategies have been adapted to SAM to avoid full retraining and reduce computational requirements. SAMed [109] exemplifies this by integrating LoRA layers into the SAM image encoder, as illustrated in Fig. 7, achieving segmentation accuracy on par with SOTA baselines by tuning only a fraction of SAM's parameters. Med-SA [111] injects domain-specific knowledge into SAM by employing two key adaptation techniques: space–depth transpose to process 3D volumetric medical images and hyper-prompting adapter to support prompt-conditioned adaptation for more flexible inference. Complementarily, MoPEFT [112] adopts a modular PEFT strategy with trainable gates to adaptively select the most suitable fine-tuning strategy based on the input data, facilitating efficient adaptation across medical imaging domains without full model retraining.

Although these early 2D SAM adaptations improved slice-wise segmentation performance, medical imaging modalities are inherently volumetric, and the naive extension of 2D SAM to 3D data often fails to capture long-range inter-slice dependencies, resulting in inconsistent volumetric segmentations. To address this limitation, SAM-Med3D [113] redesigns the SAM pipeline with fully learnable 3D components, including a 3D image encoder and mask decoder trained on the SA-Med3D-140K dataset, achieving competitive Dice scores on benchmarks such as the Beyond the Cranial Vault (BTCV) Challenge.[114] Instead of redesigning the entire architecture, the 3DSAM-adapter [115] integrates lightweight 3D adapter modules into a pretrained 2D SAM backbone, enabling volumetric segmentation by fine-tuning only the adapter modules. MedLSAM [116] adopts a complementary strategy that decouples the delineation task into two distinct stages, where a few-shot localization model identifies the target anatomical region to generate bounding boxes that guide SAM segmentation. ProtoSAM-3D [117] extends SAM to volumetric segmentation by coupling instance mask generation with mask-level prototype learning using a spatially-aware transformer to fuse features and classify mask embeddings against learned prototypes to improve consistency. To further improve cross-modality generalization, MA-SAM [118] inserts lightweight 3D adapters into SAM's transformer blocks, allowing the model to capture volumetric and temporal information across CT, MRI, and surgical video, while retaining the majority of pretrained parameters.

Recent studies have also leveraged temporal and memory mechanisms to enhance 3D consistency. MedSAM2 [119] extends MedSAM to the 3D domain by leveraging the Segment Anything Model 2 (SAM2) architecture and treating volumetric segmentation as a video object tracking problem. Utilizing a self-sorting memory bank to condition each slice based on previously processed frames, MedSAM2 supports one-prompt segmentation across 3D volumes. Similarly, AFTer-SAM [120] integrates a trainable axial fusion transformer after the SAM encoder, to capture inter-slice and intra-slice contextual information to improve 3D segmentation accuracy. Extending the capabilities of MedSAM2, ECT-3DMedSAM [121] proposes a semi-supervised cross-teaching mechanism with shared frozen encoders and two LoRA-adapted mask decoders to facilitate domain adaptation with limited annotations.

#### 4.1.2 Prompt Optimization

Conventionally, SAM relies on user-defined prompts, such as points or bounding boxes, to generate segmentation masks. This intrinsic dependency has catalyzed the development of a distinct paradigm that focuses on automating and optimizing prompt generation to reduce manual interaction while maintaining segmentation accuracy.

**Prompt Automation:**

In auto-prompting, prompts are inferred directly from input images or intermediate features. AutoSAM [122] exemplifies this by implementing a fully automated pipeline that predicts prompts through a trainable auxiliary prompt encoder based on a HarDNet [123] backbone. ESP-MedSAM [124] introduces a trainable self-patch prompt generator (SPPG) to produce dense prompt embeddings from image features. Sam2Rad [125] augments the architecture with a trainable prompt predictor network to automate the generation of bounding boxes and mask prompts. Notably, Sam2Rad also supports a semi-autonomous workflow, where users can iteratively refine the segmentation through additional prompts when needed. Several complementary approaches also leverage auxiliary detection or coarse localization to derive spatial prompts. Pandey et al. [126] integrate YOLOv8 with SAM and HQ-SAM for spatial prompt generation by training YOLOv8 on a small set of ~100 image-mask pairs to predict bounding boxes. Similarly, ProtoSAM [127] uses a DINOv2-augmented prototypical network to generate coarse masks, from which point-and-box prompts are extracted for one-shot segmentation. Alleviating the reliance on external modules, Semi-Supervised SAM [128] incorporates physical constraints with a sliding window mechanism to automatically generate reliable pseudo-mask prompts from unlabeled data. RRL-MedSAM [129] further optimizes the segmentation pipeline by introducing an auto-prompting decoder that uses intermediate segmentation outputs to guide the final segmentation.

Reducing dependence on explicit auxiliary networks, SIMSAM [130] improves zero-shot segmentation performance by simulating user interactions during inference. In contrast, EviPrompt [131] proposes a training-free evidential prompt generation technique that requires only a single reference image-annotation pair to generate prompts for new target images based on feature similarity. Extending prompt automation to volumetric segmentation, AutoProSAM [132] combines PEFT with U-Net-like trainable auto-prompt generator (APG) for 3D multi-organ segmentation. Similarly, RFMedSAM2 [133] utilizes U-Net as an initial mask predictor within a multi-stage refinement pipeline to generate bounding box prompts for SAM2. SAM2-SGP [134] leverages ICL with a pseudo-mask generation module to produce initial masks from a support set, which are then used by a pseudo-mask attention module to guide segmentation without manual prompting.

As an alternative to spatial prompts, Zeus [135] utilizes an LLM to generate natural language instructions that act as prompts, enabling zero-shot segmentation. Extending this language-driven paradigm to visual prompting, CLISC eliminates the need for manual interaction by leveraging CLIP to generate enhanced class activation maps (CAMs), which are then automatically converted into bounding box and point prompts to guide SAM for unsupervised brain tumor segmentation.

**Prompt Optimization:**

Complementing auto prompt generation, iterative prompt optimization refines existing prompts to improve segmentation quality. For example, SAM-U [136] employs test-time prompt augmentation to generate multiple perturbed bounding boxes around a target and utilizes uncertainty estimation to select the best mask. ClickSAM [137] adopts a two-stage fine-tuning strategy that iteratively improves segmentation accuracy during training by incorporating positive and negative click prompts, simulating clinical interactions.

#### 4.1.3 Hybrid SAM Architectures

Hybrid SAM architecture integrates SAM with complementary DL approaches to enhance local feature representations and fine-grained boundary delineation. For example, SAM-UNet [138] augments the frozen SAM ViT encoder with a trainable CNN branch (ResNet) and employs a multiscale fusion decoder to capture fine-grained local features, enhancing zero-shot segmentation performance on medical images. Similarly, MedSAM-CA [139] and SAM-LST [140] augment the SAM encoder with a lightweight CNN encoder. However, they adopt distinct feature fusion strategies: MedSAM-CA incorporates an attention-enhanced feature fusion block in the decoder to adaptively combine local and global features, whereas SAM-LST employs a learnable gating mechanism to fuse encoder feature maps before passing them to the decoder. SAMDA [141] addresses few-shot domain adaptation by concatenating feature embeddings from a frozen SAM-based encoder and a trainable U-Net encoder, utilizing perceptual loss to minimize domain divergence in the embedding space. Another notable implementation in the hybrid SAM paradigm is the one-shot localization and segmentation framework [142], which integrates SAM into a two-step pipeline, wherein a pretrained vision backbone is used to generate localized point prompts to guide the final segmentation.

Adapting hybrid SAM architectures to volumetric segmentation. ProMISe [143] adapts the 2D SAM model for 3D tasks by using a hybrid encoder that combines a frozen SAM ViT with a lightweight 3D CNN, by introducing depth-wise adapters into the transformer to capture volumetric spatial context. To improve inter-slice consistency, SLM-SAM2 [144] incorporates a dynamic short-long memory module between the SAM2 encoder and decoder, leveraging both recent and long-range contextual information to prevent error propagation during volumetric annotation. TAGS [145] extends the hybrid paradigm by introducing a 3D SAM Adapter and a multi-prompt fusion framework that integrates automated organ prompts. SAM2Med3D [146] treats 3D breast MRI scans as video sequences by combining SAM2 with an nnU-Net backbone. In addition, it uses spatial filtering to select high-confidence slices from the nnU-Net prediction as prompts and uses a confidence-driven fusion mechanism to adaptively merge the coarse and refined outputs for improved volumetric segmentation consistency.

#### 4.1.4 Efficient and Lightweight SAM

Despite its strong generalization capability, the deployment of SAM in resource-constrained medical settings remains a challenge due to its heavy computational and memory requirements [147]. To address this, numerous lightweight SAM variants have been proposed through architectural adaptation, knowledge distillation, and hardware-focused optimization.

Knowledge distillation is a key strategy used for compressing the SAM ViT encoder into a lighter backbone. For instance, LiteMedSAM [148] employs distillation to transfer knowledge from the MedSAM encoder into a lightweight TinyViT [149] encoder, facilitating efficient segmentation across diverse modalities. Swin-LiteMedSAM [150] substitutes the TinyViT encoder with a tiny Swin transformer and establishes skip connections between the encoder and mask decoder to enhance feature integration, achieving better stability across modalities. To further reduce the parameter count, SwiftMedSAM [151] compresses LiteMedSAM by reducing the transformer block depths, MLP dimensions, and attention heads, achieving an ultralightweight configuration with a 40% reduction in parameters (~5.9M).

Lightweight CNN backbones have also been adopted instead of transformers to accelerate inference on edge devices. RepViT-MedSAM [152] substitutes the MedSAM encoder with a RepViT [153] CNN backbone, employing aggressive data augmentation during distillation to achieve an average CPU inference time of 2.53 seconds per 2D image, surpassing LiteMedSAM (~ 3.18 seconds). Similarly, RepMedSAM [154] utilizes a RepViT and adopts a two-stage training pipeline: knowledge distillation using a pretrained TinyViT teacher, followed by fine-tuning to optimize inference speed. In addition to architectural modifications, recent studies have demonstrated that runtime optimizations can significantly improve deployment efficiency. MedficientSAM [155] leverages an EfficientViT [156] encoder and re-implements the C++ inference pipeline, achieving a CPU inference time of ~1.0 second per 2D image. Likewise, DAFT [157] combines an EfficientViT backbone with the OpenVINO runtime and domain-specific model selection, achieving an inference speedup by a factor of 6.5 over LiteMedSAM. For low-latency memory-efficient volumetric segmentation, FastSAM3D [158] combines progressive distillation with 3D sparse flash attention to reduce the computational demand achieving 8.75× speedup over 3D SAM variants, enabling interactive 3D segmentation on GPU platforms.

To enhance generalization and reduce manual intervention, several computational approaches have been proposed to compress image encoders by integrating computationally efficient automatic prompting mechanisms. De-LightSAM [159] proposes a modality-decoupled architecture featuring a domain-controllable encoder and a query-decoupled modality decoder to handle cross-modality interference. It also employs an SPPG to eliminate manual annotation, achieving a parameter count of ~ 2% SAM-H. In parallel, Med-FastSAM [160] integrates a lightweight knowledge aggregation encoder with a coarse prompt encoder to reduce the parameter count (15.45% of conventional SAM), enabling fully automated segmentation. MCP-MedSAM [161] adopts a different approach compared to auto-prompting by enhancing the standard bounding box with two trainable prompts: a modality prompt that aids the model in adjusting to different types of medical images, and a content prompt to accurately identify the target object. Furthermore, MCP-MedSAM replaces standard slice-wise training with a target modality-based data sampling strategy to mitigate training bias, enabling the model to converge in one day using a single GPU.

#### 4.1.5 Domain-Specific SAM

Domain-specific SAM encompasses SAM-based adaptations that enhance segmentation fidelity within a particular imaging domain by addressing modality-specific challenges and constraints. Although the foundational SAM architecture demonstrates superior generalization performance, its clinical applications necessitate architectural modifications, training strategies, or prompt engineering techniques. Accordingly, this section categorizes domain-specific SAM variants according to the target imaging modality and clinical context for which they have been developed and deployed.

**Ultrasound Imaging:**

In US imaging, SAM adaptations address modality-specific characteristics that cause standard encoders to discard high-frequency boundary details, such as speckle noise, low signal-to-noise ratio, and high anatomical variability. To recover these local features, hybrid architectures such as SAMUS [162] and CC-SAM [163] integrate a CNN branch alongside the SAM ViT encoder to capture high-frequency details. In particular, SAMUS utilizes a cross-branch attention mechanism to fuse global ViT features and local CNN features. CC-SAM instead employs a variational attention fusion module to probabilistically merge global and local features to handle the uncertainty inherent in low-contrast US images. Furthermore, CC-SAM incorporates ChatGPT-generated text prompts converted into bounding box prompts to guide US segmentation. Utilizing a robust, data-centric approach, SonoSAM [164] adapts the conventional SAM to US imaging by fine-tuning the mask decoder and prompt encoder on ~200,000 US images. SonoSAMLite [164] further distills SonoSAM via knowledge distillation, thus reducing the parameter count from 90M to 28M with minimal loss in segmentation accuracy. To alleviate manual interaction, AutoSAMUS [162] extends SAMUS by replacing the manual prompt encoder with a learnable APG to generate prompt embeddings for end-to-end segmentation. To address temporal consistency in echocardiography video segmentation, MemSAM [165] introduces a spatiotemporal memory mechanism to prompt the segmentation sequentially frame-by-frame, using a memory reinforcement module to filter out speckle noise before updating the memory bank. Similarly, SonoSAMTrack [166] extends SonoSAM to US video by combining the DeAOT tracking framework, which propagates dense segmentations across frames for efficient segmentation.

**Surgical Imaging:**

In surgical and endoscopic imaging, SAM adaptations primarily focus on automating prompt generation for real-time instrument tracking and addressing the scarcity of dense annotations for anatomical structures. For instance, Surgical-DeSAM [167] eliminates the need for manual interaction during surgery by decoupling the SAM architecture and replacing the SAM image encoder with a Swin-DETR [167] detector. This approach enables end-to-end instrument detection by automatically feeding generated bounding boxes to prompt the mask decoder. SurgicalSAM [168] enhances SAM by introducing a lightweight prototype-based class prompt encoder that leverages contrastive learning to generate prompt embeddings from instrument category prototypes, bypassing the dependence on explicit spatial prompts. To achieve fine-grained, part-specific delineation, SurgicalPart-SAM [169] incorporates expert knowledge of instrument structure through collaborative prompts combining category and part-level text descriptions, fused via a cross-modal encoder to accurately segment intricate components, such as shafts and wrists. Emphasizing generalization, SAMSurg [170] fine-tunes the SAM mask decoder on a curated dataset of 77,000 labeled surgical image–mask pairs, demonstrating strong performance on unseen instrument types. To facilitate text-prompted segmentation, AdaptiveSAM [171] employs a text affine layer to adapt CLIP embeddings to medical terminology alongside a parameter-efficient bias-and-norm-tuning strategy to adapt to surgical scenes with minimal computational cost. SurgiSAM2 [172] adapts SAM2 for surgical video segmentation by fine-tuning on limited subsets, exhibiting superior performance in segmenting both anatomical tissues and instruments.

**Pathology:**

In digital pathology, SAM adaptations address domain-specific challenges arising from the multiscale pyramidal structure of whole slide images (WSIs), high density cellular objects, significant staining heterogeneity and limited inherent semantic discrimination. For example, SAM-Path [173] enables multiclass semantic segmentation without manual input by utilizing trainable class prompts as cues for pathological-specific entities. In particular, it incorporates a parallel pathology-specific encoder derived from a hierarchical image pyramid transformer [174] to inject domain-specific features alongside the standard SAM image encoder. To address the multi-resolution nature of WSIs, WSI-SAM [175] adapts SAM to large structure segmentation by introducing high-resolution and low-resolution tokens within a dual mask decoder to fuse features across scales, preserving zero-shot capabilities. Similarly, SegAnyPath [176] utilizes a task-guided mixture of experts (MoE) decoder to route features for distinct tasks, such as cell, tissue, or tumor segmentation, and mitigates staining variations through self-distillation and augmentation.

A critical bottleneck in adapting SAM to WSIs is the impracticality of providing manual prompts for segmentation tasks involving numerous object instances. UN-SAM [177] addresses this limitation by devising a multiscale prompt generator that automatically creates high-quality masks to guide the decoder. It further utilizes a domain-adaptive tuning encoder to harmonize common and domain-specific knowledge for robust nuclei instance segmentation. Similarly, CellSAM [178] combines the SAM encoder with an object detector termed CellFinder to detect cells and generate bounding box prompts for the mask decoder to produce the final segmentation masks. This approach enables CellSAM to generalize across cellular imaging data with a strong zero-shot capability. μSAM [179] extends SAM to electron microscopy by implementing a new decoder architecture to improve instance segmentation performance on multidimensional data (2D and 3D).

### 4.2 Generalist VFMs

Beyond SAM, generalist VFMs have garnered significant attention in medical image analysis due to their ability to support diverse downstream tasks utilizing task-agnostic backbones pretrained via large-scale SSL or weakly SL. Unlike SAM-based models, which share a common architectural scaffold, generalist VFMs exhibit substantial architectural diversity, spanning across ViTs, hybrid CNN-transformers, U-Net variants, diffusion models, and state-space models. Similarly, they also combine multiple training paradigms, ranging from MIM and contrastive learning to generative pretraining, limiting categorization based solely on the optimization approach. Accordingly, this section organizes generalist VFMs according to their primary downstream task into task-agnostic generalists, followed by VFMs specialized for segmentation, classification, registration, and image synthesis.

#### 4.2.1 Task-Agnostic VFMs

In medical image analysis, task-agnostic generalist VFMs are designed to generalize across downstream tasks using a single pretrained backbone. Broadly, these task-agnostic models can be categorized into cross-modality VFMs pretrained on heterogeneous cross-domain datasets and modality-specific VFMs pretrained on large modality-specific corpora.

**Cross-Modality VFMs:**

In cross-modality VFMs, a key challenge is the development of unified representations that can generalize across heterogeneous imaging modalities. MerMED-FM [180] addresses this challenge by pretraining on over 3.3M images across seven imaging modalities and ten clinical specialties. By combining a self-supervised teacher-student framework with a dynamic memory module, MerMED-FM retains cross-modality knowledge, enabling classification across radiology, ophthalmology, pathology, and dermatology. Scaling this paradigm, Curia [181] leverages DINOv2 [78] pretraining on over 200M CT and MRI images from clinical examinations, demonstrating robust generalization across organ identification, pathology detection, and tumor staging.

**Modality-Specific VFMs:**

**Radiology:** In radiology, chest X-ray (CXR) imaging has witnessed a rapid growth in VFMs due to the widespread availability of large public datasets. For instance, CheXFound [22] utilizes a task-agnostic ViT backbone pretrained on ~1M CXR images, supporting tasks such as cardiovascular risk prediction and anatomical segmentation. Similarly, RayDINO [182] employs DINOv2 [78] self-distillation on 873,000 CXR images, surpassing supervised baselines including classification and segmentation. Moving beyond single-objective SSL, EVA-X [24] unifies contrastive learning and MIM within a single pretraining framework to capture both semantic and geometric features, enabling effective few-shot transfer to unseen tasks. CXRBase [183] and RadLAS [23] further extend these capabilities: CXRBase adopts a two-stage MAE pretraining approach on ~1.04M CXR images, while RadLAS utilizes lesion-grounded reconstruction with contrastive learning on ~ 294, 000 CXR images, supporting generalizable multi-disease diagnosis and localization.

In US imaging, USFM [184] leverages the 3M-US database and spatial-frequency dual masking technique to extract robust features for segmentation, classification, and enhancement. OpenUS [53] explores a Vim backbone with masked contrastive learning to improve label-efficient classification and segmentation across anatomical regions. In CT imaging, LCTFound [185] demonstrates the effectiveness of a hybrid U-Net transformer architecture with diffusion-based pretraining for diagnostic and restoration tasks. Complementing this, CT-FM [186] utilizes contrastive pretraining on ~148,000 CT scans to yield anatomically coherent 3D representations, generalizing to whole-body and tumor segmentation, head CT triage, and semantic retrieval.

Extending task-agnostic learning to volumetric MRI, the Medical Transformer [187] transforms 3D scans into multi-view 2D slice sequences within a hybrid CNN-transformer framework to model inter-slice dependencies for classification, regression, and segmentation tasks. MRI-CORE [188] scales this approach by pretraining on ~116,000 3D volumes (Duke-110K) using DINOv2 [78] with SAM-based initialization, supporting few-shot classification and zero-shot segmentation across anatomical regions. Focusing on image enhancement, BME-X [189] serves as a tissue-aware foundation model for motion correction, super-resolution, and denoising, significantly benefiting downstream tasks, such as registration. Similarly, BDFM [190] targets brain diseases by pretraining on ~15,300 3D MRI scans using a spatial-frequency dual-domain decoder to capture high-frequency lesion features. Complementing BDFM's lesion-focused design, BrainIAC [191] utilizes a SimCLR-based contrastive pretraining strategy with a modified ResNet50 backbone to extract robust representations from unlabeled MRI scans, enabling downstream tasks such as sequence classification, brain-age estimation, molecular status prediction, and survival modeling. For cardiac imaging, CineMA [192] pretrains a multi-view convolution-transformer MAE on ~75,000 cine CMR studies, demonstrating robust generalization in ventricular segmentation, ejection fraction estimation, and disease diagnosis across diverse populations.

**Ophthalmology:** In ophthalmology, the development of generalists is driven by the diversity of modalities, such as color fundus photography (CFP), OCT, and slit-lamp imaging. VisionFM [193] addresses this diversity by pretraining on ~3.4M ophthalmic images across eight modalities, learning both modality-specific and cross-modality representations for downstream tasks, such as disease screening, prognostic modeling, and anatomical structure segmentation. Similarly, EyeFound [21] learns unified representations from ~2.78 M unlabeled images from 11 ophthalmic modalities, exhibiting strong generalizability and performance across retinal disease classification, disease prediction, and zero-shot VQA. Focusing specifically on retinal imaging, RETFound [194] pretrains a ViT-Large backbone via MAE on 1.6 million unlabelled CFP and OCT images, enabling label-efficient adaptation to ocular disease diagnosis, prognosis, and systemic disease prediction including cardiovascular and neurodegenerative conditions.

**Pathology:** In pathology, bridging the gap between tile-level and whole-slide representations has led to the development of hierarchical VFMs. CHIEF [195] adopts a two-stage pipeline utilizing tile-level SSL, followed by weakly supervised WSI-level pretraining, to capture multiscale features for cancer detection and survival prediction. Augmenting hierarchical encoding, TITAN [196] extends this paradigm through vision-language alignment. Employing a three-stage strategy: iBOT-based visual self-distillation, followed by CoCa-style cross-modal pretraining using 423,122 synthetic ROI captions and 182,862 pathology reports. TITAN demonstrates superior performance, outperforming CHIEF on morphological subtyping, zero-shot classification, and zero-shot report generation [196]. In contrast to these hierarchical approaches, BEPH [197] leverages BEiTv2-based MIM on ~11.77M patches from 32 TCGA [198] cancer types, achieving strong performance in patch-level classification, WSI subtyping, and survival prediction under limited supervision. PathOrchestra [199], extends this approach by pretraining on 287,424 WSIs across 21 tissue types and evaluating on 112 clinical tasks, including pan-cancer classification, segmentation, biomarker assessment, gene expression prediction, and structured report generation. Complementing these pan-cancer frameworks, HistoEncoder [200] demonstrates the value of organ-specific pretraining by utilizing DINO self-distillation on 48 million prostate tissue tiles with an XCiT backbone, enabling efficient adaptation to cancer classification, unsupervised tissue annotation, and survival prediction.

Addressing the computational demands in gigapixel WSI processing. PLUTO [201] adopted a lightweight ViT-S backbone and a flexible patch-size framework for multiscale feature extraction across different WSI resolutions, enabling deployment in resource-constrained settings while maintaining strong generalization for instance segmentation, tile classification, and WSI-level prediction.

### 4.2.2 Segmentation VFMs

Generalist segmentation VFMs have emerged as non-SAM alternatives for delineating anatomical structures and pathological regions. Although some of these VFMs utilize task-agnostic backbones, their primary design objective focuses on segmentation. Moreover, their generalization scope varies in practice, ranging from universal segmentation across modalities to domain-specific delineation. Accordingly, we categorize them into universal segmentation and domain-specific segmentation VFMs.

**Universal Segmentation VFMs:**

Universal segmentation VFMs achieve broad generalization capability across diverse imaging modalities utilizing large-scale SSL, adaptive conditioning, or supervised pretraining. In adaptive conditioning, a task context is utilized to induce generalization without fine-tuning. A notable implementation is UniverSeg [20], which employs a cross-block mechanism to transfer information from a small support set of image-label pairs to a query image, enabling the segmentation of unseen tasks without additional training. In contrast, SSL-driven universal segmentation VFMs leverage masked modeling or self-distillation pretraining (Section 3.2.2) to learn modality-agnostic features with minimal downstream adaptation. For example, VIS-MAE [71] leverages a MAE architecture pretrained on 2.5M images across five modalities to learn modality-invariant representations for medical image segmentation. Similarly, MIS-FM [202] pretrains both the encoder and decoder via a volume-fusion self-supervision strategy within a parallel convolution and transformer network, enabling robust performance across diverse 3D CT segmentation tasks under limited task-specific supervision.

In contrast, DINO-based segmentation backbones, such as MedDINOv3 [203] and Dino U-Net [204], adapt self-distilled representations for segmentation via multiscale token aggregation or fidelity-preserving projection modules, acheiving superior segmentation performance across CT, and MRI benchmarks. To explicitly address segmentation robustness across modalities, modality-adaptive mechanisms introduce dynamic, modality-aware components within shared architectures. For instance, M$^4$oE [205] improves cross-modality performance by augmenting a SwinUNet backbone with modality-specific expert MLPs and a dynamic gating network that adaptively fuses expert outputs for efficient segmentation across CT and MRI. Similarly, MPUM [206] employs a modality-projection controller to generate modality-specific convolutional kernels while preserving shared anatomical knowledge, enabling whole-body segmentation across CT, MRI, and PET. Extending beyond modality-specific adaptive fusion, MedSegX [207] introduces a contextual mixture of adapter experts to dynamically route task-relevant features through context-aware adapters to mitigate negative transfer, achieving strong cross-site and cross-task segmentation performance across 10 imaging modalities.

Specific architectures have also been developed to address tumor segmentation and longitudinal tracking. TSFM [208] combines a ResBlock backbone with a transformer bottleneck to capture both local and global features for unseen tumor segmentation across MRI and CT. Extending tumor-focused universal segmentation to volumetric data, SAT3D [209] integrates a Swin transformer with an uncertainty-aware training pipeline, explicitly leveraging uncertainty estimates as prompts to guide reliable boundary prediction in low-contrast regions across CT, MRI, PET, and US. Recently, universal segmentation has been extended to longitudinal and spatiotemporal settings. LesionLocator [210] couples zero-shot lesion segmentation with follow-up tracking by using dense prompts and autoregressive mask propagation to locate and segment lesions across serial scans. Similarly, iMOS [211] adapts a pretrained memory-based segmentation backbone with lightweight PEFT adapters to support moving-object segmentation across modalities, including MRI, CT, US, and endoscopy. Beyond modality conditioning and self-supervised pretraining, universal segmentation has also been achieved through large-scale supervised pretraining. STU-Net [212] exemplifies this by training on the TotalSegmentator dataset, demonstrating strong generalization across 14 downstream datasets spanning diverse modalities and anatomical targets.

**Domain-Specific Segmentation VFMs:**

Domain specific segmentation VFMs aim to generalize across multiple segmentation tasks within a specific imaging modality or clinical domain, leveraging modality-specific priors and specialized architectures. In practice, these domain-specific VFMs can be broadly categorized based on their clinical domain.

**Radiology:** Within radiology, domain-specific VFMs span multiple imaging modalities and anatomical subdomains. In US imaging, ULS4US [213] proposes a multiple-in-multiple-out (MIMO) U-Net backbone with a two-stage, lesion-aware learning strategy to refine multi-organ lesion segmentation across diverse US views. Similarly, MOFO [214] introduces a universal framework for multi-organ US segmentation by combining a Swin transformer-based encoder-decoder with a prompt branch that generates organ-specific representations and an anatomical prior branch to impose structural constraints.

In MRI, F3-Net [215] extends the conventional nnU-Net backbone with modality-specific encoders and a zero-image strategy for handling missing brain MRI sequences for robust lesion segmentation. To further exploit multimodal data, the mixture of modality experts [216] framework integrates modality-specific expert networks with a hierarchical gating mechanism to achieve voxel-wise adaptive fusion of expert outputs for robust brain MRI lesion segmentation. To improve task adaptability, UniMRISegNet [217] uses a contextual prompt generation module and prompt-conditioned dynamic convolutions within a shared U-Net backbone, enabling flexible adaptation to diverse MRI segmentation tasks. BrainSegFounder [218] and RoMedFormer [219] utilizes staged SSL with task-specific fine-tuning to improve cross-task generalization in MRI. Hybrid domain-specific designs improve anatomical segmentation by combining complementary backbones with SSL-based pretraining. For example, 3D-SCUMamba [220] embeds a Mamba Bottleneck module into a hybrid 3D U-Net to capture long-range spatial dependencies for improved CT abdominal tumor segmentation.

**Ophthalmology and Pathology:** Beyond radiology, domain-specific models address challenges in retinal and pathological imaging. In retinal imaging, URVSM [221] employs a two-stage pipeline integrating a CycleGAN-based image translation with a topology-aware segmentation to achieve robust vessel delineation across multiple retinal imaging modalities. In pathology, PrPSeg [222] extends a residual U-Net with a universal proposition matrix and token-based dynamic head for multiscale renal structure segmentation, including the cortex, medulla, glomeruli, and mesangial cells.

### 4.2.3 Classification VFMs

Unlike segmentation-oriented frameworks that generate spatial masks over anatomical or pathological regions, classification-focused VFMs learn high-level semantic representations that are mapped to clinical labels representing disease presence, severity, or progression. Because these prediction tasks are tightly coupled to the clinical context of a specific anatomical region or imaging modality, classification-focused VFMs are often domain-specific. Accordingly, this section categorizes these classification VFMs by clinical domain, summarizing their scope, architectural design, and pretraining strategies.

**Radiology:** In radiology, VFMs can be classified based on CXR, CT, MRI, and US imaging modalities. In CXR imaging, Ark$^{+}$ [223] combines a Swin transformer backbone with a cyclical teacher–student strategy to integrate knowledge from six public CXR datasets, improving thoracic disease classification and diagnosis. CNTD-Net [224] targets neuro-trauma triage on non-contrast head CT by leveraging an LLM-assisted multilabel annotations, achieving high accuracy in neuro-trauma findings. FM-HCT [225] further extends head CT analysis by pretraining a 3D ViT on 361,663 scans via dual DINO and MAE objectives for generalizable disease detection. DenseFormer-MoE [226] further improves brain MRI analysis by combining a hybrid DenseNet-Transformer backbone with MAE pretraining and MoE to support concurrent neurodegenerative disease diagnosis alongside brain age prediction. MRI-PTPCa [227] extends cross-modal MRI representation learning to prostate cancer by utilizing a multi-input ViT trained on a large-scale MRI dataset, enabling noninvasive Gleason grading.

In US imaging, ProstNFound [228] integrates the image encoder and decoder of MedSAM with a conditional prompting module to fuse high-resolution micro-US cues with structured clinical data to support domain-aware prostate cancer analysis. URFM [229] improves multi-organ generalization by pretraining on over 1M images across 15 anatomical organs using representation-based MIM, achieving superior performance across 10 diagnostic tasks including nodule classification, tumor grading, liver mass diagnosis, and fetal plane recognition. Spanning multiple modalities, Radio DINO [230] explores self-supervised ViT pretraining for radiomics using the RadImageNet dataset, leveraging DINO and DINOv2 self-distillation techniques, achieving state-of-the-art classification on multiple MedMNISTv2 benchmarks.

**Ophthalmology:** In ophthalmology, XFM [231] targets rare retinal disease classification under constrained data settings by fine-tuning pretrained ophthalmic backbones (RETFound and VisionFM). Furthermore, to refine the activation maps, XFM incorporates saliency-based prior-knowledge infusion and SAM-guided cycle constraint regularization, achieving ~91% accuracy on the FundusData-FS dataset [231].

**Pathology:** In pathology, classification-oriented VFMs have focused on learning discriminative representations from high-resolution WSIs for cancer diagnosis and prognosis. For example, UNI [232] combines DINOv2 self-distillation with MIM to pretrain a ViT-L backbone on Mass-100k (100M tissue patches) supporting cancer subtype classification, tissue structure recognition, and few-shot learning. Scaling this paradigm further, Virchow [233] utilizes a ViT-H backbone combined with DINOv2 to train on ~1.5M WSIs, achieving an area under the curve (AUC) of ~0.95 for pan-cancer detection, including common and rare cancers.

To fully capture contextual information from gigapixel WSIs, models must address the representational disparity between local tile-level features and global whole-slide architectures. Prov-GigaPath [234] addresses this through a two-level encoder architecture combining tile-level DINOv2 pretraining and slide-level LongNet modeling. Pretrained on the Prov-Path dataset comprising 1.3B tiles from 171,189 WSIs, Prov-GigaPath demonstrates strong performance in cancer subtyping and mutation prediction [234].

Addressing the computational demands of gigapixel WSI processing. VFMs, such as OpenSlideFM [235], and Giga-SSL [236] adopts efficient pretraining strategies. Specifically, OpenSlideFM utilizes a lightweight ConvNeXt-Tiny architecture (35M parameters) with dual-scale feature extraction for training and deployment on a single GPU. In contrast, Giga-SSL leverages offline tile augmentations and a submanifold sparse convolutional network to handle gigapixel images with reduced memory bottlenecks. To address stain heterogeneity, PathoDuet [237] trains separate ViT encoders for hematoxylin and eosin (H&E) and immunohistochemistry (IH) slides using cross-scale positioning and cross-stain transferring pretext tasks, enabling the model to translate H&E structural understanding into IHC marker evaluation.

#### 4.2.4 Image Registration and Synthesis VFMs

**Image Registration:**

Image registration is a core component of medical imaging pipelines, enabling longitudinal monitoring, multimodal fusion, and anatomical alignment. Accordingly, these registration-oriented VFMs can be broadly categorized into end-to-end frameworks that learn transferable deformation priors, feature-based methods utilizing FM representations, and interactive frameworks incorporating user guidance.

For instance, uniGradICON [238] combines GradICON [239] with a U-Net backbone pretrained on a composite multi-anatomy dataset, enabling robust zero-shot generalization across monomodal registration tasks. MultiGradICON [240] extends uniGradICON to multimodal registration by incorporating a squared local normalized cross-correlation loss ($1 - LNCC^2$) to handle cross-modality intensity differences, outperforming uniGradICON on multimodal pairs. Complementing these deformation-based approaches, DINO-Reg [241] adopts DINOv2 as a universal feature extractor to project slice-wise features into a shared space via principal component analysis (PCA), achieving strong performance on both monomodal and multimodal registrations. DINO-Reg-Eco [241] replaces the DINOv2 encoder with a lightweight 3D U-Net to reduce the encoding time by 99% while preserving registration accuracy. Det-SAMReg [242] couples Grounding DINO [243] text-prompted detection with SAM2 segmentation to automatically generate organ masks for optimization-based MRI-to-US alignment, achieving strong performance with as few as 14 fine-tuning samples. UniReg [244] introduces an interactive registration VFM by combining self-supervised anatomical embedding with a conditional control mechanism to adapt for diverse registration tasks with minimal fine-tuning.

**Image Synthesis:**

Generalist VFMs have also been extended to produce high-fidelity synthetic medical images while preserving fine anatomical structures. A notable implementation is MedDiff-FM [74], a diffusion-based FM trained on 5376 multi-region CT volumes, combining multilevel processing, 3D positional embeddings, and conditional anatomical guidance to achieve superior denoising, anomaly detection, and image synthesis without fine-tuning. MAISI [245] scales this approach to volumetric synthesis by using a three-stage diffusion framework combining a variational autoencoder–generative adversarial network based volume compression, latent diffusion generation, and ControlNet based fine-grained anatomical conditioning across 127 structures, enabling the synthesis of high-quality 3D CT volumes with flexible dimensions and voxel spacing. To overcome the reliance on real data, SynthFM [246] introduces a modality-agnostic synthetic data framework that generates anatomically plausible shapes, diverse textures, and low-contrast boundaries for training segmentation VFMs without real patient data. In pathology, ToPoFM [247] addresses the challenge of synthesizing high-resolution tissue images through a topology-controlled VFM, which uses an LLM guided by a multiscale K-function topology estimator to generate anatomically accurate cell layouts that condition a latent diffusion model to generate high-fidelity histopathology images.

### 4.3 Quantitative Analysis of VFMs

Among the 137 VFM studies included for qualitative synthesis 135 studies published between January 2022 and December 2025 were considered for quantitative analysis. Across these studies, a clear upward trajectory was observed from 2022 to 2025 [Fig. 8(a)], reflecting the expanding interest in the development and application of VFMs in medical imaging analysis. The steep growth between 2023 and 2024 coincides with the release of large-scale pretraining frameworks and publicly available medical imaging datasets. The distribution of these studies based on the dataset accessibility used for VFM development is illustrated in Fig. 8(a). Among the 135 studies, most of the studies (88 studies, 65.2%) were developed exclusively using publicly available datasets, while 35 studies (25.9%) utilized private datasets, and 12 studies (8.9%) employed a combination of private and public data sources. This predominance of publicly accessible datasets highlights the pivotal role of open-access data repositories in enabling large-scale VFM development and fostering reproducibility across the medical imaging community.

Recognizing the pivotal role of open-access data repositories in VFM development, [Fig. 8(b)] illustrates the top 10% of the most frequently used publicly available datasets within each imaging modality for VFM training, categorized by imaging modality. The characteristics of these datasets, including their scale, supported tasks, and access links, are summarized in Table 1Among these, CT-based datasets constitute the largest proportion, followed by endoscopy, MRI, and pathology. While these datasets are grouped by their primary modality, VFM studies have also leveraged heterogeneous datasets from diverse clinical domains to strengthen cross-modal generalization in medical imaging tasks. Notably, multimodal datasets such as AMOS [248] and MSD [249] have been utilized across both multimodal and unimodal studies because of their diverse and high-quality annotations encompassing multiple organs across CT and MRI modalities.

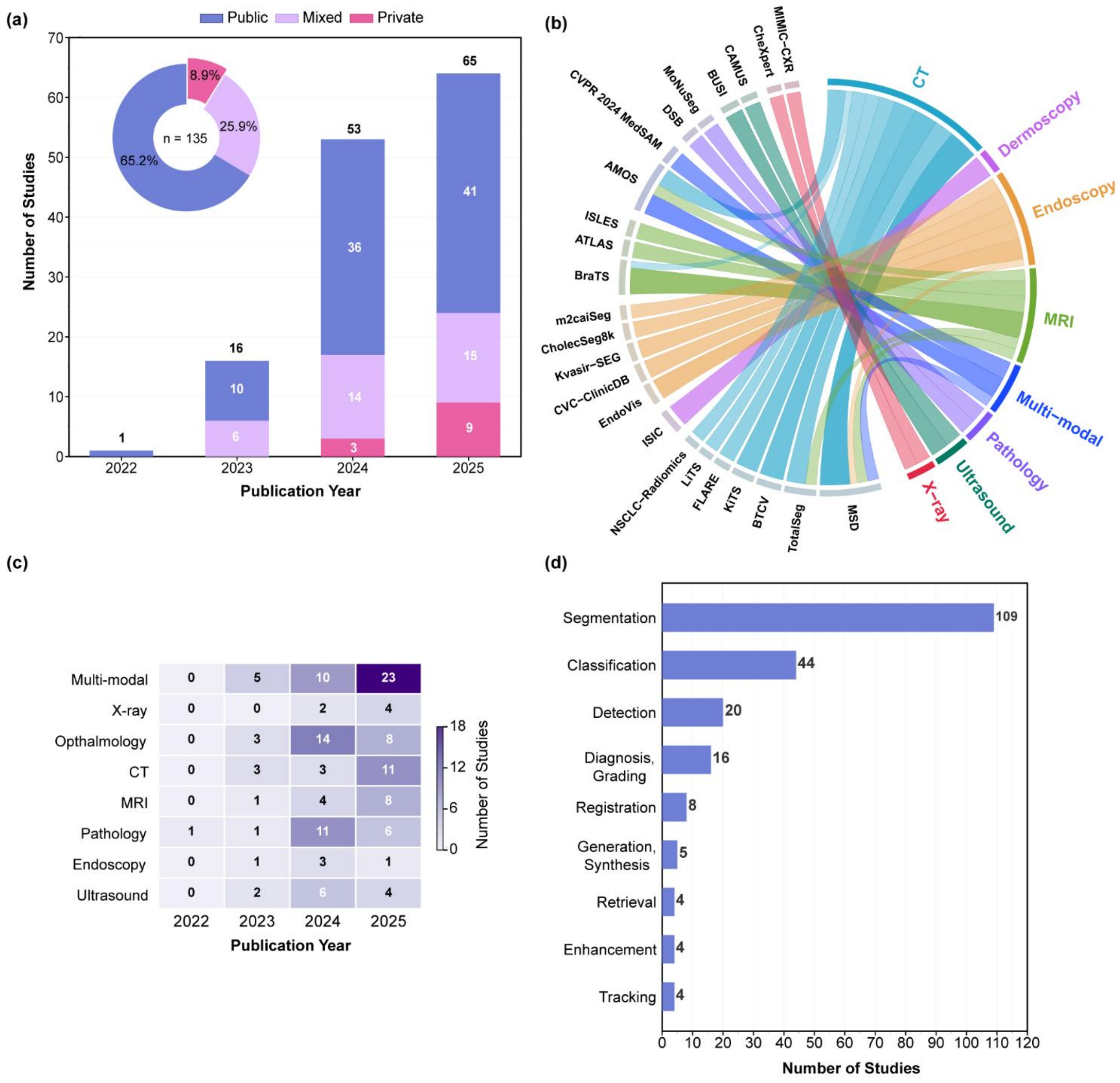

**Fig. 8.** (a) Temporal trend of VFM publications from 2022 to 2025, including the distribution of data types used for VFM development (public, private, or mixed); (b) chord diagram illustrating chord diagram illustrating the top 10% most frequently used public datasets within each imaging modality for VFM development; (c) heatmap of VFM study counts stratified by the primary imaging modality and publication year; and (d) distribution of VFM downstream tasks.

### 4.3.1 Modality and Task Distribution

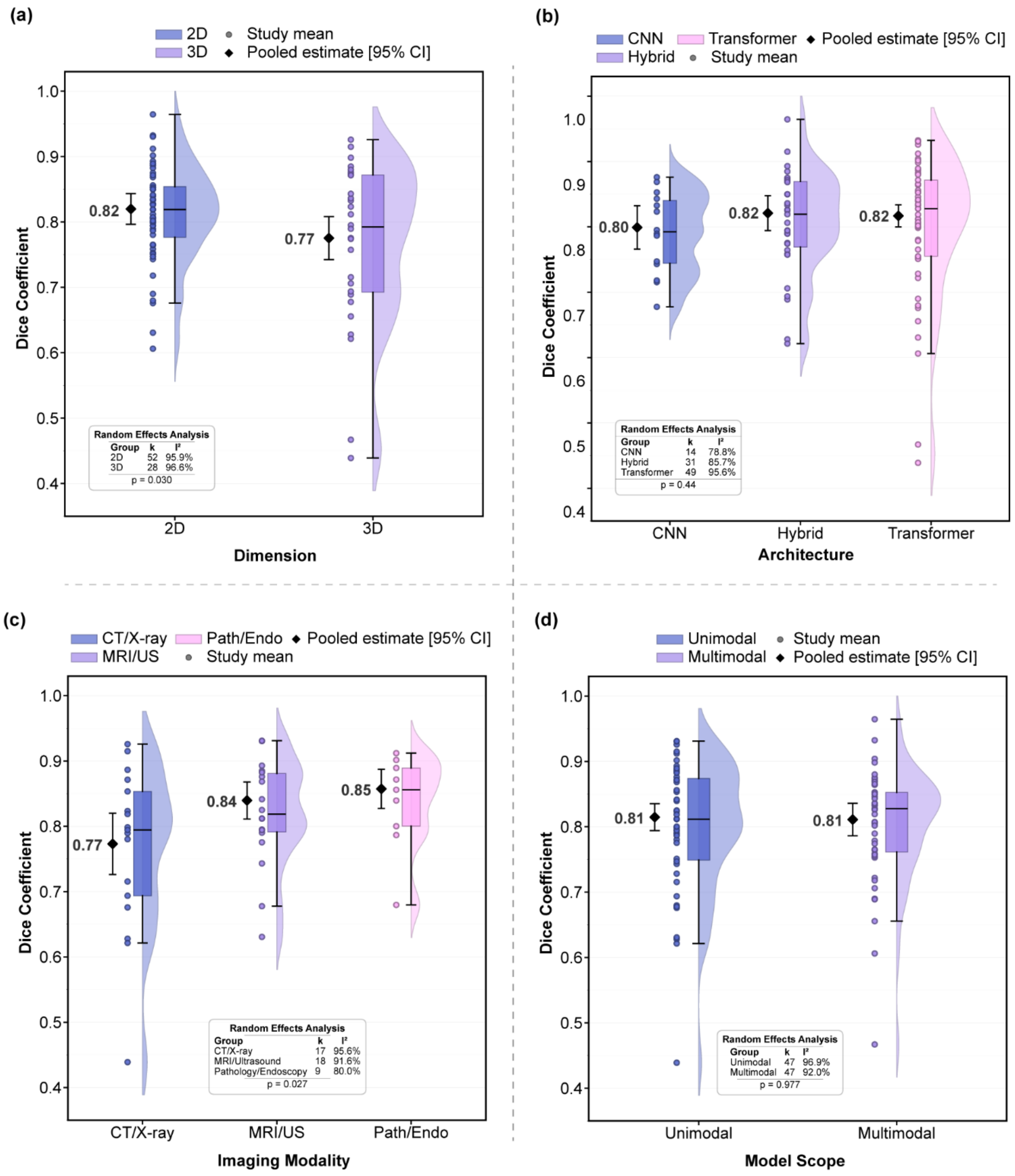

**Fig. 9**. Subgroup analyses of pooled Dice coefficients for VFM segmentation studies stratified by (a) VFM dimensionality (2D vs. 3D), (b) encoder architecture (CNN, hybrid, transformer), (c) imaging modality (CT/X-ray, MRI/US, pathology/endoscopy), and (d) model scope (unimodal vs. multimodal).

The temporal distribution of VFM studies according to their primary imaging modality is depicted in Fig. 8(c). Studies designed for cross-domain applications are grouped under the multimodal category, which constitutes the largest proportion reflecting the community's shift toward multimodal VFMs. Among unimodal studies, ophthalmology-based VFMs exhibited a notable surge in 2024, likely driven by the release of large-scale ophthalmic datasets and dedicated VFMs, such as RETFound and VisionFM. Similarly, CT and MRI-based studies have demonstrated a steady upward trajectory from 2022 to 2025. However, modalities such as endoscopy and X-ray remain underrepresented in VFMs and may benefit from further dataset development and architectural adaptations. [Fig. 8(d)] summarizes the distribution of downstream tasks addressed by VFM studies. As noted in the qualitative synthesis, segmentation emerged as the most predominant downstream task, accounting for the largest share of reported applications, followed by classification, detection, and diagnosis/grading. In contrast, tasks such as image retrieval, image enhancement, and tracking were less frequently explored, indicating that these areas remain relatively underrepresented in current VFM research. The dominance of segmentation reflects the strong influence of SAM-based architectures and their adaptations in the VFM landscape.

### 4.3.2 Pooled Performance Analysis of VFMs

To quantitatively assess the performance of VFMs, a random-effects analysis was conducted on segmentation, the most predominant downstream task. The Dice coefficient was selected as the primary pooled metric, consistent with its widespread adaptation as a standard evaluation measure of segmentation performance [Fig. 8(d)]. Among the 135 VFM studies considered for quantitative analysis, 94 reported segmentation performance using the Dice coefficient and were included in the pooled analysis. Following extraction strategies adopted in recent quantitative syntheses of VLFMs in medical imaging [36]. If a study reports the segmentation performance across multiple datasets, the mean Dice across datasets was computed with standard error. For studies reporting multiple models addressing the same downstream task, the best-performing model meeting the inclusion criteria was extracted. For studies evaluated on a single dataset, a tiered imputation approach was adopted. A DerSimonian–Laird random-effects model [250] was then applied to pool the Dice coefficients across the included studies. The random-effects model was chosen given the anticipated high degree of inter-study heterogeneity arising from differences in model architecture, training strategies, datasets, and evaluation protocols. Heterogeneity was quantified using the $I^2$ statistic and assessed using the Q-test. All statistical analyses were performed using R (version 4.5.2) with the meta [251] and metafor [252, 253] packages, with a two-tailed significance level of $\alpha = 0.05$. The overall pooled mean Dice coefficient across the 94 VFM studies was 0.81 (95% CI: 0.80–0.83, $p < 0.001$), indicating strong segmentation performance of VFMs in medical imaging applications.

### 4.3.3 Subgroup Analysis of VFMs

To investigate potential sources of heterogeneity and evaluate the VFM segmentation performance, we conducted subgroup analyses stratified by four factors: dimensionality, encoder architecture, imaging modality, and model scope. The heterogeneity between the subgroups was assessed using the Q-test for intergroup differences, and $I^2$ statistics were utilized to quantify within-group heterogeneity. The subgroup-level pooled estimates along with 95% confidence interval (CI) and heterogeneity statistics for each subgroup are shown in Fig. 9. The detailed forest plots for each subgroup are provided in the supplementary material.

**Model Dimensionality:** We categorized the 94 studies based on model dimensionality into 2D (k=52) and 3D (k=28) VFMs, excluding the hybrid studies (k=12). Within the subgroups, 2D VFMs achieved a pooled Dice score of 0.82 ($I^2$ = 95.9%), whereas the 3D subgroup exhibited a pooled Dice of 0.77 ($I^2$ = 96.6%) [Fig. 9(a)]. This intergroup difference was statistically significant ($p = 0.03$), suggesting that the segmentation performance of 2D VFMs is superior to that of 3D VFMs. This difference in performance may be attributed to the complexities associated with 3D volumetric processing, such as patch-based inference and down sampling.

**Encoder Architecture:** To assess the influence of the encoder backbone on segmentation performance, we categorized the studies into three subgroups: CNN-based VFMs (k = 14), transformer-based VFMs (k = 49), and hybrid encoder-based VFMs (k = 31). The CNN subgroup achieved a pooled Dice of 0.80 ($I^2$ = 78.8%), while the hybrid and transformer subgroups yielded comparable pooled Dice values of 0.82 ($I^2$ = 85.7%) and 0.82 ($I^2$ = 95.6%), respectively [Fig. 9(b)]. No statistically significant difference in performance was observed among the groups ($p = 0.44$), indicating a limited impact of the encoder backbone on segmentation performance once FMs are sufficiently pretrained. Notably, the CNN subgroup exhibited the lowest heterogeneity owing to the architectural homogeneity within the subgroup.

**Imaging Modality:** To evaluate the variation in segmentation performance across imaging modalities, we categorized the studies into three dominant subgroups based on their physical acquisition characteristics: CT/X-ray (k = 17), which utilizes ionizing radiation; MRI/US (k = 18), which are non-ionizing; and pathology/endoscopy (k = 9), which rely on optical imaging. Other modalities were excluded because of insufficient study counts to establish statistically meaningful subgroups. Among these subgroups, pathology- and endoscopy-based VFMs achieved the highest pooled Dice of 0.85 ($I^2$

= 80.0%), followed by MRI/US (Dice = 0.84; $I^2$ = 91.6%) and CT/X-ray (Dice = 0.77; $I^2$ = 95.6%) [Fig. 9(c)]. The intergroup difference was statistically significant (p = 0.027), demonstrating that imaging modality is a significant source of segmentation performance heterogeneity.

**Model Scope:** To analyze the generalizability of VFMs, we stratified the studies into unimodal (k = 47) and multimodal (k = 47) categories based on whether the FMs were specifically developed for a single imaging modality or designed for diverse cross-modality applications. The unimodal VFM studies achieved a pooled Dice score of 0.81 ($I^2$ = 96.9%), and multimodal studies yielded an identical pooled Dice score of 0.81 ($I^2$ = 92.0%) [Fig. 9(d)]. However, the intergroup difference was not statistically significant (p = 0.977), suggesting that multimodal VFMs achieve robust segmentation performance comparable to unimodal VFMs.

Among the subgroup analyses performed, input dimensionality (p = 0.030) and imaging modality (p = 0.027) emerged as statistically significant sources of between-group heterogeneity, whereas encoder architecture (p = 0.44) and model scope (p = 0.977) did not demonstrate significant intergroup differences. These findings suggest that the main factors influencing performance variability in VFM segmentation are related to the complexity and dimensionality of the input data rather than architectural choices or model scope. Furthermore, the comparable performance between unimodal and multimodal VFMs indicates that current VFMs demonstrate robust cross-domain transferability and generalizability across diverse clinical imaging settings.

## 5. VISION-LANGUAGE FOUNDATION MODELS

VLFMs extend vision-only approaches by learning multimodal representations from paired image–text data, facilitating reasoning across various tasks, including classification, VQA, report generation, segmentation, and image synthesis. This section categorizes VLFM studies based on their pretraining paradigms into two distinct groups: contrastive learning-based VLFMs and generative VLFMs. The contrastive VLFMs are further stratified according to their alignment strategy into multiscale contrastive alignment, knowledge-enhanced adaptations, CLIP-based generalist, CLIP-based domain-specific frameworks, CLIP-based segmentation frameworks, and interpretable VLFMs. Conversely, generative VLFMs are classified based on their generalization scope into generalist, cross-modal, and domain-specific VLFMs. A detailed overview of all the VLFM studies used in qualitative synthesis is provided in Supplementary Table 2. Following the qualitative synthesis, a quantitative analysis of the VLFMs is conducted to assess overall performance and trends in dataset utilization.

### 5.1 Contrastive Learning-Based VLFMs

In medical imaging, contrastive learning has emerged as a dominant paradigm for vision–language pretraining, where the primary objective is to align image and text embeddings into a shared latent space. While generalized frameworks like CLIP [15] have demonstrated superior generalization capability, their direct application to medical imaging presents unique challenges, including susceptibility to false negatives from similar semantic features, and domain gaps between natural and medical image–text pairs. To address these challenges, contrastive VLFMs in medical imaging have evolved along several complementary directions, including multiscale alignment and knowledge-enhanced adaptations to CLIP-based generalist and domain-specific frameworks. Accordingly, this section organizes contrastive learning VLFMs based on their alignment strategy and application scope.

#### 5.1.1 Multiscale Contrastive Alignment

A major limitation in utilizing contrastive pretraining for medical imaging is the need to align highly localized visual cues with region-specific textual descriptions, rather than relying on global semantic matching. Because pathologies often represent only a small fraction of the image amidst a large normal background, operations such as global pooling tend to dilute critical local features. Recent frameworks have adopted multiscale contrastive alignment to explicitly align features at varying levels of granularity.

GLoRIA [254] exemplifies this by using cross-modal attention learning to weigh image subregions. It optimizes both global and local contrastive losses by calculating the similarity between localized image patches and individual words in the medical report, ensuring the mapping of clinical terminologies to the relevant anatomical regions. By explicitly modeling these fine-grained semantic associations, GLoRIA achieves strong downstream performance in image-text retrieval and zero-shot classification. Similarly, PRIOR [255] computes cross-modal attention between image subregions and sentence-level textual embeddings to capture fine-grained representations through a dedicated local alignment module. To further enforce the learning of low-level visual details, PRIOR combines local alignment with a cross-modal conditional reconstruction task, forcing the model to reconstruct masked images.

Extending multiscale contrastive alignment, AFLoc [256] introduces a hierarchical, multilevel semantic alignment framework to explicitly align shallow local visual features with word-level text features, deep local visual features with sentence-level text features, and global image features with report-level text features. This approach enables AFLoc to accurately identify regions of interest without manual annotations and generalize across disease categories, outperforming frameworks such as GLoRIA [256]. LIMITR [257] addresses the inherent localization ambiguity in medical imaging by introducing a local alignment aggregation strategy. As diagnostic saliency is nonuniform across imaging regions, LIMITR employs self-attention to learn the significance of local features across both frontal and lateral views. These alignments are then aggregated into a weighted image-report similarity score, mimicking the radiologist's workflow of focusing on the relevant views where the pathology is most apparent. Extending multiscale alignment to the temporal domain, BioViL-T [83] demonstrates that localized self-supervision can be integrated with other complex objectives, such as longitudinal modeling. In particular, BioViL-T complements global alignment and image-guided MLM using a local contrastive loss to compute weighted pairwise cosine similarities between text tokens and image patch embeddings. This fine-grained visual grounding enables BioViL-T to achieve multiscale performance on temporal classification and factually accurate report generation benchmarks.

### 5.1.2 Knowledge-Enhanced Contrastive Adaptations

Although multiscale feature alignment significantly improves fine-grained feature matching, conventional contrastive pretraining frameworks are susceptible to false negatives, where semantically similar samples are treated as negatives and pushed apart in the latent space. Furthermore, relying solely on unstructured reports obscures critical cues due to grammatical complexity, overlapping vocabulary, and an imbalance between positive findings and normal observations. To overcome these challenges, knowledge-infused frameworks explicitly incorporate external medical knowledge, such as clinical priors, standardized terminologies, and detailed pathological descriptions, into the alignment process.

The prevalence of false negatives in contrastive pretraining has led to the development of several frameworks that reconstruct contrastive objectives using semantic similarities instead of binary instance discrimination. For example, MedCLIP [80] decouples image-text pairs and leverages the Unified Medical Language System (UMLS) entities to build a semantic similarity matrix for generating soft targets. Specifically, it replaces the conventional InfoNCE loss with a knowledge-infused semantic matching loss to reduce false negatives improving robustness in retrieval and zero-shot classification. Similarly, leveraging UMLS expert knowledge, MLIP [258] incorporates a UMLS-guided category-level contrastive learning approach, using prototype-based clustering (Sinkhorn-Knopp clustering) to cluster semantically similar samples for differentiating false negatives from true negatives. IMITATE [259] exploits the natural hierarchical structure of radiology reports by separating descriptive findings from clinical impressions as clinical priors and implements a clinical informed contrastive loss that treats clinically correlated patients as soft positives, outperforming GLoRIA.

Rather than aligning entire sentences or reports, several studies have extracted structured clinical concepts to guide cross-modal alignment. MedKLIP [260] parses raw reports into a structured format containing the entity, position, existence, and translates them into detailed attribute descriptions using external knowledge bases. These descriptions are then used as queries by the transformer-based fusion module to attend to specific image regions, significantly reducing semantic ambiguity and improving zero-shot classification and localization tasks. Complementing global alignment, ASIMSA [27] reduces the discrimination of semantically similar samples by integrating a disease-entity alignment module. By utilizing report filtering to extract medical observations, ASIMSA provides explicit entity-level supervised signals enhancing downstream tasks, such as disease classification, report generation, and cross-modal retrieval.

To achieve precise comprehension of anatomy and pathology, external definitions were utilized with structural priors. For instance, DeViDe [261] augments standard medical definitions by curating radiographic feature descriptions from Radiopaedia and generating additional visual descriptors with LLMs. Based on the align-before-fuse (ALBEF)-style architecture [262], DeViDe aligns the visual descriptors with localized image patches using local patch-text alignment loss, supplemented by a negative contrastive loss to distinguish closely analogous but non-matching conditions. GK-MVLP [263] similarly employs a grounded knowledge-enhanced module to explicitly align abnormalities with anatomical structures, achieving superior performance over typical contrastive approaches, such as GLoRIA. Furthermore, ECAMP [264] mitigates the complexity and imbalance of clinical language by using LLMs like ChatGPT, to distill entity-centered context from complex reports. The distilled knowledge is then used as a guide to the entity-centered MLM task, where the descriptive words preceding an entity are explicitly masked, and a re-balancing factor is applied to minority-positive disease descriptors to counter class imbalance in clinical language. SAT [265] extends knowledge enhancement to universal 3D medical image segmentation by constructing a large multimodal knowledge tree comprising over 6,500 anatomical terminologies derived from UMLS and web sources. By injecting this structured domain knowledge into a text encoder via contrastive learning, SAT explicitly aligns visual anatomical features with their textual definitions in the latent space, achieving strong segmentation performance across 72 distinct datasets without requiring specialized architectures.

### 5.1.3 CLIP-Based Generalist VLFMs

Generalist CLIP-based [15] VLFMs aim to be task-agnostic by learning vision–language representations that can transfer across diverse datasets, imaging modalities, and languages. Unlike the contrastive approaches discussed in Sections 5.1.1 and 5.1.2, generalist CLIP-based frameworks predominantly build upon the foundational global contrastive paradigm of CLIP and achieve robust cross-domain generalization through large-scale pretraining, structural adaptations, or multi-model knowledge distillation.

**Large-Scale Pretraining:**

A primary driver of generalist VLFMs is the curation of large-scale image-text corpora comprising multiple modalities. BiomedCLIP [266] exemplifies this by curating PMC-15M, a large-scale dataset comprising 15M image-text pairs across 30 biomedical imaging modalities. By pretraining on PMC-15M using PubMedBERT [267] text encoder and ViT-B image encoder, BiomedCLIP enables robust performance across diverse downstream biomedical tasks by capturing fine-grained biomedical semantics. Similarly, PMC-CLIP [81] uses PMC-OA, a corpus comprising 1.6M image-text pairs extracted from PubMed Central. Utilizing a ResNet image encoder and PubMedBERT text encoder along with a dual-training strategy, PMC-CLIP surpasses M3AE and ARL on image-text retrieval and zero-shot classification [81]. To address the scarcity of naturally paired multimodal data, UniMed-CLIP [268] curates the UniMed dataset (5.3M image-text pairs) through a scalable LLM-in-the-loop framework, which converts existing high-quality uni-modal datasets into pseudo-image-text pairs by generating diverse, biomedical-style captions. RadCLIP [269] extends contrastive learning to volumetric imaging by fusing 2D slice-level features via a slice pooling adapter with a slice-wise attention mechanism, serving as a generalist framework for tasks such as disease classification, retrieval, and cross-modal report generation. MEDBind [270] introduces a tri-modal contrastive framework unifying CXR, electrocardiogram (ECG), and medical text within a shared latent space. By utilizing an edge-modality contrastive loss for robust cross-modal alignment between CXR and ECG data, MEDBind achieves state-of-the-art performance in image retrieval and zero-shot classification tasks [270].

**Structural Adaptation and Knowledge Distillation:**

To alleviate the need for extensive end-to-end pretraining, several generalist VLFMs have adopted architectural unification and distillation-based techniques. In particular, MMKD-CLIP [2] achieves generalist capabilities through multiple medical CLIP knowledge distillation, where MMKD-CLIP acts as a student model distilled from nine CLIP architectures, achieving cross-modal generalization across 26 imaging modalities. PTUnifier [271] unifies the dual-encoder and fusion-encoder architectures into a single, task-agnostic FM to handle image-only, text-only, and image-text paired tasks by substituting missing modalities using visual and textual soft prompts. Med-UniC [272] tackles the community bias prevalent in VLFMs by learning language-agnostic representations across English and Spanish clinical data. It introduces cross-lingual text alignment regularization, a negative-free optimization method that utilizes latent language disentanglement to minimize linguistic disparities, preventing bias toward a specific language.

### 5.1.4 CLIP-Based Domain-Specific VLFMs

CLIP-based domain-specific VLFMs rely on the foundational CLIP [15] paradigm to generalize across multiple downstream tasks within a specific imaging modality or clinical domain. Accordingly, this section categorizes these VLFMs according to the target imaging modality and clinical context for which they have been developed and deployed.

**Radiology:**

**CXR:** In radiology, CXR represents the dominant imaging modality, and it accounts for the majority of CLIP-based VLFMs proposed to date. CheXzero [273] directly adapts the CLIP pretraining strategy by jointly training a ViT image encoder and the OpenAI CLIP text transformer on over 377,000 image–text pairs from the MIMIC-CXR [274]dataset. By optimizing the cosine similarity between the images and clinical impressions from radiology reports, CheXzero supports zero-shot multilabel classification using a positive-negative softmax evaluation. To address dataset heterogeneity, UniChest [275] integrates a CLIP-style contrastive loss with MoE to align image-text features across multiple sources, including MIMIC-CXR [274], CheXpert [276], ChestX-ray14 [277], and VinDr-CXR [278], thereby mitigating domain bias and improving cross-dataset generalization. Conversely, M-FLAG [279] mitigates dimensional collapse in standard CLIP training by freezing the text encoder and optimizing only the vision encoder with an orthogonality loss alongside contrastive alignment, yielding more robust CXR representations. ELIXR [28] enhances contrastive pretraining via a two-stage framework. Initially, it uses a CLIP-style loss to align the CXR vision encoder with the T5 [280]text encoder using image-report pairs and then projects these visual representations into the PALM 2 [281] embedding space, achieving state-of-the-art performance on classification and semantic search, along with VQA and report quality assurance. Beyond adult imaging, NeoCLIP [282] introduces a neonatal-specific VLFM designed to interpret NICU radiographs. Leveraging contrastive pretraining on over 20,000 neonatal X-rays paired with clinical reports and fine-tuning on LLM-extracted structured labels, NeoCLIP achieves strong pathology detection performance across 15 diseases and five medical devices.

**Mammography:** In mammography, Mammo-CLIP [283] represents the first mammography-specific VLFM employing contrastive pretraining on mammogram-report pairs to improve efficiency and robustness. Mammo-FActOR [283] further introduces a feature attribution framework for Mammo-CLIP to explicitly map textual findings in radiology reports to the image encoder's feature channels, supporting spatial localization of abnormalities without explicit bounding box annotations.

**CT Imaging:** Extending CLIP to volumetric CT imaging, CT-CLIP [284] proposes a 3D VLFM for chest CT analysis by aligning representations between a 3D ViT encoder and a text encoder using contrastive loss. Trained on CT-RATE, CT-CLIP outperformed the fully supervised baseline CT-Net [285] on 3D abnormality detection benchmarks. However, to prevent the dilution of sparse clinical details inherent in global 3D alignment, CT-GLIP [286] adopts fine-grained contrastive learning to explicitly align organ-level visual embeddings and targeted anatomical and diagnostic text descriptions. TotalFM [287] similarly prioritizes computational efficiency through an organ-separated framework that extracts organ-specific 3D patches to perform self-supervised pretraining with VideoMAE, followed by InfoNCE contrastive alignment to reduce background computation. For cardiac CT imaging, Cardiac-CLIP [288] introduces a two-stage pipeline combining a 3D MAE with contrastive vision-text alignment to capture complex cardiovascular features. By parsing free-text reports into structured pathology vectors to construct a semantic affinity matrix, Cardiac-CLIP achieves fine-grained contrastive supervision through soft labels. This approach enables Cardiac-CLIP to achieve state-of-the-art performance in real-world clinical tasks, such as the prediction of acute coronary syndrome, diagnosis of functional coronary stenosis, and coronary artery calcium grading [288]. Advancing CT vision-language pretraining to abdominal imaging, Merlin [289] integrates a 3D ResNet152 image encoder with a Clinical Longformer text encoder, trained on over 15,000 paired abdominal CT scans, electronic health record diagnosis codes, and radiology reports utilizing multitask contrastive and phenotype classification pretraining. Merlin supports zero-shot classification, cross-modal retrieval, report generation, and 3D segmentation, demonstrating high generalizability across multiple institutions.

**MRI:** In MRI, Decipher-MR [290] proposes a general-purpose 3D MRI VLFM by using a two-stage pretraining pipeline: independent self-supervised learning of vision and text encoders, followed by joint multimodal contrastive pretraining utilizing a unique organ-based, study-level sampling strategy. This approach enables Decipher-MR to learn fine-grained differences within the same anatomical region, supporting modular designs where lightweight, task-specific decoders can be flexibly attached atop a frozen pretrained encoder for diverse downstream tasks.

**Ultrasound and Echocardiography:** Adapting VLFMs to US relies on aligning temporal video frames with clinical text. For instance, EchoCLIP [291] combines a ConvNeXt [292] image encoder with a decoder-only text transformer to analyze ECG videos. By employing a custom domain-specific tokenizer to compress repetitive reports, EchoCLIP achieves robust zero-shot cardiovascular prediction. To overcome the scarcity of paired image-text pairs in obstetric imaging, FetalCLIP [293] utilizes GPT-4o [294] to generate multiple pseudo-caption templates from clinician labels, gestational age, and pixel spacing. Pretrained using the CLIP framework, FetalCLIP yields image embeddings for various downstream tasks, including zero-shot fetal view classification, gestational age estimation, and congenital heart disease detection.

**Ophthalmology:**

In ophthalmology, domain-specific CLIP-based VLFMs are developed primarily for retinal image analysis and ocular disease screening by leveraging knowledge infusion and domain-specific datasets. Notably, FLAIR [295] infuses ophthalmology expert knowledge into the pretraining by converting categorical disease labels into rich descriptive text prompts. Leveraging CLIP-style dual-encoder architecture (ResNet-50 vision encoder and a BioClinicalBERT text encoder) FLAIR optimizes a category-aware contrastive loss to align categorical labels with textual descriptions, improving robust generalization across diverse retinal disease recognition tasks. Utilizing real-world clinical data, RET-CLIP [296] learns from over 193,000 binocular image-text triplets, including diagnostic reports in Chinese. By adopting a tripartite contrastive training strategy to jointly align representations from the left and right eyes, along with patient-level semantics, RET-CLIP achieves state-of-the-art performance against strong baselines (RETFound, FLAIR, and PMC-CLIP) across diverse tasks, such as diabetic retinopathy and glaucoma diagnosis RET-CLIP.

Broadening the diversity of ophthalmic modalities, UrFound [297] introduces a retinal VLFM to process both CFP and OCT images using a single ophthalmic modality-agnostic encoder. Employing a knowledge-guided masked modeling strategy to simultaneously predict masked image patches and masked word tokens from clinical descriptions, UrFound facilitates robust and comprehensive retinal disease analysis. Addressing the issue of limited interpretability and lack of user interaction in standard VLFMs, VisionUnite [298] incorporates a fine-tuned Llama -7B and a specialized vision adapter in the transformer-based vision encoder to explicitly classify distinct clinical signs to support open-ended multi-disease diagnosis and multi-turn clinical dialog. This approach mimics the step-by-step diagnostic reasoning employed by human medical professionals, achieving diagnostic accuracy and relevance comparable to ophthalmologists, while outperforming GPT-4V [299]and Gemini Pro [13].

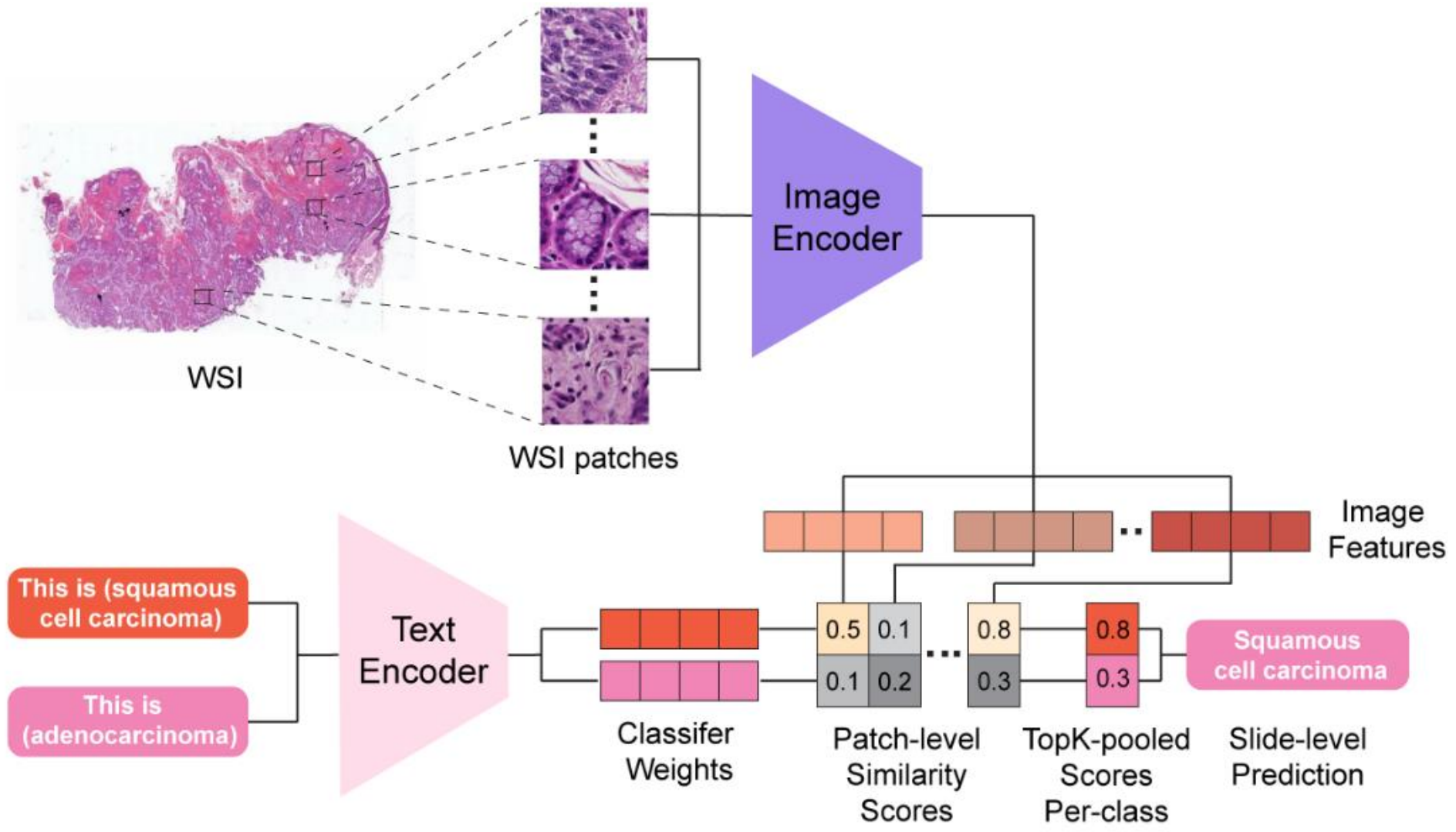

**Fig. 10.** Architecture of MI-Zero [300] illustrating the conversion of a WSI into patch-level embeddings and their aggregation using top-K max pooling to produce WSI-level classification predictions.

**Pathology:**

In digital pathology, the development of VLFMs is constrained by the scarcity of high-quality image-text pairs and the computational challenges associated with gigapixel WSI processing. For example, PLIP [301] overcomes data scarcity by fine-tuning CLIP on OpenPath, a curated dataset of over 208,000 image-text pairs from medical Twitter and the LAION dataset [302], enabling zero-shot classification and cross-modal retrieval without domain-specific fine-tuning. Similarly, IP-CLIP [303] utilizes the IPATH dataset, comprising over 45,000 image-text pairs sourced from Instagram. By using LLMs to extract textual descriptions and clean noisy captions, IP-CLIP demonstrates that social media can provide a robust signal for adapting general VLFMs to pathology tasks. CPLIP [304] eliminates the need for strictly paired image-text datasets by replacing traditional one-to-one image-text pairing with an unsupervised many-to-many alignment. CPLIP aligns GPT-3-generated text "bags" with PLIP-retrieved visual "bags" using a multiple-instance contrastive loss, achieving superior zero-shot performance without ground-truth annotations [304].

Exploiting hierarchical WSI processing in VFMs. MI-Zero [300] adapts zero-shot transfer by introducing a multiple-instance learning framework. As illustrated in Fig. 10, MI-Zero divides a massive WSI into smaller patches, computing the cosine similarity between each patch embedding and a text prompt embedding, which are then aggregated using top K pooling to generate slide-level prediction. Furthermore, MI-Zero leverages HistPathGPT to handle specialized medical vocabulary, achieving zero-shot accuracy of 70.2% across three different real-world cancer subtyping tasks [300].

### 5.1.5 CLIP-Based Segmentation VLFMs

Distinct from Section 5.1.4, which organizes CLIP-based frameworks according to their clinical domain, this section organizes CLIP based frameworks around a single shared downstream task, segmentation. Rather than being defined by their clinical domain, these CLIP-based segmentation VLFMs repurpose CLIP's image-text alignment capabilities for dense pixel or voxel-level prediction.

**Radiology:**

One of the earliest adaptations of CLIP-style contrastive pretraining for medical image segmentation is the CLIP-Driven Universal Model [305], which addresses dataset label inconsistencies by replacing conventional static one-hot labels with text embeddings generated from a pretrained CLIP text encoder. By mapping related anatomical structures closely together within the embedding space, the universal model successfully captures the intrinsic semantic relationships between 25 distinct organs and six tumor types, achieving state-of-the-art performance on the MSD [249] and BTCV [114] benchmarks.

Complementing SAM, CLIP, and its variants has also been used to derive spatial prompts from language-based interactions. For instance, CLISC [306] utilizes CLIP to generate enhanced CAMs, which are automatically converted into bounding box and point prompts to guide SAM for unsupervised brain tumor segmentation. Trained on 3M synthetic images, FluoroSAM [307] introduces a text-prompted SAM variant for X-ray imaging by combining a frozen CLIP encoder with a novel vector quantization module, enabling robust zero-shot segmentation of anatomical structures and surgical tools. Similarly, MedCLIP-SAM [308] and MedCLIP-SAMv2 [309] integrate BiomedCLIP with SAM to support text-prompted zero-shot segmentation across modalities like US, MRI, and X-ray. Extending CLIP-based segmentation to volumetric CT imaging, OpenVocabCT [310] aligns 3D CT images with radiology reports via a multigranularity contrastive objective. As standard reports lack granular organ-level details, OpenVocabCT utilizes a large LLM to decompose radiology reports into concise organ captions, improving the segmentation of merged sub-organs and complex structures. VOILA [311] addresses class imbalance and computational overhead by adopting a voxel-centric approach to align 3D voxel representations with pretrained CLIP text tokens using cosine similarity and a complexity-aware sampling module to focus on hard-to-segment regions, improving generalization while reducing computational costs.

**Pathology:**

In computational pathology, language-based prompting is more advantageous than spatial prompting, as discussed in Section 4.1.5. CRNS-Net [312] improves referring segmentation by leveraging CLIP embeddings to reduce feature competition between closely adjacent nuclei and their boundaries. In particular, CRNS-Net introduces a class guidance block to compute the cosine similarity between image features and CLIP text embeddings, explicitly guiding the model to disentangle and aggregate features specific to the nuclei, edges, and background. Emphasizing generalization and interaction, VISTA-PATH [313] introduces a class-aware interactive VLFM that supports segmentation across 93 tissue classes and nine organs. Initialized with PLIP [301], VISTA-PATH supports dynamic human-in-the-loop active learning, allowing pathologists to provide sparse patch-level corrections to refine whole slide, pixel-level segmentations.

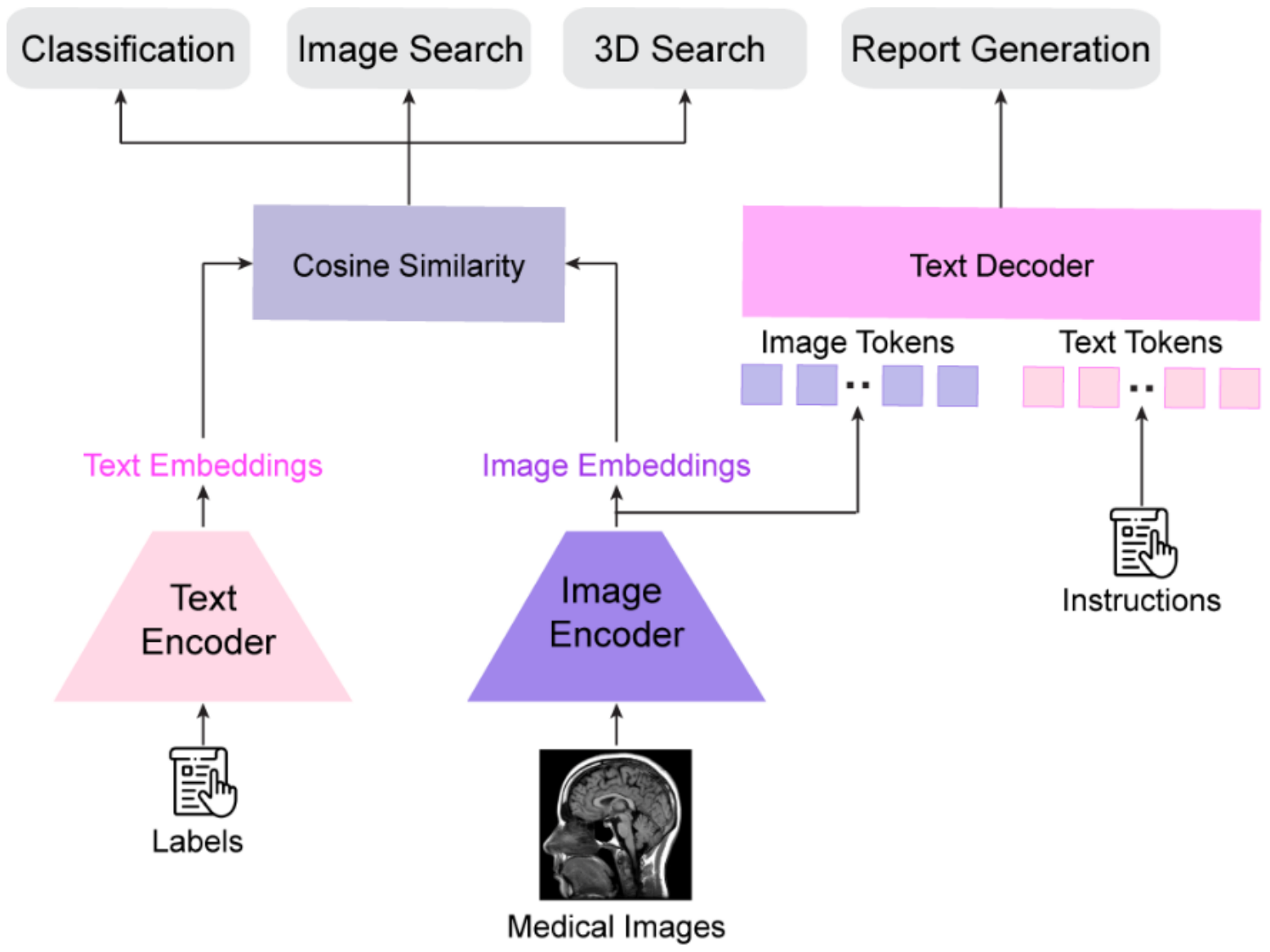

**Fig. 11.** Overview of the MedImageInsight [314] architecture.

### 5.1.6 Interpretable Contrastive VLFMs

A persistent barrier to the clinical adoption of FMs is their limited interpretability, as their predictions are often difficult to justify due to the opaque nature of these models. To address this challenge, recent contrastive VLFMs have introduced interpretable frameworks, whose predictions can be explicitly traced back to clinically meaningful concepts. CLEAR [315] exemplifies this paradigm by computing similarity scores between CXR representations and LLM-extracted clinical observations. These similarity scores are subsequently used to weigh the pre-computed LLM embeddings for each observation, ensuring that the predictions are directly traceable to specific radiological findings, fostering trustworthiness. Similarly, MONET [316] employs contrastive learning to automatically score dermatology images against user-defined clinical concepts without relying on dense manual annotations. The resulting concept scores then serve as inputs to interpretable concept bottleneck models (CBMs), enabling physicians to decipher the specific clinical factors influencing the FM's decisions. Extending this interpretability across diverse medical imaging modalities, ConceptCLIP [317] advances standard concept scoring frameworks by integrating global features and fine-grained local region alignments to compute similarity scores between input images and specific medical concepts derived from the UMLS. These concept-level similarities form a transparent bottleneck layer that can be directly mapped to disease predictions through a linear classifier, significantly enhancing the transparency and accountability of automated diagnostic reasoning. While CBMs provide transparency through text-based clinical concepts, MedImageInsight [314] achieves interpretability through direct visual evidence [Fig. 11]. Specifically, MedImageInsight computes the k-nearest neighbor image-to-image search on contrastively aligned representations and retrieves the most visually and pathologically similar historical cases from a training database to provide clinicians with direct evidence, demonstrating transparent decision-making while also preserving the model's ability to generate reliable confidence scores.

### 5.2 Generative VLFMs

Generative VLFMs are FMs that use diverse generative learning frameworks, such as MRM, autoregressive decoding, diffusion-based synthesis, generative adversarial networks, and variational autoencoders, to combine multimodal clinical data, including medical images, clinical reports, and clinical knowledge. Through large-scale multimodal pretraining, these models learn context-aware representations enabling free-form reasoning and adaptation to diverse downstream tasks, including report generation, disease interpretation, cross-modal image synthesis, segmentation, and clinical decision-making. Accordingly, this section categorizes generative VLFMs into three broad categories: generalist multitask frameworks using a unified framework to handle multiple downstream tasks. cross-modal synthesis approaches focusing on generating high-fidelity medical images, and domain-specific frameworks tailored to specific imaging domain or modality.

#### 5.2.1 Generalist Generative VLFMs

In recent years, generative VLFMs have evolved from processing simple image-text pairs into multitask generalists capable of supporting heterogeneous clinical workflows utilizing a single unified backbone. Rather than adapting general-purpose models for medical text generation, these generalist VLFMs are designed to handle diverse, high-dimensional clinical inputs (2D images, 3D volumes, and surgical videos) and generate versatile outputs, including text reports, bounding boxes, and segmentation masks. However, this paradigm is constrained by the scarcity of paired medical datasets, high computational overhead, cross-task interference, and hallucination risks.

**Data-Efficient Generalists:**

To overcome the scarcity of large-scale paired datasets in medical imaging, recent studies have leveraged advanced data synthesis, distinct training frameworks, and ICL. For instance, LLaVA-Med [25] addresses data scarcity by using GPT-4 [318] to generate open-ended conversational instruction data from PMC-15M image-text pairs. Employing a two-stage training strategy that aligns biomedical visual concepts and text vocabulary, followed by end-to-end instruction tuning, LLaVA-Med outperforms BiomedCLIP in biomedical VQA [25]. Alternatively, Med-Flamingo [29] reduces the need for downstream fine-tuning datasets by adopting multimodal ICL. By pretraining on interleaved medical image-text data curated from PubMed and medical textbooks, Med-Flamingo supports multimodal text generation and visual-language reasoning without task-specific fine-tuning. Lingshu [319] infuses extensive medical knowledge through a rigorous four-stage training paradigm: shallow alignment using coarsely annotated image-text pairs with a frozen language model; deep alignment using structurally complex multimodal data; comprehensive medical instruction tuning; and reinforcement learning with verifiable rewards. Furthermore, Lingshu tackles data scarcity by efficiently acquiring and synthesizing medical knowledge from texts, images, and general-domain data.

**Interference-Mitigating Generalists:**

Although multitask and multimodal learning are the defining characteristics of generalist VLFMs, training a single unified framework across diverse medical datasets often leads to cross-task interference, commonly referred to as the "tug-of-war" phenomenon. To mitigate this cross-task interference, several generalist VLFMs have incorporated adaptive routing to channel task-specific information within a shared backbone. For example, MedVersa [320] uses an LLM as an optimizable orchestrator to dynamically select the appropriate modules for visual-centric tasks, such as segmentation and object detection. Furthermore, it applies domain-aware minibatch gradient descent to balance learning and develop shared representations to support task transfer and reduce interference. Similarly, Uni-Med [321] alleviates the multitask tug-of-war problem by introducing a connector-MoE module to adaptively select projection experts, enabling a single VLFM to generalize across diverse tasks, such as VQA, report generation, and image classification, without task-specific architectures. Alternatively, HealthGPT [322] decouples the learning processes to store heterogeneous comprehension and generation knowledge in independent plugins by adapting heterogeneous LoRA and hierarchical visual perception for robust multimodal alignment and instruction tuning.

**Volumetric Generalists:**

As medical imaging is inherently volumetric, generalist VLFMs have been extended to volumetric imaging by incorporating specialized encoding techniques, dimensionality compression, and adaptive spatial fusion, while preserving computational efficiency. RadFM [73] unifies 2D and 3D data by padding 2D slices into a pseudo-3D format for processing through a 3D ViT. Utilizing a perceiver module to aggregate and compress volumetric features into a fixed number of tokens, RadFM reduces the computational load enabling seamless integration with textual embeddings outperforming GPT-4V [299] in VQA and automated report generation [73] Similarly, BiomedParse-V [323] compresses 3D data to a 2.5D RGB representation by implementing fractal volumetric encoding to downsample context slices based on the distance from the central focal slice. This approach demonstrates that a 2D FM can be translated into 3D segmentation tasks through compression. Med3DVLM [324] integrates a DCFormer to decompose 3D convolutions into three parallel convolutions across the three axes: depth, height, and width. These convolutions are then paired with a dual-stream MLP-Mixer projector to fuse both low-level spatial features and high-level spatial features with text embeddings, supporting image retrieval, report generation, and VQA within a single framework. In contrast, Med-2E3 [325] mimics the clinical workflow of radiologists by integrating a dual 3D-2D encoder architecture with a text-guided inter-slice scoring module to adaptively attend to different 2D slices based on their visual content and textual task instructions. OmniV-Med [326] processes 2D images, 3D volumes, and temporal videos within a single unified architecture. To handle volumetric and temporal data, it employs a rotary position-adaptive encoder that extrapolates 2D spatial positional encodings across the depth and temporal dimensions, supporting coherent spatial reasoning without an architectural redesign. Furthermore, to reduce the computational overhead, OmniV-Med introduces a medical-aware token pruning mechanism to calculate the latent distance between adjacent slices, effectively discarding ~ 60% of redundant visual tokens without degrading performance.

A persistent limitation in 3D VLFMs is their reliance on inter-slice interpolation to handle volumes of varying depths. However, interpolation introduces artifacts arising from redundant information across adjacent slices. VividMed [327] addresses this by dynamically adjusting the weights in its patch embedding layer based on the number of slices in the input volume, preserving the native spatial resolution without constraining the volume to a fixed, uniform size.

**Interpretable Generalists:**

As discussed in Section 5.1.6, clinical interpretability is intrinsic to reduce hallucinations and build trustworthy VLFMs. UniBiomed [328] exemplifies this paradigm by integrating SAM with a multimodal LLM to simultaneously generate diagnostic findings and segmentation masks. Specifically, UniBiomed utilize the LLM to generate text prompts to condition SAM's mask decoder, enabling end-to-end interpretation by directly correlating LLM-generated text and visual evidence. Similarly, MIMO [329] maps visual referring inputs to pixel grounding outputs by introducing special tokens into the LLM vocabulary. It explicitly wraps the medical entities in the generated text with <p> and </p> bracket tokens followed by a <SEG> token to trigger the mask decoder, ensuring an auditable alignment between the text response and medical entities. Analogous to MIMO token-based grounding, RadDiagSeg-M [330] addresses complex, hierarchical clinical workflows by jointly detecting abnormalities, generating textual diagnoses, and flexible multi-target segmentation masks through <seg> tokens embedded directly within its generated responses.

In contrast to the grounding approach, which uses specialized tokens to trigger pixel-level decoding, holistic VLFMs, such as BiomedParse [331], frame visual reasoning as a unified image parsing task. By utilizing text prompts for joint segmentation, detection, and recognition tasks, BiomedParse improves clinical interpretability by simultaneously segmenting and labeling all relevant objects in an image using an established biomedical object ontology, eliminating the need for manual, user-specified bounding boxes. Conversely, MedDAM [332] targets both interpretability and data scarcity by introducing a framework designed for region-specific captioning through the fusion of focal crops with global image features via gated cross-attention. By integrating localized captioning with flexible region-of-interest detection, MedDAM achieves superior performance in report generation, effectively reducing hallucinations in comparison with LLMs such as GPT-4o and Claude 3.7 Sonnet [332].

### 5.2.2 Generative VLFMs for Cross-Modal Synthesis

In contrast to unidirectional text-to-image generation, Generative VLFMs have also been used to synthesize high-fidelity medical images, sequences, and paired segmentation masks across diverse modalities by leveraging text prompts, clinical reports, and structured metadata.

**Text-Guided Image Synthesis:**

Among the text-guided synthesis, MINIM [91] represents a unified generative VLFM based on latent stable diffusion, capable of synthesizing images across OCT, FP, CXR, CT, and MRI modalities. By using a U-Net enhanced with cross-attention to align text encoded from a BERT tokenizer, MINIM applies a two-stage RLHF to iteratively refine the quality of the generated images, alleviating the need for real-world medical datasets. TUMSyn [333] tackles brain MRI synthesis guided by text prompts generated from demographic data and imaging parameters. TUMSyn adopts a two-stage training strategy in which it first leverages contrastive pretraining to align image and text semantics and then performs cross-sequence synthesis using a ResNet-34 image encoder and a local implicit image function-based decoder to generate target MR images at the desired resolution. UniSyn [334] extends TUMSyn to MRI, CT, and PET. Specifically, UniSyn replaces TUMSyn's cross-attention with a dual-parameter arithmetic operation to map NeuroCLIP-derived text features weighted with scaling ($\alpha$) and bias ($\beta$) terms, thereby adjusting modality-specific contrast differences and intensity shifts.

**Any-to-Any Synthesis:**

In contrast to text-based image synthesis, XGeM [335] introduces an "any-to-any" multimodal generation paradigm to support flexible synthesis between modalities. For example, XGeM can generate CXR directly from clinical reports, translate between X-ray projections, and synthesize missing projections along with diagnostic reports from another X-ray projection. To support multi-output generation, XGeM couples three different diffusion models linked by projection encoders and cross-attention layers along with a multi-prompt training strategy to support diverse input-output conditioning patterns.

**Paired Image-Mask Synthesis:**

To reduce reliance on expert annotations in CAD, generative VLFMs have also been utilized to generate complex paired datasets. MedSegFactory [336] introduces a versatile framework to simultaneously generate paired medical images and corresponding segmentation maps from concise user-defined text prompts specifying the target label, imaging modality, and anatomical region. Using a dual-stream diffusion model integrated with joint cross-attention, MedSegFactory iteratively refines and enforces robust semantic and spatial alignment, enhancing the fidelity of synthesized pairs.

### 5.2.3 Domain-Specific Generative VLFMs

Domain-specific generative VLFMs are tailored to address modality-specific challenges. In contrast to generalist frameworks that emphasize broad adaptability, these models incorporate specialized datasets, knowledge-guided training strategies, and task-aware architectures to capture fine-grained features and the clinical context. This section categorizes generative VLFMs based on the medical domains for which they have been developed and applied.

**Radiology:**

**CXR:** In CXR, the development of generative VLFMs has been focused on automating CXR report generation and interactive diagnostics. For instance, CXR-LLaVA [337] extends the LLaVA framework to CXR interpretation by integrating a ViT-L vision encoder with the Llama-2 [338] text encoder. This approach enables clinicians to interact with CXR images through natural language queries and receive dialog-driven diagnostic reasoning. Similarly, CheXagent [339] is an instruction-tuned FM designed to analyze and summarize CXRs. Trained on CheXinstruct, a large-scale dataset of 6M instruction-image-answer triplets, CheXagent achieves superior performance across eight CXR interpretation tasks on the CheXbench benchmark [339]. In contrast, MAIRA [340] demonstrates that incorporating lateral imaging views alongwith frontal radiographs can significantly improve clinical and lexical output metrics by using a pretrained image encoder and fine-tuning only the LLM and adapter.

Emphasizing dialog-driven interactions between clinicians and AI systems, XrayGPT [341] proposes a conversational VLFM for CXR analysis. Specifically, XrayGPT instruction-tunes a frozen MedCLIP visual encoder with Vicuna LLM on 217,000 curated interactive summaries derived from free-text radiology reports to support open-ended clinical inquiries. Advancing this paradigm, RaDialog [342] introduces a dual-branch architecture supporting iterative refinement through dialog-based interaction. Incorporating a secondary structured findings extractor alongside visual feature extraction, RaDialog captures core clinical findings, improving the FMs clinical correctness score by 13.3%.

To improve the trustworthiness of CXR report generation, ChestX-Reasoner [93] simulates the systematic workflow of human radiologists through a two-stage framework combining supervised fine-tuning and RL to improve interpretability by generating intermediate reasoning steps supervised by reasoning processes mined directly from routine clinical reports. EditGRPO [92] mitigates shortcut bias and hallucinations encountered in RL by applying GRPO [92] to inject sentence-level expert corrections during training to optimize the clinical efficacy rewards enhancing out-of-domain generalization.

Diverging from the monolithic end-to-end report generation paradigm, MedRAX [343] proposes a dynamic medical reasoning agent. Operating via a reasoning and acting loop, MedRAX orchestrates specialized external agents for VQA, segmentation, classification, and grounding to sequentially execute complex, multi-step diagnostic traces without additional model training.

**CT:** In CT imaging, M3FM [295] proposes a specialty-oriented VLFM framework for low-dose CT lung cancer screening. By employing multiscale ViT to process volumetric CT patches alongside clinical data within a multimodal question-answering framework, M3FM demonstrates significant improvements in predicting multi-year lung cancer risks and cardiovascular disease mortality. In contrast, MedVL-SAM2 [344] integrates a 3D LLaVA-style vision-language backbone with a SAM2-based segmentation module to maintain cross-slice consistency throughout the 3D volume. This approach enables MedVL-SAM2 to concurrently support interactive VQA and report generation.

**MRI and US:** OmniMRI [345] introduces a unified VLFM to manage the entire MRI workflow by training on a corpus of over 220, 000 MRI volumes through an autoregressive transformer backbone equipped with a MoE feedforward network. Adopting a dual encoder architecture comprising an image decoder that handles pixel-level tasks and a text encoder for semantic tasks, OmniMRI can perform under-sampled image reconstruction, volumetric segmentation, abnormality detection, and report generation without requiring task-specific architectural modifications. In US imaging, LLaVA-Ultra [346] introduces a vision-language assistant tailored for fine-grained multimodal conversations leveraging SAM alongside CLIP. By incorporating an adaptive sampling module to automatically screen and filter out redundant frames, LLaVA-Ultra excels in VQA compared to LLaVA-Med [346].

**Ophthalmology:**

In ophthalmology, knowledge infusion and domain-specific datasets are leveraged to enhance VLFMs for retinal image analysis and ocular disease screening. Berger et al. [347] extend the FLAIR framework for ocular disease screening by incorporating multimodal contextual information such as clinical data, diabetic health conditions alongside the retinal images to improve predictive accuracy. Trained on the OPHDIAT dataset, comprising over 700,000 FPs, a context-aware VLFM achieves accurate and generalizable screening for diabetic retinopathy. OphGLM [348] introduces an interactive ophthalmology VLFM by integrating an image encoder, a text encoder, a fusion module, and an LLM to support professional visual dialog interactions. To inject domain-specific expertise, OphGLM employs a dynamic label pairing strategy informed by a specialized dataset comprising 100,000 instances of ophthalmic instructions and real-world doctor–patient conversations. Using a dynamic label pairing strategy OphGLM bridges visual features with natural language to improve fine-tuning on Fundus Tuning-CN, enabling accurate classification of common fundus diseases.

**Pathology:**

Recent adaptations of generative VLFMs in pathology have focused on bridging the gap between visual perception and complex clinical reasoning. Notably, CPath-Omni [349] proposes a unified VLFM for the joint analysis of both microscopic patches and gigapixel WSIs. Integrating an in-house DINOv2-based CLIP visual processor with the Qwen2.5-14B [350] LLM, CPath-Omni establishes the first 15B FM capable of consolidating diverse analytical tasks, including classification, VQA, captioning, and visual referring prompting [349]. Similarly, PathChat [30] is an AI copilot developed for pathology by adapting the UNI vision encoder with the Llama-2 [338] LLM. Fine-tuned on over 456,000 visual-language instructions, PathChat supports multi-turn conversations, multiple-choice diagnostic queries, and open-ended questions based on diverse tissues, outperforming VLFMs such as LLaVA-Med and GPT-4V [30].

To overcome the hallucinations in diagnostic reasoning, recent VLFM frameworks have adopted RL along with structured reasoning protocols. For example, PathReasoner-R1 [94] uses a knowledge-guided generation pipeline that leverages medical knowledge graphs to instill structured chain-of-thought capabilities by combining supervised fine-tuning with RL, mimicking pathologist workflows. Compared to reasoning frameworks such as Patho-R1, PathReasoner-R1 demonstrates interpretable WSI reasoning through step-by-step logic grounded on verifiable medical facts. Enhancing rigorous reasoning, TeamPath [351] introduces a multitask AI copilot with an LLM-driven router to adaptively select optimal solutions, such as RL for complex diagnostic reasoning and supervised fine-tuning for cross-modality spatial transcriptomic generation. In particular, TeamPath improves clinical reliability through a verifier-corrector pipeline to identify and revise flawed reasoning paths provided by human experts. Similarly, Patho-AgenticRAG [95] framework introduces multimodal retrieval-augmented generation system to autonomously plan multi-turn searches by utilizing an RL-optimized intelligent agent to query a specialized knowledge base of page-level embeddings from pathology textbooks. Unlike traditional text-only retrieval systems, Patho-AgenticRAG retrieves both textual and visual cues, grounding diagnostic reasoning with evidence substantiating trustworthiness.

**Oncology:**

In radiation oncology, generative VLFMs have emerged as powerful tools capable of integrating complex, multimodal data to improve cancer diagnosis and treatment planning. For instance, NasVLM [352] proposes a VLFM for nasal disease diagnosis by adopting a multigranular alignment architecture to effectively align clinical reports with unlabeled nasal endoscopic images across different anatomical regions, achieving superior nasopharyngeal carcinoma (NPC) classification and lesion localization performance compared to BiomedCLIP, MedSAM, and ViT-B. Advancing multimodal integration for NPC, the NPC model [353] combines self-supervised image pretraining with clinical text fusion utilizing a ViT backbone for MRI feature extraction and an attention-to-mask decoder for joint segmentation of gross tumor volumes and metastatic lymph nodes. To bridge the dimensional gap between dense image data and sparse clinical metadata, the NPC model leverages a Q-Former with a Vicuna-based LLM to fuse medical records, such as age, gender, and tumor staging enhancing chemotherapy sensitivity prediction. Likewise, HiCur-NPC [354] introduces a three-stage training called hierarchical feature fusion curriculum learning framework to integrate large-scale self-supervised visual pretraining with language alignment and multimodal fusion for NPC. By extracting visual features through a hybrid contrastive masked autoencoder, language representations with Llama-3-8B, and deep-fusing them through a MoE cross-attention module, HiCur-NPC achieves superior performance across key clinical tasks, including diagnosis, automated report generation, tumor segmentation, and patient prognosis, compared to FMs like BiomedCLIP and MedSAM [354].

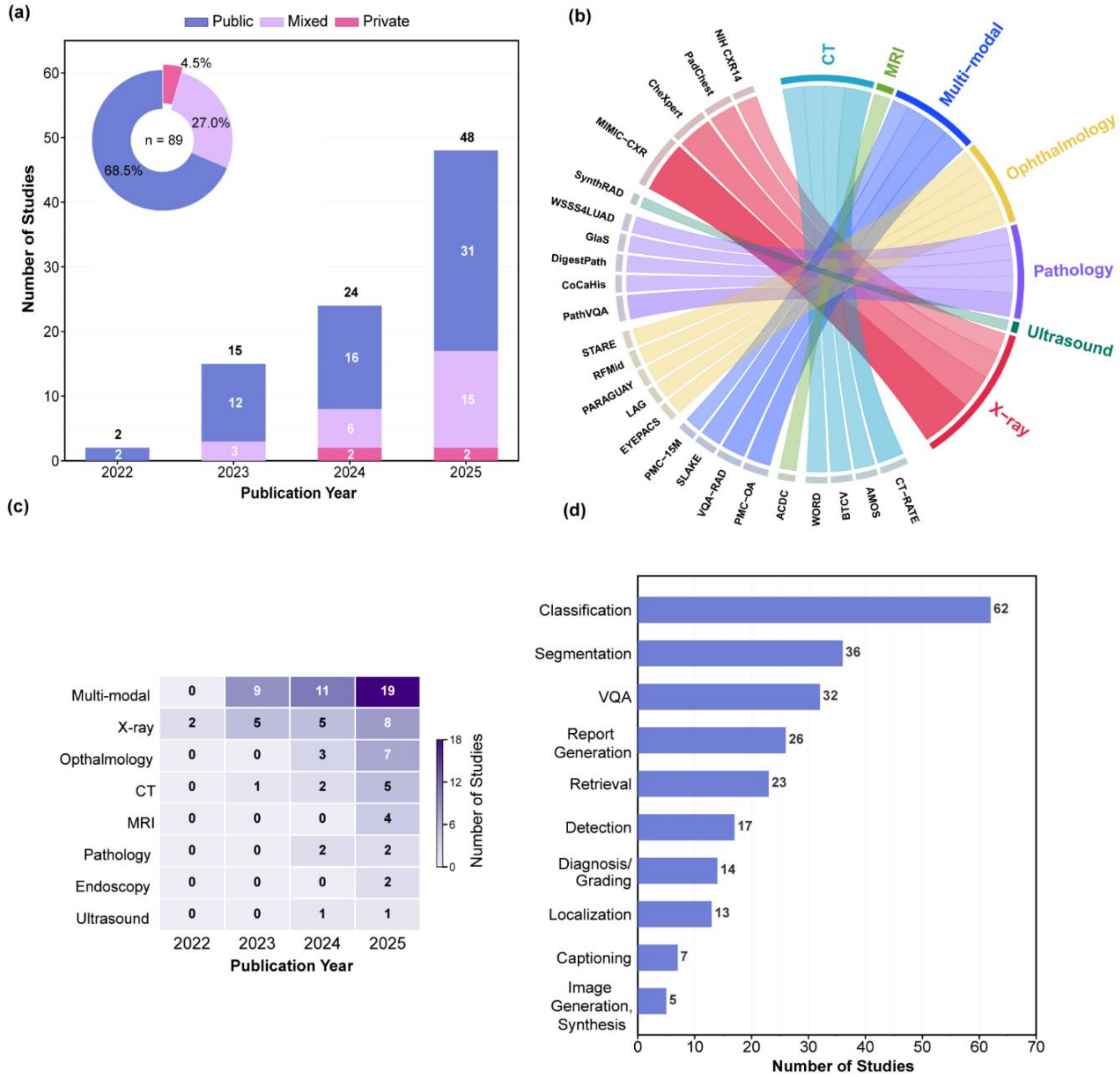

**Fig. 12**. (a) Temporal trend of VLFM publications from 2022 to 2025 along with the distribution of dataset types used for VLFM development (public, private, or mixed).; (b) chord diagram illustrating the top 10% most frequently used public datasets within each imaging modality for VLFM development; (c) heatmap of VLFM studies stratified by primary imaging modality; (d) distribution of downstream clinical tasks addressed by VLFM studies.

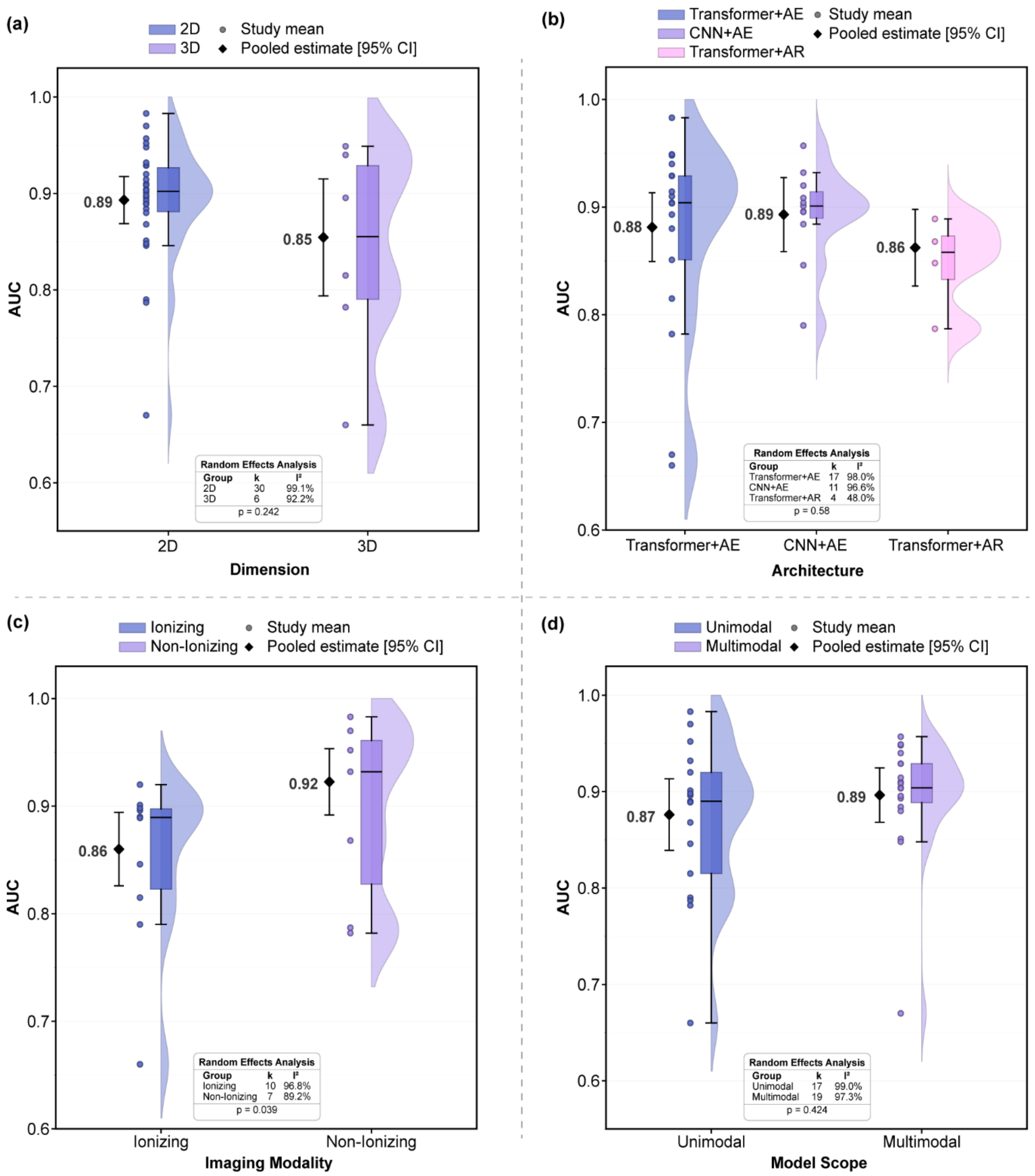

**Fig. 13.** Subgroup analyses of pooled AUC for VLFM classification studies stratified by (a) FM dimensionality (2D vs. 3D), (b) encoder architecture (Transformer+AE, CNN+AE, and Transformer+AR), (c) imaging modality (ionizing and non-ionizing), and (d) model scope (unimodal and multimodal).

### 5.3 Quantitative Analysis of VLFMs

For quantitative synthesis, 89 VLFMs published between January 2022 and December 2025 were included. The number. of VLFM publications increased exponentially from 2022 to 2025 [Fig. 12 (a)], reflecting the growing interest and importance of VLFMs in medical image analysis. With respect to dataset utilization [Fig. 12 (a)], 61 studies (68.5%) relied exclusively on public datasets, four studies (4.5%) employed private datasets, and 24 studies (27.0%) combined both public and private sources. Notably, the predominance of studies utilizing public datasets has continued to grow over recent years, underscoring the importance of open-access data for VLFM development and reproducibility. A detailed breakdown of the top 10% most frequently used public datasets within each imaging modality for VLFM development is illustrated in Fig. 12 (b). The characteristics of these datasets, including their scale and access links, are summarized in Table 2. Among these, X-ray based datasets constitute the largest proportion, followed by pathology, ophthalmology, CT, and multimodal datasets. Notably, MIMIC-CXR [274]and CheXpert emerged as the most widely used open access datasets in X-ray. Other commonly used datasets include ROCO [355], CT-RATE [284], and PathVQA [356].

### 5.3.1 Modality and Task Distribution

The temporal distribution of VLFM studies with respect to imaging modalities is shown in Fig. 12(c). Among the single-modality studies, X-ray-based VLFMs exhibited steady growth between 2022 and 2025, aligning with the prevalent use of CXR datasets in the VLFM paradigm. Similarly, CT- and ophthalmology-based VLFM studies demonstrated a progressive upward trend from 2023 to 2025. However, modalities such as US and endoscopy remain underrepresented, highlighting opportunities for future VLFM development in these domains. Fig.12(d) illustrates the distribution of downstream clinical tasks targeted by VLFM studies. Notably, classification emerged as the most predominant downstream application, reflecting the inherent synergy between contrastive vision-language pretraining and discriminative tasks, such as zero-shot and few-shot disease classification. Segmentation represented the second most common task, driven by CLIP-based segmentation frameworks and language-guided referring segmentation approaches. However, tasks such as captioning, image generation, and synthesis have been less frequently explored.

### 5.3.2 Pooled Performance Analysis of VLFMs

To quantitatively evaluate VLFM performance, we conducted a random-effects quantiative analysis focused on classification, the predominant downstream application [Fig. 12(d)]. The AUC was selected as the primary metric due to its widespread reporting among the literatures included in the quantitative synthesis. Among the 89 VLFM studies considered for quantitative analysis, 36 reported classification performance using the AUC metric and were included in the pooled analysis. The extraction, imputation, and pooling of AUC metrics followed the methodology described in Section 4.3.2. The overall pooled AUC across the 36 VLFM classification studies was 0.89 (95% CI, 0.87–0.91; $p < 0.001$), demonstrating strong classification performance across diverse clinical applications and imaging modalities.

### 5.3.3 Subgroup Analysis of VLFMs:

To evaluate the VLFM classification performance and investigate the potential sources of heterogeneity, we conducted subgroup analyses stratified by four factors: dimensionality, vision and text encoder architecture, imaging modality, and model scope. Similar to Section 4.3.3, the heterogeneity between the subgroups was assessed using the Q-test for intergroup differences, and the $I^2$ statistic was used to quantify within-group heterogeneity. The subgroup-level pooled estimates along with 95% CI and heterogeneity statistics are illustrated in Fig. 13. The detailed forest plots for each subgroup are provided in the supplementary material.

**Model Dimensionality:** We categorized the 36 studies based on model dimensionality into 2D (k=30) and 3D (k = 6) VLFM subgroups. In the random-effects analysis, the 2D subgroup achieved a pooled AUC of 0.89 ($I^2$ = 99.1%), while the 3D subgroup yielded a pooled AUC of 0.85 ($I^2$ = 92.2%) [Fig. 13(a)]. Although the pooled AUC of the 2D subgroup was higher, the intergroup difference was not statistically significant ($p = 0.242$). Notably, the wider difference in CI on the 3D subgroup may be attributed to the small number of 3D studies included in the analysis.

**Encoder Architecture:** To assess the influence of the backbone architecture on classification performance, evaluations were categorized into three subgroups based on the vision encoder and text encoder architecture combinations: a transformer vision encoder paired with an autoencoding text encoder (Transformer+AE; k = 17), a CNN vision encoder paired with an autoencoding text encoder (CNN+AE; k = 11), and a transformer vision encoder paired with an autoregressive text decoder (Transformer+AR; k = 4). Other architectural combinations were excluded due to insufficient study counts. Among the subgroups, CNN+AE achieved a pooled AUC of 0.89 ($I^2$ = 96.6%), followed by Transformer+AE subgroup with 0.88 ($I^2$ = 98.0%), and Transformer+AR subgroup with 0.86 ($I^2$ = 48.0%) [Fig. 13(b)]. No statistically significant difference was observed between these architectural combinations ($p = 0.58$), suggesting that the architectural choice of vision encoder and language backbone has a limited effect on downstream classification performance.

**Imaging Modality:** To evaluate the variations in classification performance across imaging modalities, we categorized the studies into two dominant subgroups based on their physical acquisition characteristics: modalities which use ionizing radiation (k = 10; X-ray and CT) and non-ionizing modalities (k = 7; including MRI, ophthalmology, endoscopy, dermatology, and US). In the analysis, the non-ionizing subgroup achieved a pooled AUC of 0.92 ($I^2$ = 89.2%), significantly outperforming ionizing modalities (AUC = 0.86, $I^2$ = 96.8%), and this intergroup difference was statistically significant (p

= 0.039) [Fig. 13(c)], demonstrating that imaging modality is a significant source of classification performance heterogeneity.

**Model Scope:** To analyze the generalizability of VLFMs, studies were stratified into unimodal (k = 17) and multimodal categories (k = 19) based on whether the FMs were specifically developed for unimodal or cross-modal applications. The unimodal VLFM studies achieved a pooled AUC of 0.87 ($I^2$ = 99.0%), and multimodal studies yielded a pooled AUC of 0.89 ($I^2$ = 97.3%). However, the intergroup difference was not statistically significant (p = 0.424) [Fig. 13(d)], suggesting that unimodal VLFMs achieve comparable classification performance to multimodal VLFMs.

In summary, among the four stratification factors examined, only imaging modality (p = 0.039) emerged as a statistically significant source of intergroup heterogeneity and, suggesting that the primary driver of performance variability in the VLFM classification is related to the inherent characteristics of the imaging modality rather than architectural choices, dimensionality, and model scope.

# 6. CHALLENGES AND FUTURE DIRECTIONS

## 6.1 Universal vs. Specialist FMs

The current landscape of FMs in medical imaging can be broadly classified into universal and specialist FMs. Universal FMs emphasize breadth by learning transferable representations across multiple modalities, anatomical regions, clinical tasks, and institutions. They typically adopt general-domain architectures, such as ViTs, Swin Transformers, or multimodal encoders (CLIP or Flamingo-style frameworks), along with weakly supervised and self-supervised objectives to generalize across downstream tasks. By leveraging heterogeneous datasets spanning across modalities, anatomical regions, institutions, and tasks, universal FMs enable zero-shot or few-shot adaptation through lightweight strategies, such as linear probes, prompt tuning, and PEFT. However, universal FMs often suffer from limited sensitivity to domain-specific nuances, which can reduce their effectiveness in highly specialized applications.

Specialist FMs, in contrast, emphasize depth by tailoring model architectures, training objectives, and datasets to specific clinical contexts. These models often incorporate domain-informed inductive biases, such as pyramid tiling, slide-level aggregation, or long-range 3D attention, to learn fine-grained features that universal FMs may overlook. Unlike universal FMs, which rely on weak supervision or self-supervision, specialist models utilize expert-annotated curated datasets to achieve task-specific accuracy, reliability, and robustness in narrowly defined tasks. Models such as RETFound [194] in ophthalmology and CHIEF [195] in pathology exemplify this paradigm and consistently outperform generalist backbones in narrow but clinically critical tasks.

The computational demands also differ between universal and specialist FMs. As universal FMs leverage large-scale heterogeneous datasets for pretraining, they demand substantial pretraining resources, often relying on techniques such as mixed precision, memory-efficient attention, and distributed training. Specialist FMs are comparatively more resource-efficient because they are task-constrained and typically adopt PEFT architecture to reduce training and inference costs. For instance, LiteMedSAM [150] compresses the MedSAM architecture while retaining competitive accuracy for segmentation, making it more feasible for deployment in resource-limited hospital environments.

In practice, the adoption of universal and specialist FMs depends on the clinical context, workflow integration, and deployment constraints. Universal FMs are more suitable for platform-style integration, where a single backbone can generalize across modalities and tasks, particularly in cases where labeled data are scarce. However, the broader scope of universal FMs complicates clinical validation and regulatory approval, as failure boundaries are less transparent and performance can vary substantially across institutions. Specialist FMs offer higher interpretability, consistent performance, and streamlined validation, making them better aligned with regulatory pathways and clinical workflows that demand reliability and precision.

In recent years, hybrid strategies have emerged as a bridge between universal and specialist paradigms, where the universal backbone is pretrained on a large-scale heterogeneous dataset and adapted to specific domains through lightweight mechanisms, such as PEFT, modular adapters, and task-specific prompting. For instance, MedSAM [59] extends the pretrained SAM backbone to CT, MRI, and US segmentation by adapting LoRA-based adapters. Similarly, MoPEFT [112] introduces modular PEFT strategies for dynamic adaptor selection based on the input modality. UniMed-CLIP [268] extends universal vision–language embeddings to clinical domains by adding expert-aware projection layers.

### 6.2 Training Frameworks

In the development of FMs, training frameworks serve as strategies to overcome fundamental barriers, such as label scarcity, modality fragmentation, and annotation costs. These frameworks are not standalone FMs but provide methodological scaffolding upon which FMs can be built. SSL approaches, such as DeSD and LVM-Med, demonstrate how robust and generalizable representations can be learned without extensive manual labeling. DeSD [357] mitigates weak shallow-layer representations through deep self-distillation, while LVM-Med [358] employs second-order graph matching across 1.3M images from 55 datasets to achieve modality-agnostic learning. MatchAnything [359] and UniMiSS [360] illustrate how shared embedding spaces and unified backbones can enhance FM generalization across modalities. MatchAnything addresses the challenge of cross-modality matching by pretraining with synthetic cross-modal signals and diverse datasets, enabling robust alignment between modalities, such as CT, MRI, and US. UniMiSS breaks the dimensionality barrier by combining 2D and 3D data within a medical transformer backbone with switchable patch embeddings, facilitating joint pretraining and consistent feature learning across both dimensions. Similarly, MOSMOS [361] leverages free-text reports as weak supervision, introducing global image–report and local pixel–tag alignment to reduce annotation cost and improve multi-organ segmentation, while foreshadowing the integration of LLMs to provide semantic anchors for multimodal pretraining. Collectively, these strategies substantiate that the progress of FMs depends not only on novel architectures but also on training frameworks that enable scalability, cross-modal generalization, and clinical adaptability.

### 6.3 FM Evaluation

The evaluation of FM in medical imaging analysis extends beyond traditional accuracy metrics. Unlike task-specific approaches, FMs are designed to generalize across diverse modalities, institutions, and clinical tasks. Therefore, FM evaluation requires a multifaceted framework that considers the quality of learned representations, robustness of performance under distribution shifts, and interpretability of predictions for clinical validation.

#### 6.3.1 Representation Evaluation

Representation evaluation assesses the capability of FMs to encode meaningful features in the latent space to support downstream tasks and transferability. This section summarizes the evaluation approaches employed to analyze the quality of learned representations in FMs.

**Cross-Modal Similarity:**

Cross-modal similarity evaluates the alignment of embedded representations in the latent space. In VLFMs, this is typically assessed by computing the cosine similarity between embeddings and testing cross-modal retrieval in both text-to-image and image-to-text directions, using metrics such as Recall@k, mean average precision, and normalized discounted cumulative gain. For instance, BiomedCLIP [266] is trained on PMC-15M, achieving Recall@1 of ~56% and Recall@5 of ~77% on 725,739 held-out pairs, demonstrating robust alignment between coarse- and fine-grained biomedical semantics.

Beyond retrieval, cross-modal alignment can also be examined through zero-shot classification, where text prompts are used to classify images, and performance is measured with AUROC, AUPRC, and the F1-score, complemented by per-class or per-site breakdown [266]. Semantic correlation analysis further evaluates whether clinically meaningful relationships are preserved across modalities. Correlation measures, such as Spearman's rank correlation, can probe how the latent similarity structure aligns with known clinical taxonomies, reinforcing the interpretability of biomedical VLFM embeddings.

**Clustering:**

Clustering evaluates whether the representations in the latent space preserve the intrinsic structure of the data by grouping semantically similar data points together, without task-specific supervision. In medical image analysis, clustering of embedded features in FMs are typically assessed by utilizing unsupervised techniques, such as k-means or hierarchical clustering, and quantifying cluster quality with metrics such as normalized mutual information, adjusted Rand index, or silhouette scores. For instance, Virchow [233] performed unsupervised feature analysis to separate cellular compartments on CoNSeP [362] without task-specific training.

In addition to quantitative metrics, qualitative visualization techniques such as PCA [363], t-SNE [364], and UMAP [365] are often used alongside clustering to provide intuitive assessment of meaningful structure, cluster separation, or batch effects. For example, Virchow [233] PCA maps highlight malignant epithelium versus other compartments on CoNSeP. To ensure robustness, such visualizations should be paired with formal clustering metrics (e.g., silhouette score, Davies–Bouldin index) and stratified by factors, such as site or scanner, to detect spurious or dataset-driven groupings.

**Linear Separability:**

Linear separability evaluates the quality of embedded representations by testing whether simple classifiers can effectively distinguish classes in the latent space. A common approach is the linear probing, where a fixed linear head is trained on top of frozen embeddings to measure discriminative strength. For example, Virchow [233] reported tile-level linear probe benchmarks across public and internal datasets, ranking first on most tasks and showing limited degradation under stain shifts in colorectal cancer tests.

Beyond linear heads, prototype-based methods extend this evaluation. Notably, SimpleShot utilized by UNI [232] classifies samples by assigning them to the nearest class centroids in the embedding space. This approach achieves strong prototype-based classification and retrieval across diverse organs, demonstrating that clinically coherent decision boundaries can emerge from FM embeddings without fine-tuning. A large cross-domain benchmark spanning 16 FMs and 19 datasets further reinforces these findings by demonstrating that linear probes and lightweight alternatives, such as k-NN, are reliable, data-efficient tools for evaluating representations, especially when labeled data are limited [366]. CXR-CLIP [82] further demonstrates that contrastive objectives and careful dataset design can boost classification accuracy, however at the cost of retrieval performance. Thus, both probe-based and retrieval metrics should be reported together to provide a more complete picture of representation quality and downstream utility.

### 6.3.2 Robustness Evaluation

Robustness evaluation examines the stability of FMs when exposed to distribution shifts or adversarial perturbations that arise in clinical imaging. As FMs are often subject to variations in acquisition protocols, patient populations, and imaging devices in clinics, robust generalization is intrinsic to the development of trustworthy FMs. This section summarizes the commonly used approaches for robustness evaluation, including cross-domain testing, long-tail evaluation, synthetic perturbation analysis, and adversarial robustness with uncertainty estimation.

**Cross-Domain Testing:**

Cross-domain testing evaluates whether an FM maintains stable performance across institutions, acquisition protocols, and scanners, distinct from the training distribution. Domain-generalization surveys recommend multi-site external validation with standardized preprocessing and transparent reporting, emphasizing cross-site transfer as the most prevalent stress test in medical imaging. Notably, CT-CLIP [284] validates zero-shot multi-abnormality detection and case retrieval utilizing external cohorts, demonstrating robustness to dataset shifts beyond its training corpus. Similarly, 3D CT FM Merlin utilizes ~7,000 clinical CTs and public datasets, such as VerSe [367] and TotalSegmentator [19], to demonstrate that retrieval and phenotype-classification performance can transfer across scanner manufacturers and imaging protocols. In pathology, Virchow [233] demonstrated robustness to stain variation, reporting minimal performance degradation under stain shifts in colorectal cancer cohorts. In addition, methodological reviews underscore the importance of decentralized, cross-institutional validation pipelines and systematic documentation of calibration methods, stratification strategies, and site-wise performance, to ensure that robustness claims are both clinically interpretable and reproducible [368, 369].

**Long-Tail Evaluation:**

Long-tail evaluation probes FM behavior under conditions of class imbalance and rare condition regimes to test whether FMs can generalize beyond common cases and reliably recognize underrepresented diseases, subtle abnormalities, or rare phenotypes. A common approach for long-tail evaluation is to report per-class or per-label metrics rather than aggregated scores, ensuring that performance on rare categories is not masked by dominant classes [233]. Tail-aware metrics, such as macro-AUROC, macro-F1, and balanced accuracy, are frequently used to capture performance across both frequent and rare labels [284, 369]. Reporting calibration (e.g., expected calibration error, ECE) or approximate CI (e.g., Monte Carlo dropout) is useful for safety-critical use. Listing one or two representative failure cases (e.g., protocol change, strong motion) helps with reproducibility and stress testing.

### 6.3.3 Interpretability Evaluation

Interpretability evaluation assesses whether the reasoning of FMs is grounded in clinically meaningful evidence that is transparent, reproducible, and trustworthy. Typically, interpretability evaluation involves three complementary strategies: region–entity grounding, retrieval-based evidence, and saliency-based visualization. This section summarizes these strategies and highlights their application in medical imaging FMs.

**Region-Entity Grounding:**

In medical image analysis, FMs treat grounding as a primary objective for anchoring predictions to specific anatomical or pathological regions, rather than ad hoc heatmaps. GK-MVLP [370] leverages medical knowledge aligned with anatomical regions during pretraining, allowing disease localization to be reported alongside classification, report generation, and VQA. This design transforms interpretability into quantifiable localization metrics that can be consistently compared across datasets. Similarly, KAD [371] leverages knowledge-guided disease queries to focus attention to relevant image evidence, improving both interpretability and zero/few-shot recognition.

FMs in pathology have demonstrated region-level interpretability by utilizing whole-slide and region-level outputs that can be directly linked to morphological cues. For example, CHIEF [195] systematically inspects both slide-level and localized predictions, enabling pathologists to trace model outputs back to specific histological structures. Likewise, Virchow [233] grounds predictions in tissue- and cell-level compartments, supporting pan cancer detection with external validation and offering interpretable evidence at the clinical scale. Prov-GigaPath [234] further extends this principle by highlighting human-interpretable features that correlate with molecular phenotypes, thereby substantiating the importance of region-entity grounding as a cornerstone for trust and for bridging morphologic cues with underlying biology.

**Retrieval-Based Evidence:**

VLFMs employ text-image retrieval to provide case-based evidence, allowing clinicians to audit predictions through similar cases or aligned captions. By retrieving images or reports that align with a given query, these VLFMs create an interpretable evidence trail that supports transparency and clinical validation. For example, UNI [232] evaluate retrieval alongside diverse downstream tasks, making case-based justification a routine part of their evaluation. BiomedCLIP [266] positions cross-modal retrieval as a core component, providing queryable medical entities that also serve as interpretable evidence channels. CT-CLIP [284] extends this paradigm to 3D CT imaging, enabling zero-shot detection of multiple abnormalities and retrieving comparable cases directly from volumetric scans, thereby demonstrating that retrieval can serve as an effective interpretability channel in high-dimensional medical data. Merlin [372] scales retrieval to full volumetric CT with zero-shot cross-modal capabilities and extensive external validation, demonstrating case-based evidence is reliable and useful at multiple scales, from individual slices to whole-volume analysis.

**Saliency as a Complement:**

Saliency maps, such as Grad-CAM [373], remain a primary qualitative benchmark for evaluating FMs. However, quantitative saliency benchmarks, such as expert-annotated masks with defined evaluation points, are uncommon in FM evaluations. To address this limitation, some FMs embed region–entity grounding directly into their objectives. For instance, GK-MVLP [370] reports region-level localization metrics rather than relying on free-form heatmaps. Field-wide reviews also recommend pairing saliency maps with region-level scores or retrieval-based evidence to ensure reproducibility and to avoid overinterpreting a single visualization [368, 374].

#### 6.3.4 Benchmarking and Standardized Evaluation

While the preceding subsections evaluate individual FM characteristics, recent initiatives have introduced composite benchmark suites to facilitate standardized, multitask comparisons across FMs. For instance, MultiMedBench [375] encompasses 14 biomedical tasks, including medical VQA, report generation, and classification, across diverse imaging modalities, providing a unified evaluation framework for generalist medical FMs. Similarly, RadBench [73] introduced a radiology-specific dataset across five task categories. In addition, benchmarking platforms such as OpenMEDLab 2.0 [376] have established multidimensional evaluation modules that utilize common segmented datasets across clinical domains, complementing the cross-institutional validation approaches discussed in Section 6.3.2. Despite these advancements, most existing benchmarks are limited in scope and few incorporate longitudinal or prospective evaluation components.

An unaddressed challenge in medical imaging is the evaluation of open-ended tasks, such as report generation, medical VQA, and clinical dialogue. Although metrics such as BLEU, ROUGE, and BERTScore capture surface-level textual overlaps, they often fail to reflect clinical correctness. Recently, LLMs such as GPT-4 [318] have been explored as automated judges to score model outputs against reference answers [377]. However, their reliability in capturing medical nuances remains debatable. Furthermore, hallucination of FMs remains a critical safety concern, and existing benchmarks inadequately address this issue. Emerging protocols such as NOPE [378] specifically target hallucination detection in vision-language models, but comprehensive hallucination benchmarks tailored to medical FMs remain nascent.

### 6.4 Fairness and Equity in FMs

Beyond accuracy, robustness, and interpretability, fairness represents a critical dimension in the evaluation of FMs. Empirical studies have demonstrated FMs inherit and amplify demographic and geographic biases present in their training data, resulting in unequal performance patholo age, sex, race, and regional populations [379-381]. In clinical settings, these disparities exacerbate preexisting health inequities in underrepresented groups. In FMs, bias arises from multiple stages of the pipeline. One prominent source is data imbalance, as most large-scale datasets originate from North America, Europe, and East Asia, leading to the underrepresentation of populations in the global south. This geographic skew restricts the diversity of disease phenotypes and imaging protocols, limiting model generalizability across global populations. Annotation practices also contribute to bias, as labels reflect subjective clinical judgments that may inadvertently encode social or institutional biases [382, 383]. Model architecture can also exacerbate disparities by exploiting spurious correlations that exist in the training data [384]. VLFMs may compound these challenges; a bias in the language corpora can reinforce visual disparities, amplifying inequitable outcomes.

Bias in FMs can be mitigated at different stages of the pipeline. In the preprocessing stage, bias mitigation focuses on improving dataset quality and representation. These include curating balanced datasets, supplementing underrepresented groups, and applying synthetic augmentation techniques to improve diversity in imaging protocols and disease phenotypes. In addition, fairness constraints can be embedded directly into the training by employing techniques such as adversarial debiasing or reweighting techniques to reduce spurious correlations. In the post-processing stage, strategies such as calibration, output adjustment, or threshold optimization can be applied to achieve equitable performance without retraining the FM. However, recent reviews emphasize that fairness in FMs cannot be achieved through isolated interventions. Instead, an effective mitigation strategy should employ integrated strategies that innervate over the entire FM lifecycle, encompassing systematic data documentation, metadata collection, rigorous evaluation, deployment monitoring, and governance frameworks to ensure equitable outcomes [379-381]. In addition, bias interventions must account for the utility-fairness trade-off, as mitigation strategies may reduce the accuracy of majority groups. Thus, evaluating this balance requires both group fairness and individual fairness metrics. Recent benchmarking studies, such as FairMedFM [385], have demonstrated that adaptation strategies can simultaneously improve fairness and utility.

The concentration of FM development in North America, Europe, and East Asia further aggravates inequities, as global southern populations remain severely underrepresented [386, 387]. This imbalance threatens generalizability and risks widening healthcare gaps. Addressing such disparities requires greater dataset diversification, open-weight models, and continuous monitoring of deployed systems. Ensuring equity is not only an ethical obligation but also a regulatory mandate. For example, the EU AI Act [388] classifies medical AI systems, including FMs, as high-risk, mandating fairness, transparency, and accountability. Complementary reporting frameworks, such as DECIDE-AI [389], TRIPOD+AI [390], and CLAIM 2024 [391], further substantiates that equitable outcomes are indispensable for FM validation and clinical translation. In practice, integrating fairness assessments into routine external validation and post-deployment monitoring is often sufficient for early-stage FM studies. As the field matures, embedding fairness into the design, evaluation, and governance of FMs is intrinsic to mitigate healthcare disparities.

### 6.5 Cross-Paradigm Synthesis

A combined analysis of VFMs and VLFMs on their predominant tasks (Section 4.3.3 and Section 5.3.3) yields three cross-paradigm findings. First, imaging modality is a statistically significant source of performance heterogeneity in both paradigms, suggesting that modality-specific data characteristics consistently influence model performance. Second, broad encoder architecture does not produce significant performance differences in either paradigm, indicating that architectural category alone may not explain the observed variability. Third, model scope is also non-significant across both paradigms, with unimodal and multimodal models demonstrating comparable pooled performance within the evaluated tasks. Beyond these findings, dimensionality is a significant source of performance heterogeneity for VFM segmentation but not for VLFM classification, suggesting that its impact is contingent on the specific paradigm and task rather than being consistent across domains. Collectively these results indicate that performance across these domains is primarily shaped by data and modality related factors rather than by architectural design or model scope. Notably, modality-associated differences motivate greater emphasis on modality-aware pretraining, domain adaptation, and evaluation across underrepresented imaging modalities. Furthermore, the absence of significant subgroup differences across broad architectural categories further suggests that future studies should evaluate the relative contributions of data diversity, pretraining objectives, and adaptation strategies rather than focusing predominantly on backbone modification. Finally, the emerging convergence of the two paradigms, including text-prompted segmentation and vision–language systems built upon strong visual encoders, together with the current mismatch in evaluation tasks and metrics, highlights the need for standardized cross-paradigm benchmarks that evaluate VFMs and VLFMs on common clinical tasks.

### 6.6 Practical Guidelines for FM Selection and Deployment

Translating the qualitative and quantitative synthesis presented in this study, we put forward actionable recommendations to guide researchers in FM selection and deployment. Therefore, we have organized these recommendations along four interrelated decision axes: downstream tasks, imaging modality, data dimensionality, and computational resources.

**Task-Driven Paradigm Selection:**

Downstream tasks are primary drivers of FM selection. For tasks requiring pixel-level precision, such as anatomical segmentation, tumor delineation, and organ localization, VFMs remain a natural choice. In particular, SAM-based adaptations, such as MedSAM [59] and SAM-Med3D [113], are well-suited for interactive promptable workflows. For tasks requiring automated batch segmentation, generalist segmentation frameworks, such as UniverSeg [20] and SegVol [48], are more appropriate due to their generalizability across diverse anatomical structures and imaging modalities without task-specific training. For tasks requiring cross-modal alignment, such as report generation, semantic understanding, and zero-shot disease classification, VLFMs are preferred. Specifically, contrastive VLFMs, such as BioViL-T [83]and BiomedCLIP [266], are effective for image-text retrieval and classification. For interactive and language-promptable workflows, LLaVA-Med [25]and MedVersa [320]are preferred because of their combination of a transformer-based vision encoder and an autoregressive language decoder, enabling generative, dialog-driven clinical reasoning, and flexible multi-step interactions.

**Modality-Driven Paradigm Selection:**

During FM selection, imaging modality plays a crucial role in determining both model compatibility and performance. Substantiated by quantitative analysis (Section 4.3.2), VFMs typically yield higher segmentation performance on pathology and endoscopy relative to other modalities ($p = 0.027$). For CT and X-ray applications, FMs leveraging large-scale pretraining and task-specific fine-tuning should be preferred, regardless of the paradigm; however, models pretrained on generic natural images are likely to yield suboptimal results without modality-specific adaptation. For computational pathology, modality-specific VFMs, such as PLIP [301], and CHIEF [195], offer the best starting point, given their tissue-specific pretraining objectives. For retinal imaging, RETFound [194]serves as the most reliable foundation due to its large-scale pretraining on diverse retinal fundus images. Similarly, general-purpose VFMs and VLFMs pretrained on diverse medical datasets remain an appropriate choice.

**Dimensionality-Driven Architecture Selection:**

For volumetric segmentation tasks involving CT and MRI, architectures with native 3D inductive biases, such as SAM-Med3D [113], SegVol [48], and 3D-SCUMamba [54], should be preferred as FMs relying on 2D slice-wise processing of 3D volumes often exhibit suboptimal performance and can be considered as a fallback rather than the primary choice. For 2D segmentation and classification tasks, regardless of the paradigm, FMs should be deployed at the native resolution, as upsampling of input images tends to introduce interpolation artifacts and degrade feature quality.

**Resource-Driven Selection:**

In resource-constrained environments, FMs exhibiting linear computational complexity, such as SSM-based frameworks or lightweight adaptations like Swin-LiteMedSAM [150], should be prioritized. Furthermore, for fine-tuning FMs encompassing ViT-L or ViT-H based encoders, techniques such as LoRA [100] are strongly recommended. In scenarios where data is scarce, PEFT-adapted generalist VFMs and zero-shot VLFMs exhibit competitive performance while significantly reducing annotation overhead. In summary, these recommendations are intended to serve as a practical starting point. Researchers are encouraged to incorporate prospective testing, site-specific validation, and regulatory considerations when selecting and deploying FMs in clinical settings, where interpretability–an active area of research in FM, as discussed in Sections 5.1.6 and Section 6.3.3.

### 6.6 Limitations of This Review

While this review provides a comprehensive synthesis of the current VFM and VLFM landscape with quantitative analysis, it subjected to limitations. The literature search was confined to four databases (PubMed, IEEE Xplore, ArXiv, and Google

Scholar), potentially excluding pertinent studies indexed in other repositories. Additionally, given the rapidly evolving nature of the FM landscape, this study may not encompass all studies published within the search period and works published after our search cutoff (28 February 2026). The quantitative analysis depends on performance metrics reported in the original studies, which vary in evaluation protocols, dataset splits, and preprocessing pipelines, introducing potential heterogeneity which cannot be fully controlled through subgroup analyses alone. Furthermore, the pooled performance estimates for VFMs and VLFMs are susceptible to differences in target tasks, imaging modalities, and institutional datasets; thus, the reported metrics should be interpreted as summary trends rather than definitive benchmarks.

Table. 1. Top 10% of the most frequently used datasets per imaging modality in VFM studies and their characteristics.

| Modality | Dataset | Data Scale | Task | Characteristics | Access Link |
|---|---|---|---|---|---|
| CT | BTCV | 50 volumes | Segmentation | Multi-organ segmentation benchmark; 13 abdominal organs | https://www.synapse.org/Synapse:syn3193805/wiki/217789 |
| | KiTS | 599 volumes | Segmentation | Kidney and renal tumor segmentation; paired clinical outcomes | https://kits-challenge.org/kits23/ |
| | FLARE | 2.3K volumes | Segmentation | Semi-supervised abdominal multi-organ segmentation; 13 organs | https://flare22.grand-challenge.org/ |
| | LiTS | 201 volumes | Segmentation | Liver and liver tumor segmentation benchmark | https://competitions.codalab.org/competitions/17094 |
| | NSCLC-Radiomics | 422 volumes | Segmentation, Prognosis | Lung CT radiomics; links imaging phenotypes with genomics and clinical outcomes | https://www.cancerimagingarchive.net/collection/nsclc-radiomics/ |
| | TotalSegmentator | 1.2K volumes | Segmentation | Multi-structure segmentation dataset; 104 anatomical structures | https://zenodo.org/records/6802614 |
| MRI | BraTS | ~2.0K volumes | Segmentation | Multimodal multi-institutional brain tumor MRI dataset; T1, T1ce, T2, FLAIR sequences | http://braintumorsegmentation.org/ |
| | ATLAS | ~1.3K volumes | Segmentation | Brain stroke lesion segmentation | https://fcon_1000.projects.nitrc.org/indi/retro/atlas.html |
| | ISLES | 400 volumes | Segmentation | Acute to subacute ischemic stroke lesion segmentation; FLAIR, DWI, ADC | https://isles22.grand-challenge.org/ |
| X-ray | CheXpert | ~224.3K images | Classification | CXR classification; 14 pathology labels | https://stanfordaimi.azurewebsites.net/datasets/8cbd9ed4-2eb9-4565-affc-111cf4f7ebe2 |
| Ultrasound | BUSI | 780 images | Classification, Segmentation | B-mode breast ultrasound with lesion masks; 600 female patients | https://www.kaggle.com/datasets/aryashah2k/breast-ultrasound-images-dataset |
| | CAMUS | 500 cases | Segmentation | 2D echocardiography; left ventricular endocardium, epicardium, and left atrium segmentation; varied image quality | https://www.creatis.insa-lyon.fr/Challenge/camus/ |
| Endoscopy | Kvasir-SEG | 1.0K images | Segmentation | Colonoscopy polyp segmentation | https://datasets.simula.no/kvasir-seg/ |
| | CVC-ClinicDB | 612 images | Segmentation, Detection | Colonoscopy polyp segmentation and detection | https://polyp.grand-challenge.org/CVCClinicDB/ |
| | EndoVis | ~22.1K images | Segmentation, Detection, Tracking, Recognition | Surgical video benchmark; instrument segmentation, detection, tracking, and workflow recognition | https://endovissub2018-roboticscenesegmentation.grand-challenge.org/ |
| | CholecSeg8k | 8.1K images | Segmentation | Laparoscopic cholecystectomy segmentation; 13 tissue and instrument classes | https://www.kaggle.com/datasets/newslab/cholecseg8k |
| | m2caiSeg | 307 images | Segmentation | Surgical scene segmentation; multi-organ, tissue, and instrument labeling | https://www.kaggle.com/datasets/salmanmaq/m2caiseg |
| Pathology | MoNuSeg | 44 images | Segmentation | Nucleus segmentation; H&E stained; 40x magnification; 1000x1000 tiles | https://monuseg.grand-challenge.org/ |
| | DSB 2018 | 841 images | Segmentation | Nucleus segmentation; multiple magnifications | https://www.kaggle.com/c/data-science-bowl-2018 |
| Dermoscopy | ISIC | ~550K images | Classification, Segmentation | Dermoscopic skin lesion classification and segmentation | https://www.isic-archive.com/ |
| Multi-modal | MSD | 1.4K volumes | Segmentation | Multi-organ and tumor segmentation | http://medicaldecathlon.com/ |
| | AMOS | 600 volumes | Segmentation | Multi-organ abdominal segmentation; 15 abdominal organs; multi-center | https://amos22.grand-challenge.org/ |
| | CVPR 2024 MedSAM Challenge | ~1.57M images | Segmentation | Universal medical image segmentation; FM benchmark | https://www.codabench.org/competitions/1847/ |

Table. 2. Top 10% of the most frequently used datasets per imaging modality in VLFM studies and their characteristics.

| Modality | Dataset | Data Scale | Task | Characteristics | Access Link |
|---|---|---|---|---|---|
| | CT-RATE | ~25.7K volumes | Detection, Report Generation, VQA, Retrieval | Non-contrast chest CT with radiology text reports | https://huggingface.co/datasets/ibrahimhamamci/CT-RATE |
| | WORD | 150 volumes | Segmentation | abdominal organ segmentation; 16 organs | https://github.com/HiLab-git/WORD |
| MRI | ACDC | 150 cases | Segmentation, Classification | Cardiac cine MRI; left ventricle, right ventricle, myocardium segmentation | https://www.creatis.insa-lyon.fr/Challenge/acdc/ |
| | SynthRAD | ~2.4K cases | Image Synthesis | MRI/CBCT-to-synthetic CT for radiotherapy; multi-site; head-and-neck, thorax, abdomen | https://synthrad2025.grand-challenge.org/ |
| Multi-modal | PMC-OA | 1.65M pairs | Vision-Language Pre-training | Biomedical image-caption pairs from PubMed Central | https://huggingface.co/datasets/axiong/pmc_oa |
| | VQA-RAD | 315 images | VQA | Clinician-generated QA on radiology images | https://osf.io/89kps/ |
| | SLAKE | 642 images | VQA | Bilingual medical VQA; 14,028 QA pairs; CT, MRI, X-ray | http://www.med-vqa.com/slake |
| | PMC-15M | 15M pairs | Vision-Language Pre-training | Largest biomedical image-text dataset; basis for BiomedCLIP | https://huggingface.co/microsoft/BiomedCLIP-PubMedBERT_256-vit_base_patch16_224 |
| Ophthalmology | EyePACS | ~88.7K images | Classification | Diabetic retinopathy grading; 5 severity levels | https://www.kaggle.com/c/diabetic-retinopathy-detection |
| | LAG | ~11.8K images | Detection, Classification | Glaucoma detection with attention maps from eye-tracking experiments | https://github.com/smilell/AG-CNN |
| | PARAGUAY | 757 images | Classification | Diabetic retinopathy fundus images; 7 severity levels | https://zenodo.org/records/4647952 |
| | RFMiD | 3.2K images | Multi-disease Detection | Retinal fundus multi-disease dataset; 46 conditions; 3 camera types | https://ieee-dataport.org/open-access/retinal-fundus-multi-disease-image-dataset-rfmid |
| | STARE | ~400 images | Segmentation, Detection, Diagnosis | Retinal vessel segmentation | https://cecas.clemson.edu/~ahoover/stare/ |
| Pathology | PathVQA | ~5.0K images | VQA | Pathology VQA; 32,799 QA pairs; 7 question categories | https://huggingface.co/datasets/flaviagiammarino/path-vqa |
| | CoCaHis | 82 images | Segmentation, Classification | Metastatic colon cancer in liver; H&E frozen sections; multi-annotator | http://cocahis.irb.hr/ |
| | DigestPath | ~1.5K images | Detection, Segmentation | Signet ring cell detection and colonoscopy tissue segmentation | https://digestpath2019.grand-challenge.org/ |
| | GlaS | 165 images | Instance Segmentation | Colon gland segmentation; H&E colorectal adenocarcinoma | https://www.kaggle.com/datasets/sani84/glasmiccai2015-gland-segmentation |
| | WSSS4LUAD | ~10.2K patches | Semantic Segmentation | Weakly-supervised lung adenocarcinoma | https://wsss4luad.grand-challenge.org/ |
| X-ray | PadChest | ~160K images | Classification, Detection | 174 radiographic findings; 19 diagnoses; 104 anatomic locations; Spanish reports | http://bimcv.cipf.es/bimcv-projects/padchest/ |
| | NIH CXR14 | ~112.1K images | Classification, Localization | 14 thoracic disease labels; NLP text-mined labels | https://www.kaggle.com/datasets/nih-chest-xrays/data |
| | MIMIC-CXR | ~377.1K images | Classification, Detection | De-identified CXRs with free-text radiology reports; largest public CXR dataset | https://physionet.org/content/mimic-cxr/ |

## REFERENCES

[1] D. Das, A. Sharma, P. Rajendran, and M. Pramanik, "Another decade of photoacoustic imaging," *Physics in Medicine & Biology,* vol. 66, no. 5, p. 05TR01, 2021.

[2] S. Wang *et al.*, "A generalist biomedical vision-language model via multi-CLIP knowledge distillation," *Nature Communications,* 2026.

[3] O. A. M. F. Alnaggar *et al.*, "Efficient artificial intelligence approaches for medical image processing in healthcare: comprehensive review, taxonomy, and analysis," *Artificial*

*Intelligence Review,* vol. 57, no. 8, p. 221, 2024/07/29 2024, doi: 10.1007/s10462-024-10814-2.
[4] M. Khalifa and M. Albadawy, "AI in diagnostic imaging: Revolutionising accuracy and efficiency," *Computer Methods and Programs in Biomedicine Update,* vol. 5, p. 100146, 2024/01/01/ 2024.
[5] E. S. D. Buaka and M. Z. I. Moid, "AI and medical imaging technology: evolution, impacts, and economic insights," *The Journal of Technology Transfer,* vol. 49, no. 6, pp. 2260–2272, 2024/12/01 2024, doi: 10.1007/s10961-024-10100-x.
[6] M. Chen *et al.*, "Impact of human and artificial intelligence collaboration on workload reduction in medical image interpretation," *npj Digital Medicine,* vol. 7, no. 1, p. 349, 2024/11/30 2024, doi: 10.1038/s41746-024-01328-w.
[7] R. Bommasani *et al.*, "On the opportunities and risks of foundation models," *arXiv preprint arXiv:2108.07258,* 2021.
[8] J. Devlin, M.-W. Chang, K. Lee, and K. Toutanova, "Bert: Pre-training of deep bidirectional transformers for language understanding," in *Proceedings of the 2019 conference of the North American chapter of the association for computational linguistics: human language technologies, volume 1 (long and short papers)*, 2019, pp. 4171–4186.
[9] T. Brown *et al.*, "Language models are few-shot learners," *Advances in neural information processing systems,* vol. 33, pp. 1877–1901, 2020.
[10] A. Singh *et al.*, "Openai gpt-5 system card," *arXiv preprint arXiv:2601.03267,* 2025.
[11] A. Chowdhery *et al.*, "Palm: Scaling language modeling with pathways," *Journal of Machine Learning Research,* vol. 24, no. 240, pp. 1–113, 2023.
[12] H. Touvron *et al.*, "Llama: Open and efficient foundation language models," *arXiv preprint arXiv:2302.13971,* 2023.
[13] G. Team *et al.*, "Gemini: a family of highly capable multimodal models," *arXiv preprint arXiv:2312.11805,* 2023.
[14] A. Dosovitskiy *et al.*, "An image is worth 16x16 words: Transformers for image recognition at scale," presented at the International Conference on Learning Representations, 2021, 2020.
[15] A. Radford *et al.*, "Learning transferable visual models from natural language supervision," in *International conference on machine learning*, 2021: PmLR, pp. 8748–8763.
[16] J.-B. Alayrac *et al.*, "Flamingo: a visual language model for few-shot learning," *Advances in neural information processing systems,* vol. 35, pp. 23716–23736, 2022.
[17] A. Ramesh *et al.*, "Zero-shot text-to-image generation," in *International conference on machine learning*, 2021: Pmlr, pp. 8821–8831.
[18] T. Chen, S. Kornblith, M. Norouzi, and G. Hinton, "A simple framework for contrastive learning of visual representations," in *International conference on machine learning*, 2020: PmLR, pp. 1597–1607.
[19] J. Wasserthal *et al.*, "TotalSegmentator: robust segmentation of 104 anatomic structures in CT images," *Radiology: Artificial Intelligence,* vol. 5, no. 5, p. e230024, 2023, doi: 10.1148/ryai.230024.
[20] V. I. Butoi, J. J. G. Ortiz, T. Ma, M. R. Sabuncu, J. Guttag, and A. V. Dalca, "Universeg: Universal medical image segmentation," in *Proceedings of the IEEE/CVF International Conference on Computer Vision*, 2023, pp. 21438–21451.
[21] D. Shi *et al.*, "Eyefound: a multimodal generalist foundation model for ophthalmic imaging," *arXiv preprint arXiv:2405.11338,* 2024.
[22] Z. Yang, X. Xu, J. Zhang, G. Wang, M. K. Kalra, and P. Yan, "Chest x-ray foundation model with global and local representations integration," *IEEE transactions on medical imaging,* 2025.
[23] Y. Liu, Y. Wen, L. Yang, L. He, and H. T. Shen, "RadLAS: A Foundation Model for Interpretable Radiography Image Analysis with Lesion-Aware Self-Supervised Pre-training,"

in *Proceedings of the 33rd ACM International Conference on Multimedia*, 2025, pp. 10847–10856.

[24] J. Yao *et al.*, "EVA-X: a foundation model for general chest x-ray analysis with self-supervised learning," *npj Digital Medicine,* vol. 8, no. 1, p. 678, 2025/11/17 2025, doi: 10.1038/s41746-025-02032-z.

[25] C. Li *et al.*, "LLaVA-med: training a large language-and-vision assistant for biomedicine in one day," presented at the Proceedings of the 37th International Conference on Neural Information Processing Systems, New Orleans, LA, USA, 2023.

[26] S. Srivastav *et al.*, "MAIRA at RRG24: A specialised large multimodal model for radiology report generation," in *Proceedings of the 23rd Workshop on Biomedical Natural Language Processing*, 2024, pp. 597–602.

[27] S. Xiao, Y. Zhang, L. Jiang, and Z. Wang, "ASIMSA: Advanced Semantic Information Guided Multi-Scale Alignment Framework for Medical Vision-Language Pretraining," in *2024 IEEE 9th International Conference on Computational Intelligence and Applications (ICCIA)*, 9–11 Aug. 2024 2024, pp. 99–103, doi: 10.1109/ICCIA62557.2024.10719240.

[28] S. Xu *et al.*, "Elixr: Towards a general purpose x-ray artificial intelligence system through alignment of large language models and radiology vision encoders," *arXiv preprint arXiv:2308.01317,* 2023.

[29] M. Moor *et al.*, "Med-Flamingo: a Multimodal Medical Few-shot Learner," presented at the Proceedings of the 3rd Machine Learning for Health Symposium, Proceedings of Machine Learning Research, 2023.

[30] M. Y. Lu *et al.*, "A multimodal generative AI copilot for human pathology," *Nature,* vol. 634, no. 8033, pp. 466–473, 2024, doi: 10.1038/s41586-024-07618-3.

[31] B. Azad *et al.*, "Foundational models in medical imaging: A comprehensive survey and future vision," *arXiv preprint arXiv:2310.18689,* 2023.

[32] L. Jiao *et al.*, "Foundation Model for Medical Imaging: A Comprehensive Review," *IEEE Transactions on Artificial Intelligence,* 2025.

[33] V. van Veldhuizen *et al.*, "Foundation Models in Medical Imaging--A Review and Outlook," *arXiv preprint arXiv:2506.09095,* 2025.

[34] C. Liu *et al.*, "Visual–language foundation models in medicine," *The Visual Computer,* vol. 41, no. 4, pp. 2953–2972, 2025/03/01 2025, doi: 10.1007/s00371-024-03579-w.

[35] J. S. Ryu, H. Kang, Y. Chu, and S. Yang, "Vision-language foundation models for medical imaging: a review of current practices and innovations," *Biomedical Engineering Letters,* vol. 15, no. 5, pp. 809–830, 2025.

[36] Y. Sun *et al.*, "Visual-language foundation models in medical imaging: A systematic review and meta-analysis of diagnostic and analytical applications," *Computer Methods and Programs in Biomedicine,* vol. 268, p. 108870, 2025.

[37] C. Niu, P. Wu, B. De Man, and G. Wang, "Foundation Models for Medical Imaging: Status, Challenges, and Directions," *arXiv preprint arXiv:2602.15913,* 2026.

[38] K. Simonyan and A. Zisserman, "Very deep convolutional networks for large-scale image recognition," presented at the 3rd International Conference on Learning Representations, ICLR 2015, San Diego, CA, USA, May 7-9, 2015, Conference Track Proceedings, 2015, 2015.

[39] K. He, X. Zhang, S. Ren, and J. Sun, "Deep residual learning for image recognition," in *Proceedings of the IEEE conference on computer vision and pattern recognition*, 2016, pp. 770–778.

[40] G. Huang, Z. Liu, L. Van Der Maaten, and K. Q. Weinberger, "Densely connected convolutional networks," in *Proceedings of the IEEE conference on computer vision and pattern recognition*, 2017, pp. 4700–4708.

[41] O. Ronneberger, P. Fischer, and T. Brox, "U-net: Convolutional networks for biomedical image segmentation," in *International Conference on Medical image computing and computer-assisted intervention*, 2015: Springer, pp. 234–241.
[42] F. Milletari, N. Navab, and S.-A. Ahmadi, "V-net: Fully convolutional neural networks for volumetric medical image segmentation," in *2016 fourth international conference on 3D vision (3DV)*, 2016: Ieee, pp. 565–571.
[43] T.-Y. Lin, P. Dollár, R. Girshick, K. He, B. Hariharan, and S. Belongie, "Feature pyramid networks for object detection," in *Proceedings of the IEEE conference on computer vision and pattern recognition*, 2017, pp. 2117–2125.
[44] A. Vaswani *et al.*, "Attention is all you need," *Advances in neural information processing systems,* vol. 30, 2017.
[45] Z. Liu *et al.*, "Swin transformer: Hierarchical vision transformer using shifted windows," in *Proceedings of the IEEE/CVF international conference on computer vision*, 2021, pp. 10012–10022.
[46] A. Hatamizadeh *et al.*, "Unetr: Transformers for 3d medical image segmentation," in *Proceedings of the IEEE/CVF winter conference on applications of computer vision*, 2022, pp. 574–584.
[47] A. Kirillov *et al.*, "Segment anything," in *Proceedings of the IEEE/CVF international conference on computer vision*, 2023, pp. 4015–4026.
[48] Y. Du, F. Bai, T. Huang, and B. Zhao, "Segvol: Universal and interactive volumetric medical image segmentation," *Advances in Neural Information Processing Systems,* vol. 37, pp. 110746–110783, 2024.
[49] Z. Xia, H. Li, and L. Lan, "MedFormer: hierarchical medical vision transformer with content-aware dual sparse selection attention," *Physics in Medicine & Biology,* vol. 70, no. 19, p. 195005, 2025/09/25 2025, doi: 10.1088/1361-6560/ae07a1.
[50] A. Gu and T. Dao, "Mamba: Linear-time sequence modeling with selective state spaces," in *First conference on language modeling*, 2024.
[51] L. Zhu, B. Liao, Q. Zhang, X. Wang, W. Liu, and X. Wang, "Vision mamba: efficient visual representation learning with bidirectional state space model," presented at the Proceedings of the 41st International Conference on Machine Learning, Vienna, Austria, 2024.
[52] Y. Liu *et al.*, "Vmamba: Visual state space model," *Advances in neural information processing systems,* vol. 37, pp. 103031–103063, 2024.
[53] X. Zheng *et al.*, "OpenUS: A Fully Open-Source Foundation Model for Ultrasound Image Analysis via Self-Adaptive Masked Contrastive Learning," *arXiv preprint arXiv:2511.11510,* 2025.
[54] G. M. Hassan and A. Datta, "3D-SCUMamba: An Abdominal Tumor Segmentation model," *IEEE Access,* 2025.
[55] R. Balestriero *et al.*, "A cookbook of self-supervised learning," *arXiv preprint arXiv:2304.12210,* 2023.
[56] J. E. van Engelen and H. H. Hoos, "A survey on semi-supervised learning," *Machine Learning,* vol. 109, no. 2, pp. 373–440, 2020/02/01 2020, doi: 10.1007/s10994-019-05855-6.
[57] Q. Li, W. Cai, X. Wang, Y. Zhou, D. D. Feng, and M. Chen, "Medical image classification with convolutional neural network," in *2014 13th International Conference on Control Automation Robotics & Vision (ICARCV)*, 10–12 Dec. 2014 2014, pp. 844–848, doi: 10.1109/ICARCV.2014.7064414.
[58] M. Ali *et al.*, "A review of the Segment Anything Model (SAM) for medical image analysis: Accomplishments and perspectives," *Computerized Medical Imaging and Graphics,* vol. 119, p. 102473, 2025/01/01/ 2025.
[59] J. Ma, Y. He, F. Li, L. Han, C. You, and B. Wang, "Segment anything in medical images," *Nature Communications,* vol. 15, no. 1, p. 654, 2024.

[60] X. Liu *et al.*, "Self-supervised learning: Generative or contrastive," *IEEE transactions on knowledge and data engineering,* vol. 35, no. 1, pp. 857–876, 2021.
[61] C. Doersch, A. Gupta, and A. A. Efros, "Unsupervised visual representation learning by context prediction," in *Proceedings of the IEEE international conference on computer vision*, 2015, pp. 1422–1430.
[62] M. Noroozi and P. Favaro, "Unsupervised learning of visual representations by solving jigsaw puzzles," in *European conference on computer vision*, 2016: Springer, pp. 69–84.
[63] S. Gidaris, P. Singh, and N. Komodakis, "Unsupervised representation learning by predicting image rotations," presented at the 6th International Conference on Learning Representations, ICLR 2018, Vancouver, BC, Canada, April 30 - May 3, 2018, Conference Track Proceedings, 2018, 2018.
[64] Y. LeCun, "A path towards autonomous machine intelligence version 0.9. 2, 2022-06-27," *Open Review,* vol. 62, no. 1, pp. 1–62, 2022.
[65] M. Assran *et al.*, "Self-supervised learning from images with a joint-embedding predictive architecture," in *Proceedings of the IEEE/CVF conference on computer vision and pattern recognition*, 2023, pp. 15619–15629.
[66] A. Baevski, W.-N. Hsu, Q. Xu, A. Babu, J. Gu, and M. Auli, "Data2vec: A general framework for self-supervised learning in speech, vision and language," in *International conference on machine learning*, 2022: PMLR, pp. 1298–1312.
[67] D. Chen *et al.*, "Vl-jepa: Joint embedding predictive architecture for vision-language," *arXiv preprint arXiv:2512.10942,* 2025.
[68] L. Jing and Y. Tian, "Self-supervised visual feature learning with deep neural networks: A survey," *IEEE transactions on pattern analysis and machine intelligence,* vol. 43, no. 11, pp. 4037–4058, 2020.
[69] K. He, X. Chen, S. Xie, Y. Li, P. Dollár, and R. Girshick, "Masked autoencoders are scalable vision learners," in *Proceedings of the IEEE/CVF conference on computer vision and pattern recognition*, 2022, pp. 16000–16009.
[70] D. Pathak, P. Krahenbuhl, J. Donahue, T. Darrell, and A. A. Efros, "Context encoders: Feature learning by inpainting," in *Proceedings of the IEEE conference on computer vision and pattern recognition*, 2016, pp. 2536–2544.
[71] Z. Liu *et al.*, "VIS-MAE: An Efficient Self-supervised Learning Approach on Medical Image Segmentation and Classification," in *International Workshop on Machine Learning in Medical Imaging*, 2024: Springer, pp. 95–107.
[72] W. Wang *et al.*, "Image as a foreign language: Beit pretraining for vision and vision-language tasks," in *Proceedings of the IEEE/CVF Conference on Computer Vision and Pattern Recognition*, 2023, pp. 19175–19186.
[73] C. Wu, X. Zhang, Y. Zhang, H. Hui, Y. Wang, and W. Xie, "Towards generalist foundation model for radiology by leveraging web-scale 2D&3D medical data," *Nature Communications,* vol. 16, no. 1, p. 7866, 2025/08/23 2025, doi: 10.1038/s41467-025-62385-7.
[74] Y. Yu, Y. Gu, S. Zhang, and X. Zhang, "Meddiff-fm: A diffusion-based foundation model for versatile medical image applications," *arXiv preprint arXiv:2410.15432,* 2024.
[75] K. He, H. Fan, Y. Wu, S. Xie, and R. Girshick, "Momentum contrast for unsupervised visual representation learning," in *Proceedings of the IEEE/CVF conference on computer vision and pattern recognition*, 2020, pp. 9729–9738.
[76] M. Caron, I. Misra, J. Mairal, P. Goyal, P. Bojanowski, and A. Joulin, "Unsupervised learning of visual features by contrasting cluster assignments," *Advances in neural information processing systems,* vol. 33, pp. 9912–9924, 2020.
[77] M. Caron *et al.*, "Emerging properties in self-supervised vision transformers," in *Proceedings of the IEEE/CVF international conference on computer vision*, 2021, pp. 9650–9660.

[78] M. Oquab *et al.*, "DINOv2: Learning Robust Visual Features without Supervision," *Transactions on Machine Learning Research,* 2024 2024.
[79] O. Siméoni *et al.*, "DINOv3," *arXiv preprint arXiv:2508.10104,* 2025.
[80] Z. Wang, Z. Wu, D. Agarwal, and J. Sun, "Medclip: Contrastive learning from unpaired medical images and text," in *Proceedings of the Conference on Empirical Methods in Natural Language Processing. Conference on Empirical Methods in Natural Language Processing*, 2022, vol. 2022, p. 3876.
[81] W. Lin *et al.*, "Pmc-clip: Contrastive language-image pre-training using biomedical documents," in *International Conference on Medical Image Computing and Computer-Assisted Intervention*, 2023: Springer, pp. 525–536.
[82] K. You *et al.*, "CXR-CLIP: Toward Large Scale Chest X-ray Language-Image Pre-training," in *Medical Image Computing and Computer Assisted Intervention – MICCAI 2023*, Cham, H. Greenspan *et al.*, Eds., 2023// 2023: Springer Nature Switzerland, pp. 101–111.
[83] S. Bannur *et al.*, "Learning to Exploit Temporal Structure for Biomedical Vision-Language Processing," in *2023 IEEE/CVF Conference on Computer Vision and Pattern Recognition (CVPR)*, 17–24 June 2023 2023, pp. 15016–15027.
[84] K. Sohn *et al.*, "Fixmatch: Simplifying semi-supervised learning with consistency and confidence," *Advances in neural information processing systems,* vol. 33, pp. 596–608, 2020.
[85] Z. Cai, A. Ravichandran, S. Maji, C. Fowlkes, Z. Tu, and S. Soatto, "Exponential moving average normalization for self-supervised and semi-supervised learning," in *Proceedings of the IEEE/CVF conference on computer vision and pattern recognition*, 2021, pp. 194–203.
[86] T. Chen, S. Kornblith, K. Swersky, M. Norouzi, and G. E. Hinton, "Big self-supervised models are strong semi-supervised learners," *Advances in neural information processing systems,* vol. 33, pp. 22243–22255, 2020.
[87] Z. Weng, X. Yang, A. Li, Z. Wu, and Y.-G. Jiang, "Semi-supervised vision transformers," in *European conference on computer vision*, 2022: Springer, pp. 605–620.
[88] T. Kaufmann, P. Weng, V. Bengs, and E. Hüllermeier, "A survey of reinforcement learning from human feedback," 2024.
[89] J. Schulman, F. Wolski, P. Dhariwal, A. Radford, and O. Klimov, "Proximal policy optimization algorithms," *arXiv preprint arXiv:1707.06347,* 2017.
[90] Z. Shao *et al.*, "Deepseekmath: Pushing the limits of mathematical reasoning in open language models," *arXiv preprint arXiv:2402.03300,* 2024.
[91] J. Wang *et al.*, "Self-improving generative foundation model for synthetic medical image generation and clinical applications," *Nature Medicine,* vol. 31, no. 2, pp. 609–617, 2025/02/01 2025, doi: 10.1038/s41591-024-03359-y.
[92] K. Zhang, C. Malon, L. Sun, and M. R. Min, "EditGRPO: Reinforcement Learning with Post -Rollout Edits for Clinically Accurate Chest X-Ray Report Generation," Mumbai, India, December 2025: The Asian Federation of Natural Language Processing and The Association for Computational Linguistics, in Proceedings of the 14th International Joint Conference on Natural Language Processing and the 4th Conference of the Asia-Pacific Chapter of the Association for Computational Linguistics, pp. 304–316.
[93] Z. Fan, C. Liang, C. Wu, Y. Zhang, Y. Wang, and W. Xie, "ChestX-Reasoner: Advancing Radiology Foundation Models with Reasoning through Step-by-Step Verification," *arXiv preprint arXiv:2504.20930,* 2025.
[94] S. Jiang, F. Liu, Z. Wang, L. Cai, and Y. Zhang, "PathReasoner-R1: Instilling Structured Reasoning into Pathology Vision-Language Model via Knowledge-Guided Policy Optimization," *arXiv preprint arXiv:2601.21617,* 2026.
[95] W. Zhang *et al.*, "Patho-AgenticRAG: towards multimodal agentic retrieval-augmented generation for pathology VLMs via reinforcement learning," *arXiv preprint arXiv:2508.02258,* 2025.

[96] H. Bahng, A. Jahanian, S. Sankaranarayanan, and P. Isola, "Exploring visual prompts for adapting large-scale models," *arXiv preprint arXiv:2203.17274,* 2022.
[97] M. Jia *et al.*, "Visual prompt tuning," in *European conference on computer vision*, 2022: Springer, pp. 709–727.
[98] M. Awais *et al.*, "Foundation models defining a new era in vision: a survey and outlook," *IEEE Transactions on Pattern Analysis and Machine Intelligence,* 2025.
[99] L. Xu, H. Xie, S. J. Qin, X. Tao, and F. L. Wang, "Parameter-efficient fine-tuning methods for pretrained language models: A critical review and assessment," *IEEE Transactions on Pattern Analysis and Machine Intelligence,* 2026.
[100] D. Zhang, T. Feng, L. Xue, Y. Wang, Y. Dong, and J. Tang, "Parameter-efficient fine-tuning for foundation models," *arXiv preprint arXiv:2501.13787,* 2025.
[101] Y. Zhu *et al.*, "Melo: Low-rank adaptation is better than fine-tuning for medical image diagnosis," in *2024 IEEE International Symposium on Biomedical Imaging (ISBI)*, 2024: IEEE, pp. 1–5.
[102] C. Lian, H.-Y. Zhou, Y. Yu, and L. Wang, "Less could be better: Parameter-efficient fine-tuning advances medical vision foundation models," *arXiv preprint arXiv:2401.12215,* 2024.
[103] B. Peng, C. Li, P. He, M. Galley, and J. Gao, "Instruction tuning with gpt-4," *arXiv preprint arXiv:2304.03277,* 2023.
[104] J. Wei *et al.*, "Finetuned language models are zero-shot learners," presented at the International Conference on Learning Representations, 2022, 2022.
[105] H. Liu, C. Li, Q. Wu, and Y. J. Lee, "Visual instruction tuning," *Advances in neural information processing systems,* vol. 36, pp. 34892–34916, 2023.
[106] J. Wu and M. Xu, "One-prompt to segment all medical images," in *Proceedings of the IEEE/CVF conference on computer vision and pattern recognition*, 2024, pp. 11302–11312.
[107] J. Xu, "GMISeg: General Medical Image Segmentation without Re-Training," *arXiv preprint arXiv:2311.12539,* 2023.
[108] P. Liang *et al.*, "Vision foundation models in medical image analysis: Advances and challenges," *arXiv preprint arXiv:2502.14584,* 2025.
[109] K. Zhang and D. Liu, "Customized segment anything model for medical image segmentation," *arXiv preprint arXiv:2304.13785,* 2023.
[110] J. Cheng *et al.*, "SAM-Med2D," *arXiv [cs.CV],* 2023 2023.
[111] J. Wu *et al.*, "Medical sam adapter: Adapting segment anything model for medical image segmentation," *Medical image analysis,* vol. 102, p. 103547, 2025.
[112] R. Sahay and A. Savakis, "Mopeft: A mixture-of-pefts for the segment anything model," in *Proceedings of the Computer Vision and Pattern Recognition Conference*, 2025, pp. 6500–6510.
[113] H. Wang *et al.*, "SAM-Med3D: A Vision Foundation Model for General-Purpose Segmentation on Volumetric Medical Images," *IEEE Transactions on Neural Networks and Learning Systems,* vol. 36, no. 10, pp. 17599–17612, 2025, doi: 10.1109/TNNLS.2025.3586694.
[114] Harrigr, "Segmentation Outside the Cranial Vault Challenge," 2015 2015, doi: 10.7303/SYN3193805.
[115] S. Gong *et al.*, "3DSAM-adapter: Holistic adaptation of SAM from 2D to 3D for promptable tumor segmentation," *Medical Image Analysis,* vol. 98, p. 103324, 2024/12/01/ 2024.
[116] W. Lei, W. Xu, K. Li, X. Zhang, and S. Zhang, "MedLSAM: Localize and segment anything model for 3D CT images," *Medical Image Analysis,* vol. 99, p. 103370, 2025.
[117] Y. Shen, D. Dreizin, B. Inigo, and M. Unberath, "ProtoSAM-3D: Interactive semantic segmentation in volumetric medical imaging via a Segment Anything Model and mask-level prototypes," *Computerized Medical Imaging and Graphics,* vol. 121, p. 102501, 2025.

[118] C. Chen *et al.*, "Ma-sam: Modality-agnostic sam adaptation for 3d medical image segmentation," *Medical Image Analysis,* vol. 98, p. 103310, 2024.
[119] J. Zhu, A. Hamdi, Y. Qi, Y. Jin, and J. Wu, "Medical sam 2: Segment medical images as video via segment anything model 2," *arXiv preprint arXiv:2408.00874,* 2024.
[120] X. Yan *et al.*, "After-sam: Adapting sam with axial fusion transformer for medical imaging segmentation," in *Proceedings of the IEEE/CVF Winter Conference on Applications of Computer Vision*, 2024, pp. 7975–7984.
[121] Z. Huang, S. R. Guariglia, J. Yang, and C.-L. Tsai, "ECT-3DMedSAM: Efficient Cross Teaching Using Segment Anything Model for Semi-Supervised 3D Medical Image Segmentation," in *Medical Imaging with Deep Learning*, 2026.
[122] T. Shaharabany, A. Dahan, R. Giryes, and L. Wolf, "Autosam: Adapting sam to medical images by overloading the prompt encoder," *arXiv preprint arXiv:2306.06370,* 2023.
[123] P. Chao, C.-Y. Kao, Y.-S. Ruan, C.-H. Huang, and Y.-L. Lin, "Hardnet: A low memory traffic network," in *Proceedings of the IEEE/CVF international conference on computer vision*, 2019, pp. 3552–3561.
[124] Q. Xu *et al.*, "Esp-medsam: Efficient self-prompting sam for universal domain-generalized medical image segmentation," *arXiv preprint arXiv:2407.14153,* 2024.
[125] A. S. Wahd *et al.*, "Sam2Rad: A segmentation model for medical images with learnable prompts," *Computers in Biology and Medicine,* vol. 187, p. 109725, 2025.
[126] S. Pandey, K.-F. Chen, and E. B. Dam, "Comprehensive multimodal segmentation in medical imaging: Combining yolov8 with sam and hq-sam models," in *Proceedings of the IEEE/CVF international conference on computer vision*, 2023, pp. 2592–2598.
[127] L. Ayzenberg, R. Giryes, and H. Greenspan, "ProtoSAM for automated one shot medical image segmentation using foundational models," *Scientific Reports,* vol. 15, no. 1, p. 41482, 2025/11/24 2025, doi: 10.1038/s41598-025-06643-0.
[128] H. Zhu *et al.*, "SSS: Semi-Supervised SAM-2 with Efficient Prompting for Medical Imaging Segmentation," *arXiv preprint arXiv:2506.08949,* 2025.
[129] J. Wang, Y. Mei, J. Liu, and X. Fan, "SAM-Guided Robust Representation Learning for One-Shot 3D Medical Image Segmentation," *arXiv preprint arXiv:2504.20501,* 2025.
[130] B. Towle, X. Chen, and K. Zhou, "SimSAM: Zero-shot medical image segmentation via simulated interaction," in *2024 IEEE International Symposium on Biomedical Imaging (ISBI)*, 2024: IEEE, pp. 1–5.
[131] Y. Xu, J. Tang, A. Men, and Q. Chen, "EviPrompt: A training-free evidential prompt generation method for adapting segment anything model in medical images," *IEEE Transactions on Image Processing,* 2024.
[132] C. Li, R. I. Sultan, P. Khanduri, Y. Qiang, C. Indrin, and D. Zhu, "Autoprosam: Automated prompting sam for 3d multi-organ segmentation," in *2025 IEEE/CVF Winter Conference on Applications of Computer Vision (WACV)*, 2025: IEEE, pp. 3570–3580.
[133] B. Xie, H. Tang, Y. Yan, and G. Agam, "Rfmedsam 2: Automatic prompt refinement for enhanced volumetric medical image segmentation with sam 2," *arXiv preprint arXiv:2502.02741,* 2025.
[134] Y. Xing, J. Wu, Y. Bu, and K. Gong, "SAM2-SGP: Enhancing SAM2 for Medical Image Segmentation via Support-Set Guided Prompting," *arXiv preprint arXiv:2506.19658,* 2025.
[135] S. Dai, K. Ye, G. Liu, H. Tang, and L. Zhan, "Zeus: Zero-shot LLM Instruction for Union Segmentation in Multimodal Medical Imaging," *arXiv preprint arXiv:2504.07336,* 2025.
[136] G. Deng *et al.*, "Sam-u: Multi-box prompts triggered uncertainty estimation for reliable sam in medical image," in *International Conference on Medical Image Computing and Computer-Assisted Intervention*, 2023: Springer, pp. 368–377.

[137] A. Guo, G. Fei, H. Pasupuleti, and J. Wang, "ClickSAM: Fine-tuning Segment Anything Model using click prompts for ultrasound image segmentation," in *Medical Imaging 2024: Ultrasonic Imaging and Tomography*, 2024, vol. 12932: SPIE, pp. 240–244.
[138] S. Yang, H. Bi, H. Zhang, and J. Sun, "Sam-unet: Enhancing zero-shot segmentation of sam for universal medical images," *arXiv preprint arXiv:2408.09886,* vol. 10, 2024.
[139] P. Tian, X. Chen, H. Bi, and F. Li, "MedSAM-CA: A CNN-Augmented ViT with Attention-Enhanced Multi-Scale Fusion for Medical Image Segmentation," *arXiv preprint arXiv:2506.23700,* 2025.
[140] S. Chai *et al.*, "Ladder fine-tuning approach for sam integrating complementary network," *Procedia Computer Science,* vol. 246, pp. 4951–4958, 2024.
[141] Y. Wang and L. Xiao, "Samda: Leveraging sam on few-shot domain adaptation for electronic microscopy segmentation," in *2025 IEEE 22nd International Symposium on Biomedical Imaging (ISBI)*, 2025: IEEE, pp. 1–5.
[142] D. Anand *et al.*, "One-shot localization and segmentation of medical images with foundation models," *arXiv preprint arXiv:2310.18642,* 2023.
[143] H. Li, H. Liu, D. Hu, J. Wang, and I. Oguz, "Promise: Prompt-driven 3d medical image segmentation using pretrained image foundation models," in *2024 IEEE international symposium on biomedical imaging (ISBI)*, 2024: IEEE, pp. 1–5.
[144] Y. Chen *et al.*, "Accelerating Volumetric Medical Image Annotation via Short-Long Memory SAM 2," *IEEE Transactions on Medical Imaging,* pp. 1–1, 2025, doi: 10.1109/TMI.2025.3627954.
[145] S. Li, L. Peng, Z. Zhang, G. Durak, and U. Bagci, "TAGS: 3D Tumor-Adaptive Guidance for SAM," in *Proceedings of the IEEE/CVF International Conference on Computer Vision*, 2025, pp. 4352–4362.
[146] Y. Chen, W. Cui, X. Dong, S. Zhou, and Z. Wang, "SAM2Med3D: Leveraging video foundation models for 3D breast MRI segmentation," *Computers & Graphics,* vol. 132, p. 104341, 2025/11/01/ 2025.
[147] Y. Xiong *et al.*, "Efficientsam: Leveraged masked image pretraining for efficient segment anything," in *Proceedings of the IEEE/CVF Conference on Computer Vision and Pattern Recognition*, 2024, pp. 16111–16121.
[148] W. Liu, R. Bian, H. Li, and T. Tian, "LiteMedSAM with low-rank adaptation and multi-box efficient inference for medical image segmentation," 2024.
[149] K. Wu *et al.*, "Tinyvit: Fast pretraining distillation for small vision transformers," in *European conference on computer vision*, 2022: Springer, pp. 68–85.
[150] R. Gao, D. Lyu, and M. Staring, "Swin-LiteMedSAM: A lightweight box-based segment anything model for large-scale medical image datasets," in *Medical Image Segmentation Challenge*: Springer, 2024, pp. 70–82.
[151] Y. Kong, K. Kim, S. Jeong, K. E. Lee, and H.-J. Kong, "SwiftMedSAM: An Ultra-lightweight Prompt-Based Universal Medical Image Segmentation Model for Highly Constrained Environments," in *Medical Image Segmentation Challenge*: Springer, 2024, pp. 180–194.
[152] Q. Ali, Y. Chen, and A. Wong, "RepViT-MedSAM: Efficient Segment Anything in the Medical Images," in *Medical Image Segmentation Challenge*: Springer, 2024, pp. 195–205.
[153] A. Wang, H. Chen, Z. Lin, J. Han, and G. Ding, "Repvit: Revisiting mobile cnn from vit perspective," in *Proceedings of the IEEE/CVF conference on computer vision and pattern recognition*, 2024, pp. 15909–15920.
[154] Z. Zhang, R. Huang, and N. Huang, "RepMedSAM: Segment Anything in Medical Images with Lightweight CNN," in *CVPR 2024: Segment Anything In Medical Images On Laptop*, 2024.

[155] B.-H. Le, D.-K. Nguyen-Vu, T.-H. Nguyen-Mau, H.-D. Nguyen, and M.-T. Tran, "MedficientSAM: a robust medical segmentation model with optimized inference pipeline for limited clinical settings," in *Medical Image Segmentation Challenge*: Springer, 2024, pp. 1–14.
[156] H. Cai, J. Li, M. Hu, C. Gan, and S. Han, "Efficientvit: Lightweight multi-scale attention for high-resolution dense prediction," in *Proceedings of the IEEE/CVF international conference on computer vision*, 2023, pp. 17302–17313.
[157] A. Pfefferle, L. Purucker, and F. Hutter, "Daft: Data-aware fine-tuning of foundation models for efficient and effective medical image segmentation," in *Medical Image Segmentation Challenge*: Springer, 2024, pp. 15–38.
[158] Y. Shen *et al.*, "Fastsam3d: An efficient segment anything model for 3d volumetric medical images," in *International Conference on Medical Image Computing and Computer-Assisted Intervention*, 2024: Springer, pp. 542–552.
[159] Q. Xu *et al.*, "De-LightSAM: Modality-Decoupled Lightweight SAM for Generalizable Medical Segmentation," *arXiv [eess.IV],* 2025 2025.
[160] Y. Luo, Q. Xu, J. Feng, G. Qian, and W. Duan, "Med-FastSAM: Improving Transfer Efficiency of SAM to Domain-Generalised Medical Image Segmentation," in *Advancements In Medical Foundation Models: Explainability, Robustness, Security, and Beyond*.
[161] D. Lyu, R. Gao, and M. Staring, "MCP-MedSAM: A Powerful Lightweight Medical Segment Anything Model Trained with a Single GPU in Just One Day," *Machine Learning for Biomedical Imaging,* vol. 3, pp. 135–151, 2025 2025, doi: 10.59275/j.melba.2025-4849.
[162] X. Lin, Y. Xiang, L. Yu, and Z. Yan, "Beyond adapting SAM: Towards end-to-end ultrasound image segmentation via auto prompting," in *International Conference on Medical Image Computing and Computer-Assisted Intervention*, 2024: Springer, pp. 24–34.
[163] S. N. Gowda and D. A. Clifton, "CC-SAM: SAM with Cross-Feature Attention and Context for Ultrasound Image Segmentation," in *Computer Vision – ECCV 2024*, Cham, A. Leonardis, E. Ricci, S. Roth, O. Russakovsky, T. Sattler, and G. Varol, Eds., 2025// 2025: Springer Nature Switzerland, pp. 108–124.
[164] H. Ravishankar, R. Patil, V. Melapudi, and P. Annangi, "Sonosam-segment anything on ultrasound images," in *International workshop on advances in simplifying medical ultrasound*, 2023: Springer, pp. 23–33.
[165] X. Deng, H. Wu, R. Zeng, and J. Qin, "Memsam: Taming segment anything model for echocardiography video segmentation," in *Proceedings of the IEEE/CVF Conference on Computer Vision and Pattern Recognition*, 2024, pp. 9622–9631.
[166] H. Ravishankar *et al.*, "SonoSAMTrack--Segment and Track Anything on Ultrasound Images," *arXiv preprint arXiv:2310.16872,* 2023.
[167] Y. Sheng, S. Bano, M. J. Clarkson, and M. Islam, "Surgical-DeSAM: decoupling SAM for instrument segmentation in robotic surgery," *International Journal of Computer Assisted Radiology and Surgery,* vol. 19, no. 7, pp. 1267–1271, 2024.
[168] W. Yue, J. Zhang, K. Hu, Y. Xia, J. Luo, and Z. Wang, "Surgicalsam: Efficient class promptable surgical instrument segmentation," in *Proceedings of the AAAI Conference on Artificial Intelligence*, 2024, vol. 38, no. 7, pp. 6890–6898.
[169] W. Yue *et al.*, "Surgicalpart-sam: Part-to-whole collaborative prompting for surgical instrument segmentation," *arXiv preprint arXiv:2312.14481,* 2023.
[170] N. M. Matasyoh, F. Mathis-Ullrich, and R. A. Zeineldin, "SAMSurg: Surgical Instrument Segmentation in Robotic Surgeries Using Vision Foundation Model," *IEEE Access,* vol. 12, pp. 193950–193959, 2024, doi: 10.1109/access.2024.3520386.
[171] J. N. Paranjape, N. G. Nair, S. Sikder, S. S. Vedula, and V. M. Patel, "Adaptivesam: Towards efficient tuning of sam for surgical scene segmentation," in *Annual Conference on Medical Image Understanding and Analysis*, 2024: Springer, pp. 187–201.

[172] D. N. Kamtam *et al.*, "Surgisam2: fine-tuning a foundational model for surgical video anatomy segmentation and detection," *arXiv preprint arXiv:2503.03942,* 2025.
[173] J. Zhang *et al.*, "Sam-path: A segment anything model for semantic segmentation in digital pathology," in *International Conference on Medical Image Computing and Computer-Assisted Intervention*, 2023: Springer, pp. 161–170.
[174] R. J. Chen *et al.*, "Scaling Vision Transformers to Gigapixel Images via Hierarchical Self-Supervised Learning," in *2022 IEEE/CVF Conference on Computer Vision and Pattern Recognition (CVPR)*, 18–24 June 2022 2022, pp. 16123–16134, doi: 10.1109/CVPR52688.2022.01567.
[175] H. Liu, H. Yang, P. J. v. Diest, J. P. W. Pluim, and M. Veta, "WSI-SAM: Multi-resolution Segment Anything Model (SAM) for histopathology whole-slide images," presented at the Proceedings of the MICCAI Workshop on Computational Pathology, Proceedings of Machine Learning Research, 2024.
[176] C. Wang *et al.*, "SegAnyPath: a foundation model for multi-resolution stain-variant and multi-task pathology image segmentation," *IEEE Transactions on Medical Imaging,* 2024.
[177] Z. Chen, Q. Xu, X. Liu, and Y. Yuan, "UN-SAM: Domain-adaptive self-prompt segmentation for universal nuclei images," *Medical Image Analysis,* vol. 103, p. 103607, 2025/07/01/ 2025.
[178] U. Israel *et al.*, "CellSAM: a foundation model for cell segmentation," *BioRxiv,* p. 2023.11. 17.567630, 2025.
[179] A. Archit *et al.*, "Segment anything for microscopy," *Nature Methods,* vol. 22, no. 3, pp. 579–591, 2025.
[180] Y. Zhou *et al.*, "Multimodal, Multi-Disease Medical Imaging Foundation Model (MerMED-FM)," *arXiv preprint arXiv:2507.00185,* 2025.
[181] C. Dancette *et al.*, "Curia: A Multi-Modal Foundation Model for Radiology," *arXiv preprint arXiv:2509.06830,* 2025.
[182] T. Moutakanni *et al.*, "Advancing human-centric ai for robust x-ray analysis through holistic self-supervised learning," *arXiv preprint arXiv:2405.01469,* 2024.
[183] L. Xu, Z. Ni, H. Sun, H. Li, and S. Zhang, "A foundation model for generalizable disease diagnosis in chest X-ray images," *arXiv preprint arXiv:2410.08861,* 2024.
[184] J. Jiao *et al.*, "Usfm: A universal ultrasound foundation model generalized to tasks and organs towards label efficient image analysis," *Medical image analysis,* vol. 96, p. 103202, 2024.
[185] Z. Gao *et al.*, "A Lung CT Foundation Model Facilitating Disease Diagnosis and Medical Imaging," Cold Spring Harbor Laboratory, 2025.
[186] S. Pai *et al.*, "Vision foundation models for computed tomography," *arXiv preprint arXiv:2501.09001,* 2025.
[187] E. Jun, S. Jeong, D.-W. Heo, and H.-I. Suk, "Medical transformer: Universal encoder for 3-D brain MRI analysis," *IEEE Transactions on Neural Networks and Learning Systems,* 2023.
[188] H. Dong *et al.*, "MRI-CORE: A Foundation Model for Magnetic Resonance Imaging," *arXiv preprint arXiv:2506.12186,* 2025.
[189] Y. Sun, L. Wang, G. Li, W. Lin, and L. Wang, "A foundation model for enhancing magnetic resonance images and downstream segmentation, registration and diagnostic tasks," *Nature Biomedical Engineering,* vol. 9, no. 4, pp. 521–538, 2025.
[190] J. Zhang, C. Xu, H. Zhong, Y. Cao, and L. Zhao, "BDFM: Foundation Model for Segmentation and Classification Tasks of Brain Diseases," *IEEE Transactions on Biomedical Engineering,* pp. 1–12, 2025, doi: 10.1109/TBME.2025.3637146.
[191] D. Tak *et al.*, "A foundation model for generalized brain MRI analysis," *medRxiv,* 2024.
[192] Y. Fu *et al.*, "A versatile foundation model for cine cardiac magnetic resonance image analysis tasks," *arXiv preprint arXiv:2506.00679,* 2025.

[193] J. Qiu *et al.*, "Development and Validation of a Multimodal Multitask Vision Foundation Model for Generalist Ophthalmic Artificial Intelligence," *NEJM AI,* vol. 1, no. 12, p. AIoa2300221, 2024, doi: doi:10.1056/AIoa2300221.
[194] Y. Zhou *et al.*, "A foundation model for generalizable disease detection from retinal images," *Nature,* vol. 622, no. 7981, pp. 156–163, 2023.
[195] X. Wang *et al.*, "A pathology foundation model for cancer diagnosis and prognosis prediction," *Nature,* vol. 634, no. 8035, pp. 970–978, 2024/10/01 2024, doi: 10.1038/s41586-024-07894-z.
[196] T. Ding *et al.*, "A multimodal whole-slide foundation model for pathology," *Nature Medicine,* vol. 31, no. 11, pp. 3749–3761, 2025/11/01 2025, doi: 10.1038/s41591-025-03982-3.
[197] Z. Yang *et al.*, "A foundation model for generalizable cancer diagnosis and survival prediction from histopathological images," *Nature Communications,* vol. 16, no. 1, p. 2366, 2025.
[198] J. N. Weinstein *et al.*, "The cancer genome atlas pan-cancer analysis project," *Nature genetics,* vol. 45, no. 10, pp. 1113–1120, 2013.
[199] F. Yan *et al.*, "PathOrchestra: a comprehensive foundation model for computational pathology with over 100 diverse clinical-grade tasks," *npj Digital Medicine,* vol. 8, no. 1, p. 695, 2025/11/19 2025, doi: 10.1038/s41746-025-02027-w.
[200] J. Pohjonen *et al.*, "HistoEncoder: a digital pathology foundation model for prostate cancer," *arXiv preprint arXiv:2411.11458,* 2024.
[201] D. Juyal *et al.*, "Pluto: Pathology-universal transformer," *arXiv preprint arXiv:2405.07905,* 2024.
[202] G. Wang, J. Wu, X. Luo, X. Liu, K. Li, and S. Zhang, "Mis-fm: 3d medical image segmentation using foundation models pretrained on a large-scale unannotated dataset," *arXiv preprint arXiv:2306.16925,* 2023.
[203] Y. Li, Y. Wu, Y. Lai, M. Hu, and X. Yang, "MedDINOv3: How to adapt vision foundation models for medical image segmentation?," *arXiv preprint arXiv:2509.02379,* 2025.
[204] Y. Gao, H. Li, F. Yuan, X. Wang, and X. Gao, "Dino U-Net: Exploiting High-Fidelity Dense Features from Foundation Models for Medical Image Segmentation," *arXiv preprint arXiv:2508.20909,* 2025.
[205] Y. Jiang and Y. Shen, "M4oe: A foundation model for medical multimodal image segmentation with mixture of experts," in *international conference on medical image computing and computer-assisted intervention*, 2024: Springer, pp. 621–631.
[206] Y. Chen *et al.*, "Modality-projection universal model for comprehensive full-body medical imaging segmentation," *Nature Communications,* vol. 16, no. 1, p. 9423, 2025/10/24 2025, doi: 10.1038/s41467-025-64469-w.
[207] S. Zhang *et al.*, "A generalist foundation model and database for open-world medical image segmentation," *Nature Biomedical Engineering,* 2025/09/05 2025, doi: 10.1038/s41551-025-01497-3.
[208] J. Xie, Z. Zhang, G. Luo, and Y. Zhu, "A Segmentation Foundation Model for Diverse-type Tumors," *arXiv preprint arXiv:2403.06396,* 2024.
[209] H. Peiris, S. Wang, G. Egan, M. Harandi, M. Law, and Z. Chen, "Segment Any Tumour: An Uncertainty-Aware Vision Foundation Model for Whole-Body Analysis," *arXiv preprint arXiv:2511.09592,* 2025.
[210] M. Rokuss *et al.*, "LesionLocator: Zero-Shot Universal Tumor Segmentation and Tracking in 3D Whole-Body Imaging," in *Proceedings of the Computer Vision and Pattern Recognition Conference*, 2025, pp. 30872–30885.

[211] Z. Yan *et al.*, "A foundation model for general moving object segmentation in medical images," in *2024 IEEE International Symposium on Biomedical Imaging (ISBI)*, 2024: IEEE, pp. 1–5.
[212] Z. Huang *et al.*, "Stu-net: Scalable and transferable medical image segmentation models empowered by large-scale supervised pre-training," *arXiv preprint arXiv:2304.06716,* 2023.
[213] X. Wu *et al.*, "ULS4US: universal lesion segmentation framework for 2D ultrasound images," *Physics in Medicine & Biology,* vol. 68, no. 16, p. 165009, 2023.
[214] H. Chen *et al.*, "Multi-organ foundation model for universal ultrasound image segmentation with task prompt and anatomical prior," *IEEE Transactions on Medical Imaging,* 2024.
[215] S. Sahar Taheri Otaghsara and R. Rahmanzadeh, "F3-Net: Foundation Model for Full Abnormality Segmentation of Medical Images with Flexible Input Modality Requirement," *arXiv e-prints,* p. arXiv: 2507.08460, 2025.
[216] X. Zhang *et al.*, "A foundation model for brain lesion segmentation with mixture of modality experts," in *International Conference on Medical Image Computing and Computer-Assisted Intervention*, 2024: Springer, pp. 379–389.
[217] Z. Zhang *et al.*, "Unimrisegnet: Universal 3d network for various organs and cancers segmentation on multi-sequence mri," *IEEE Journal of Biomedical and Health Informatics,* 2024.
[218] J. Cox *et al.*, "BrainSegFounder: Towards 3D foundation models for neuroimage segmentation," *Medical Image Analysis,* vol. 97, p. 103301, 2024.
[219] Y. Li *et al.*, "RoMedFormer: A Rotary-Embedding Transformer Foundation Model for 3D Genito-Pelvic Structure Segmentation in MRI and CT," *arXiv preprint arXiv:2503.14304,* 2025.
[220] Juwita, G. M. Hassan, and A. Datta, "3D-SCUMamba: An Abdominal Tumor Segmentation Model," *IEEE Access,* vol. 13, pp. 98362–98373, 2025, doi: 10.1109/ACCESS.2025.3573619.
[221] B. Wen *et al.*, "Universal Vessel Segmentation for Multi-Modality Retinal Images," *IEEE Transactions on Image Processing,* 2025.
[222] R. Deng *et al.*, "Prpseg: Universal proposition learning for panoramic renal pathology segmentation," in *Proceedings of the IEEE/CVF conference on computer vision and pattern recognition*, 2024, pp. 11736–11746.
[223] D. Ma, J. Pang, M. B. Gotway, and J. Liang, "A fully open AI foundation model applied to chest radiography," *Nature,* pp. 1–11, 2025.
[224] Y. Yoo *et al.*, "A Non-contrast Head CT Foundation Model for Comprehensive Neuro-Trauma Triage," in *International Conference on Medical Image Computing and Computer-Assisted Intervention*, 2025: Springer, pp. 3–13.
[225] W. Zhu *et al.*, "3D Foundation Model for Generalizable Disease Detection in Head Computed Tomography," *arXiv preprint arXiv:2502.02779,* 2025.
[226] R. Ding, H. Lu, and M. Liu, "DenseFormer-MoE: A Dense Transformer Foundation Model With Mixture of Experts for Multi-Task Brain Image Analysis," *IEEE Transactions on Medical Imaging,* vol. 44, no. 10, pp. 4037–4048, 2025, doi: 10.1109/TMI.2025.3551514.
[227] L. Shao *et al.*, "An MRI–pathology foundation model for noninvasive diagnosis and grading of prostate cancer," *Nature Cancer,* 2025/09/02 2025, doi: 10.1038/s43018-025-01041-x.
[228] P. F. Wilson *et al.*, "Prostnfound: Integrating foundation models with ultrasound domain knowledge and clinical context for robust prostate cancer detection," in *International Conference on Medical Image Computing and Computer-Assisted Intervention*, 2024: Springer, pp. 499–509.
[229] Q. Kang *et al.*, "URFM: a general Ultrasound Representation Foundation Model for advancing ultrasound image diagnosis," *IScience,* vol. 28, no. 8, 2025.

[230] L. Zedda, A. Loddo, and C. Di Ruberto, "Radio DINO: A foundation model for advanced radiomics and AI-driven medical imaging analysis," *Computers in Biology and Medicine,* vol. 195, p. 110583, 2025.

[231] H. Jiang *et al.*, "XFM: An Explainable Knowledge-Fused Vision Foundation Model for Improving Clinical Diagnosis of Low-Prevalence Retinal Diseases," *IEEE Journal of Biomedical and Health Informatics,* pp. 1–15, 2025, doi: 10.1109/JBHI.2025.3624305.

[232] R. J. Chen *et al.*, "Towards a general-purpose foundation model for computational pathology," *Nature Medicine,* vol. 30, no. 3, pp. 850–862, 2024/03/01 2024, doi: 10.1038/s41591-024-02857-3.

[233] E. Vorontsov *et al.*, "A foundation model for clinical-grade computational pathology and rare cancers detection," *Nature Medicine,* vol. 30, no. 10, pp. 2924–2935, 2024/10/01 2024, doi: 10.1038/s41591-024-03141-0.

[234] H. Xu *et al.*, "A whole-slide foundation model for digital pathology from real-world data," *Nature,* vol. 630, no. 8015, pp. 181–188, 2024.

[235] S. Ahmad Zafar *et al.*, "OpenSlideFM: A Computationally Efficient Multi-Scale Foundation Model for Computational Pathology," *medRxiv,* p. 2025.12. 09.25341769, 2025.

[236] T. Lazard, M. Lerousseau, E. Decencière, and T. Walter, "Giga-ssl: Self-supervised learning for gigapixel images," in *Proceedings of the IEEE/CVF Conference on Computer Vision and Pattern Recognition*, 2023, pp. 4305–4314.

[237] S. Hua, F. Yan, T. Shen, L. Ma, and X. Zhang, "PathoDuet: Foundation models for pathological slide analysis of H&E and IHC stains," *Medical Image Analysis,* vol. 97, p. 103289, 2024.

[238] L. Tian *et al.*, "unigradicon: A foundation model for medical image registration," in *International Conference on Medical Image Computing and Computer-Assisted Intervention*, 2024: Springer, pp. 749–760.

[239] L. Tian *et al.*, "GradICON: Approximate diffeomorphisms via gradient inverse consistency," in *Proceedings of the IEEE/CVF Conference on Computer Vision and Pattern Recognition*, 2023, pp. 18084–18094.

[240] B. Demir *et al.*, "Multigradicon: A foundation model for multimodal medical image registration," in *International Workshop on Biomedical Image Registration*, 2024: Springer, pp. 3–18.

[241] X. Song, X. Xu, J. Zhang, D. M. Reyes, and P. Yan, "Dino-Reg: Efficient Multimodal Image Registration with Distilled Features," *IEEE Transactions on Medical Imaging,* 2025.

[242] M. Yang, Q. Wang, S. Huang, W. Yan, Y. Hu, and Z. Min, "Det-SAMReg: Few-Shot Medical Image Registration Using Vision Foundation Models," Cham, 2026: Springer Nature Switzerland, in Simplifying Medical Ultrasound, pp. 76–86.

[243] S. Liu *et al.*, "Grounding dino: Marrying dino with grounded pre-training for open-set object detection," in *European conference on computer vision*, 2024: Springer, pp. 38–55.

[244] Z. Li *et al.*, "UniReg: Foundation Model for Controllable Medical Image Registration," *arXiv preprint arXiv:2503.12868,* 2025.

[245] P. Guo *et al.*, "Maisi: Medical ai for synthetic imaging," in *2025 IEEE/CVF Winter Conference on Applications of Computer Vision (WACV)*, 2025: IEEE, pp. 4430–4441.

[246] S. Sengupta, S. Chakrabarty, K. S. Ravi, G. Avinash, and R. Soni, "SynthFM: Training Modality-Agnostic Foundation Models for Medical Image Segmentation Without Real Medical Data," in *2025 IEEE 22nd International Symposium on Biomedical Imaging (ISBI)*, 2025: IEEE, pp. 1–4.

[247] J. Li *et al.*, "Topofm: Topology-guided pathology foundation model for high-resolution pathology image synthesis with cellular-level control," *IEEE Transactions on Medical Imaging,* 2025.

[248] Y. Ji *et al.*, "Amos: A large-scale abdominal multi-organ benchmark for versatile medical image segmentation," *Advances in neural information processing systems,* vol. 35, pp. 36722–36732, 2022.
[249] M. Antonelli *et al.*, "The medical segmentation decathlon," *Nature communications,* vol. 13, no. 1, p. 4128, 2022.
[250] R. DerSimonian and N. Laird, "Meta-analysis in clinical trials," *Controlled clinical trials,* vol. 7, no. 3, pp. 177–188, 1986.
[251] G. Schwarzer, "meta: An R package for meta-analysis," *R news,* vol. 7, no. 3, pp. 40–45, 2007.
[252] W. Viechtbauer and M. W. Viechtbauer, "Package 'metafor'," *The comprehensive R Archive network. Package 'metafor,* 2015.
[253] W. Viechtbauer, "Conducting meta-analyses in R with the metafor package," *Journal of statistical software,* vol. 36, pp. 1–48, 2010.
[254] S.-C. Huang, L. Shen, M. P. Lungren, and S. Yeung, "Gloria: A multimodal global-local representation learning framework for label-efficient medical image recognition," in *Proceedings of the IEEE/CVF international conference on computer vision*, 2021, pp. 3942–3951.
[255] P. Cheng, L. Lin, J. Lyu, Y. Huang, W. Luo, and X. Tang, "Prior: Prototype representation joint learning from medical images and reports," in *Proceedings of the IEEE/CVF international conference on computer vision*, 2023, pp. 21361–21371.
[256] H. Yang *et al.*, "A multimodal vision–language model for generalizable annotation-free pathology localization," *Nature Biomedical Engineering,* 2026/01/06 2026, doi: 10.1038/s41551-025-01574-7.
[257] G. Dawidowicz, E. Hirsch, and A. Tal, "Limitr: Leveraging local information for medical image-text representation," in *Proceedings of the IEEE/CVF International Conference on Computer Vision*, 2023, pp. 21165–21173.
[258] J. Liu *et al.*, "Mlip: Medical Language-Image Pre-Training With Masked Local Representation Learning," in *2024 IEEE International Symposium on Biomedical Imaging (ISBI)*, 27–30 May 2024 2024, pp. 1–5, doi: 10.1109/ISBI56570.2024.10635357.
[259] C. Liu, S. Cheng, M. Shi, A. Shah, W. Bai, and R. Arcucci, "IMITATE: Clinical Prior Guided Hierarchical Vision-Language Pre-Training," *IEEE Transactions on Medical Imaging,* vol. 44, no. 1, pp. 519–529, 2025, doi: 10.1109/TMI.2024.3449690.
[260] C. Wu, X. Zhang, Y. Zhang, Y. Wang, and W. Xie, "Medklip: Medical knowledge enhanced language-image pre-training for x-ray diagnosis," in *Proceedings of the IEEE/CVF international conference on computer vision*, 2023, pp. 21372–21383.
[261] H. Luo, Z. Zhou, C. Royer, A. Sekuboyina, and B. Menze, "Devide: Faceted medical knowledge for improved medical vision-language pre-training," *arXiv preprint arXiv:2404.03618,* 2024.
[262] J. Li, R. Selvaraju, A. Gotmare, S. Joty, C. Xiong, and S. C. H. Hoi, "Align before fuse: Vision and language representation learning with momentum distillation," *Advances in neural information processing systems,* vol. 34, pp. 9694–9705, 2021.
[263] Q. Deng *et al.*, "Grounded knowledge-enhanced medical vision–language pre-training for chest X-ray," *Biomedical Signal Processing and Control,* vol. 112, p. 108416, 2026/02/01/ 2026.
[264] R. Wang *et al.*, "ECAMP: Entity-centered Context-aware Medical Vision Language Pre-training," *Medical Image Analysis,* vol. 105, p. 103690, 2025/10/01/ 2025.
[265] Z. Zhao *et al.*, "One model to rule them all: Towards universal segmentation for medical images with text prompts," *arXiv preprint arXiv:2312.17183,* 2023.

[266] S. Zhang *et al.*, "A Multimodal Biomedical Foundation Model Trained from Fifteen Million Image–Text Pairs," *NEJM AI,* vol. 2, no. 1, p. AIoa2400640, 2025, doi: doi:10.1056/AIoa2400640.

[267] Y. Gu *et al.*, "Domain-specific language model pretraining for biomedical natural language processing," *ACM Transactions on Computing for Healthcare (HEALTH),* vol. 3, no. 1, pp. 1–23, 2021.

[268] M. U. Khattak, S. Kunhimon, M. Naseer, S. Khan, and F. S. Khan, "Unimed-clip: Towards a unified image-text pretraining paradigm for diverse medical imaging modalities," *arXiv preprint arXiv:2412.10372,* 2024.

[269] Z. Lu, H. Li, N. A. Parikh, J. R. Dillman, and L. He, "RadCLIP: Enhancing Radiologic Image Analysis Through Contrastive Language–Image Pretraining," *IEEE Transactions on Neural Networks and Learning Systems,* 2025.

[270] Y. Gao, S. Kim, D. E. Austin, and C. McIntosh, "MEDBind: Unifying Language and Multimodal Medical Data Embeddings," in *Medical Image Computing and Computer Assisted Intervention – MICCAI 2024*, Cham, M. G. Linguraru *et al.*, Eds., 2024// 2024: Springer Nature Switzerland, pp. 218–228.

[271] Z. Chen, S. Diao, B. Wang, G. Li, and X. Wan, "Towards unifying medical vision-and-language pre-training via soft prompts," in *Proceedings of the IEEE/CVF international conference on computer vision*, 2023, pp. 23403–23413.

[272] Z. Wan *et al.*, "Med-UniC: unifying cross-lingual medical vision-language pre-training by diminishing bias," presented at the Proceedings of the 37th International Conference on Neural Information Processing Systems, New Orleans, LA, USA, 2023.

[273] E. Tiu, E. Talius, P. Patel, C. P. Langlotz, A. Y. Ng, and P. Rajpurkar, "Expert-level detection of pathologies from unannotated chest X-ray images via self-supervised learning," *Nature Biomedical Engineering,* vol. 6, no. 12, pp. 1399–1406, 2022, doi: 10.1038/s41551-022-00936-9.

[274] A. E. W. Johnson *et al.*, "MIMIC-CXR, a de-identified publicly available database of chest radiographs with free-text reports," *Scientific Data,* vol. 6, no. 1, p. 317, 2019/12/12 2019, doi: 10.1038/s41597-019-0322-0.

[275] T. Dai, R. Zhang, F. Hong, J. Yao, Y. Zhang, and Y. Wang, "UniChest: Conquer-and-Divide Pre-Training for Multi-Source Chest X-Ray Classification," *IEEE Transactions on Medical Imaging,* vol. 43, no. 8, pp. 2901–2912, 2024, doi: 10.1109/tmi.2024.3381123.

[276] J. Irvin *et al.*, "Chexpert: A large chest radiograph dataset with uncertainty labels and expert comparison," in *Proceedings of the AAAI conference on artificial intelligence*, 2019, vol. 33, no. 01, pp. 590–597.

[277] X. Wang, Y. Peng, L. Lu, Z. Lu, M. Bagheri, and R. M. Summers, "Chestx-ray8: Hospital-scale chest x-ray database and benchmarks on weakly-supervised classification and localization of common thorax diseases," in *Proceedings of the IEEE conference on computer vision and pattern recognition*, 2017, pp. 2097–2106.

[278] H. Q. Nguyen *et al.*, "VinDr-CXR: An open dataset of chest X-rays with radiologist's annotations," *Scientific Data,* vol. 9, no. 1, p. 429, 2022/07/20 2022, doi: 10.1038/s41597-022-01498-w.

[279] C. Liu *et al.*, "M-flag: Medical vision-language pre-training with frozen language models and latent space geometry optimization," in *International Conference on Medical Image Computing and Computer-Assisted Intervention*, 2023: Springer, pp. 637–647.

[280] C. Raffel *et al.*, "Exploring the limits of transfer learning with a unified text-to-text transformer," *Journal of machine learning research,* vol. 21, no. 140, pp. 1–67, 2020.

[281] R. Anil *et al.*, "Palm 2 technical report," *arXiv preprint arXiv:2305.10403,* 2023.

[282] Y. Huang, P. Sharma, A. Palepu, N. Greenbaum, A. Beam, and K. Beam, "NeoCLIP: A Self-Supervised Foundation Model for the Interpretation of Neonatal Radiographs," *medRxiv,* p. 2024.12. 03.24318410, 2024.
[283] S. Ghosh, C. B. Poynton, S. Visweswaran, and K. Batmanghelich, "Mammo-clip: A vision language foundation model to enhance data efficiency and robustness in mammography," in *International conference on medical image computing and computer-assisted intervention*, 2024: Springer, pp. 632–642.
[284] I. E. Hamamci *et al.*, "Developing generalist foundation models from a multimodal dataset for 3d computed tomography," *arXiv preprint arXiv:2403.17834,* 2024.
[285] R. L. Draelos *et al.*, "Machine-learning-based multiple abnormality prediction with large-scale chest computed tomography volumes," *Medical image analysis,* vol. 67, p. 101857, 2021.
[286] J. Lin *et al.*, "Ct-glip: 3d grounded language-image pretraining with ct scans and radiology reports for full-body scenarios," *arXiv preprint arXiv:2404.15272,* 2024.
[287] K. Yamamoto and T. Kikuchi, "TotalFM: An Organ-Separated Framework for 3D-CT Vision Foundation Models," *arXiv preprint arXiv:2601.00260,* 2026.
[288] Y. Hu *et al.*, "Cardiac-CLIP: A Vision-Language Foundation Model for 3D Cardiac CT Images," *arXiv preprint arXiv:2507.22024,* 2025.
[289] L. Blankemeier *et al.*, "Merlin: a computed tomography vision–language foundation model and dataset," *Nature,* 2026/03/04 2026, doi: 10.1038/s41586-026-10181-8.
[290] Z. Yang *et al.*, "Decipher-MR: A Vision-Language Foundation Model for 3D MRI Representations," *arXiv preprint arXiv:2509.21249,* 2025.
[291] M. Christensen, M. Vukadinovic, N. Yuan, and D. Ouyang, "Vision–language foundation model for echocardiogram interpretation," *Nature Medicine,* vol. 30, no. 5, pp. 1481–1488, 2024, doi: 10.1038/s41591-024-02959-y.
[292] Z. Liu, H. Mao, C. Y. Wu, C. Feichtenhofer, T. Darrell, and S. Xie, "A ConvNet for the 2020s," in *2022 IEEE/CVF Conference on Computer Vision and Pattern Recognition (CVPR)*, 18–24 June 2022 2022, pp. 11966–11976, doi: 10.1109/CVPR52688.2022.01167.
[293] F. Maani *et al.*, "FetalCLIP: A visual-language foundation model for fetal ultrasound image analysis," *arXiv preprint arXiv:2502.14807,* 2025.
[294] A. Hurst *et al.*, "Gpt-4o system card," *arXiv preprint arXiv:2410.21276,* 2024.
[295] J. Silva-Rodríguez, H. Chakor, R. Kobbi, J. Dolz, and I. Ben Ayed, "A Foundation Language-Image Model of the Retina (FLAIR): encoding expert knowledge in text supervision," *Medical Image Analysis,* vol. 99, p. 103357, 2025, doi: 10.1016/j.media.2024.103357.
[296] J. Du *et al.*, "Ret-clip: A retinal image foundation model pre-trained with clinical diagnostic reports," in *International conference on medical image computing and computer-assisted intervention*, 2024: Springer, pp. 709–719.
[297] K. Yu *et al.*, "Urfound: Towards universal retinal foundation models via knowledge-guided masked modeling," in *International Conference on Medical Image Computing and Computer-Assisted Intervention*, 2024: Springer, pp. 753–762.
[298] Z. Li *et al.*, "VisionUnite: A Vision-Language Foundation Model for Ophthalmology Enhanced With Clinical Knowledge," *IEEE Transactions on Pattern Analysis and Machine Intelligence,* vol. 47, no. 12, pp. 11848–11862, 2025, doi: 10.1109/TPAMI.2025.3598734.
[299] OpenAI, "GPT-4V(ision) system card," 2023.
[300] M. Y. Lu *et al.*, "Visual Language Pretrained Multiple Instance Zero-Shot Transfer for Histopathology Images," in *2023 IEEE/CVF Conference on Computer Vision and Pattern Recognition (CVPR)*, 17–24 June 2023 2023, pp. 19764–19775, doi: 10.1109/CVPR52729.2023.01893.

[301] Z. Huang, F. Bianchi, M. Yuksekgonul, T. J. Montine, and J. Zou, "A visual–language foundation model for pathology image analysis using medical Twitter," *Nature Medicine,* vol. 29, no. 9, pp. 2307–2316, 2023/09/01 2023, doi: 10.1038/s41591-023-02504-3.
[302] C. Schuhmann *et al.*, "Laion-5b: An open large-scale dataset for training next generation image-text models," *Advances in neural information processing systems,* vol. 35, pp. 25278–25294, 2022.
[303] L. Wang, M. Wang, H. Fu, and D. Zhang, "Vision-language model IP protection via prompt-based learning," in *Proceedings of the Computer Vision and Pattern Recognition Conference*, 2025, pp. 9497–9506.
[304] S. Javed, A. Mahmood, I. I. Ganapathi, F. A. Dharejo, N. Werghi, and M. Bennamoun, "Cplip: Zero-shot learning for histopathology with comprehensive vision-language alignment," in *Proceedings of the IEEE/CVF conference on computer vision and pattern recognition*, 2024, pp. 11450–11459.
[305] J. Liu *et al.*, "Clip-driven universal model for organ segmentation and tumor detection," in *Proceedings of the IEEE/CVF international conference on computer vision*, 2023, pp. 21152–21164.
[306] X. Ma, J. Fu, W. Liao, S. Zhang, and G. Wang, "CLISC: Bridging Clip and SAM by Enhanced CAM for Unsupervised Brain Tumor Segmentation," in *2025 IEEE 22nd International Symposium on Biomedical Imaging (ISBI)*, 14–17 April 2025 2025, pp. 1–5, doi: 10.1109/ISBI60581.2025.10980784.
[307] B. D. Killeen *et al.*, "FluoroSAM: A Language-promptable Foundation Model for Flexible X-ray Image Segmentation," in *International Conference on Medical Image Computing and Computer-Assisted Intervention*, 2025: Springer, pp. 248–258.
[308] T. Koleilat, H. Asgariandehkordi, H. Rivaz, and Y. Xiao, "MedCLIP-SAM: Bridging Text and Image Towards Universal Medical Image Segmentation," in *Medical Image Computing and Computer Assisted Intervention – MICCAI 2024*, Cham, M. G. Linguraru *et al.*, Eds., 2024// 2024: Springer Nature Switzerland, pp. 643–653.
[309] T. Koleilat, H. Asgariandehkordi, H. Rivaz, and Y. Xiao, "Medclip-samv2: Towards universal text-driven medical image segmentation," *arXiv preprint arXiv:2409.19483,* 2024.
[310] Y. Li *et al.*, "OpenVocabCT: Towards Universal Text-driven CT Image Segmentation," *IEEE Transactions on Medical Imaging,* 2025.
[311] Z. Wan, Y. Gao, W. Pang, and D. Ding, "VOILA: Complexity-Aware Universal Segmentation of CT images by Voxel Interacting with Language," in *Proceedings of the AAAI Conference on Artificial Intelligence*, 2025, vol. 39, no. 7, pp. 7482–7490.
[312] R. Yuan, W. Zhang, X. Dong, and W. Zhang, "Crns: CLIP-driven referring nuclei segmentation," *The Journal of Supercomputing,* vol. 81, no. 1, p. 174, 2025.
[313] P. Liang *et al.*, "VISTA-PATH: An interactive foundation model for pathology image segmentation and quantitative analysis in computational pathology," *arXiv preprint arXiv:2601.16451,* 2026.
[314] N. C. Codella *et al.*, "Medimageinsight: An open-source embedding model for general domain medical imaging," *arXiv preprint arXiv:2410.06542,* 2024.
[315] T. Han *et al.*, "CLEAR: An Auditable Foundation Model for Radiology Grounded in Clinical Concepts," *medRxiv,* p. 2026.01. 15.26344222, 2026.
[316] C. Kim *et al.*, "Transparent medical image AI via an image–text foundation model grounded in medical literature," *Nature medicine,* vol. 30, no. 4, pp. 1154–1165, 2024.
[317] Y. Nie *et al.*, "Conceptclip: Towards trustworthy medical ai via concept-enhanced contrastive langauge-image pre-training," *arXiv e-prints,* p. arXiv: 2501.15579, 2025.
[318] J. Achiam *et al.*, "Gpt-4 technical report," *arXiv preprint arXiv:2303.08774,* 2023.
[319] W. Xu *et al.*, "Lingshu: A Generalist Foundation Model for Unified Multimodal Medical Understanding and Reasoning," *arXiv preprint arXiv:2506.07044,* 2025.

[320] H.-Y. Zhou, J. N. Acosta, S. Adithan, S. Datta, E. J. Topol, and P. Rajpurkar, "MedVersa: A Generalist Foundation Model for Diverse Medical Imaging Tasks," *NEJM AI,* vol. 0, no. 0, p. AIoa2500595, doi: doi:10.1056/AIoa2500595.

[321] X. Zhu, Y. Hu, F. Mo, M. Li, and J. Wu, "Uni-med: a unified medical generalist foundation model for multi-task learning via connector-MoE," *Advances in Neural Information Processing Systems,* vol. 37, pp. 81225–81256, 2024.

[322] T. Lin *et al.*, "HealthGPT: A Medical Large Vision-Language Model for Unifying Comprehension and Generation via Heterogeneous Knowledge Adaptation," presented at the Proceedings of the 42nd International Conference on Machine Learning, Proceedings of Machine Learning Research, 2025.

[323] T. Zhao *et al.*, "BiomedParse-V: Scaling Foundation Model for Universal Text-guided Volumetric Biomedical Image Segmentation," in *CVPR 2025: Foundation Models for 3D Biomedical Image Segmentation*, 2025.

[324] Y. Xin, G. C. Ates, K. Gong, and W. Shao, "Med3dvlm: An efficient vision-language model for 3d medical image analysis," *IEEE Journal of Biomedical and Health Informatics,* 2025.

[325] Y. Shi *et al.*, "Med-2E3: A 2D-Enhanced 3D Medical Multimodal Large Language Model," in *2025 IEEE International Conference on Bioinformatics and Biomedicine (BIBM)*, 15–18 Dec. 2025 2025, pp. 2754–2759, doi: 10.1109/BIBM66473.2025.11357048.

[326] S. Jiang *et al.*, "Omniv-med: Scaling medical vision-language model for universal visual understanding," *arXiv preprint arXiv:2504.14692,* 2025.

[327] L. Luo, B. Tang, X. Chen, R. Han, and T. Chen, "VividMed: Vision Language Model with Versatile Visual Grounding for Medicine," Albuquerque, New Mexico, April 2025: Association for Computational Linguistics, in Proceedings of the 2025 Conference of the Nations of the Americas Chapter of the Association for Computational Linguistics: Human Language Technologies (Volume 1: Long Papers), pp. 1800–1821.

[328] L. Wu *et al.*, "UniBiomed: A Universal Foundation Model for Grounded Biomedical Image Interpretation," *arXiv preprint arXiv:2504.21336,* 2025.

[329] Y. Chen *et al.*, "MIMO: A medical vision language model with visual referring multimodal input and pixel grounding multimodal output," in *Proceedings of the Computer Vision and Pattern Recognition Conference*, 2025, pp. 24732–24741.

[330] C. Li *et al.*, "RadDiagSeg-M: A Vision Language Model for Joint Diagnosis and Multi-Target Segmentation in Radiology," *arXiv preprint arXiv:2510.18188,* 2025.

[331] T. Zhao *et al.*, "A foundation model for joint segmentation, detection and recognition of biomedical objects across nine modalities," *Nature Methods,* vol. 22, no. 1, pp. 166–176, 2025/01/01 2025, doi: 10.1038/s41592-024-02499-w.

[332] X. Xiao *et al.*, "Describe Anything in Medical Images," *arXiv preprint arXiv:2505.05804,* 2025.

[333] Y. Wang *et al.*, "Toward general text-guided multimodal brain MRI synthesis for diagnosis and medical image analysis," *Cell Reports Medicine,* vol. 6, no. 6, p. 102182, 2025/06/17/ 2025.

[334] Y. Wang *et al.*, "Unisyn: A generative foundation model for universal medical image synthesis across mri, ct and pet," in *International Conference on Medical Image Computing and Computer-Assisted Intervention*, 2025: Springer, pp. 673–682.

[335] D. Molino *et al.*, "XGeM: A multi-prompt foundation model for multimodal medical data generation," *Computerized Medical Imaging and Graphics,* p. 102718, 2026.

[336] J. Mao *et al.*, "Medsegfactory: Text-guided generation of medical image-mask pairs," in *Proceedings of the IEEE/CVF International Conference on Computer Vision*, 2025, pp. 21525–21535.

[337] S. Lee, J. Youn, H. Kim, M. Kim, and S. H. Yoon, "CXR-LLaVA: a multimodal large language model for interpreting chest X-ray images," *European Radiology,* vol. 35, no. 7, pp. 4374–4386, 2025, doi: 10.1007/s00330-024-11339-6.
[338] H. Touvron *et al.*, "Llama 2: Open foundation and fine-tuned chat models," *arXiv preprint arXiv:2307.09288,* 2023.
[339] Z. Chen *et al.*, "Chexagent: Towards a foundation model for chest x-ray interpretation," *arXiv preprint arXiv:2401.12208,* 2024.
[340] S. Srivastav *et al.*, "MAIRA at RRG24: A specialised large multimodal model for radiology report generation," Bangkok, Thailand, August 2024: Association for Computational Linguistics, in Proceedings of the 23rd Workshop on Biomedical Natural Language Processing, pp. 597–602.
[341] O. C. Thawakar *et al.*, "XrayGPT: Chest Radiographs Summarization using Large Medical Vision-Language Models," Bangkok, Thailand, August 2024: Association for Computational Linguistics, in Proceedings of the 23rd Workshop on Biomedical Natural Language Processing, pp. 440–448.
[342] C. Pellegrini, E. Özsoy, B. Busam, B. Wiestler, N. Navab, and M. Keicher, "Radialog: Large vision-language models for x-ray reporting and dialog-driven assistance," in *Medical Imaging with Deep Learning*, 2025.
[343] A. Fallahpour, J. Ma, A. Munim, H. Lyu, and B. Wang, "Medrax: Medical reasoning agent for chest x-ray," *arXiv preprint arXiv:2502.02673,* 2025.
[344] Y. Xing *et al.*, "MedVL-SAM2: A unified 3D medical vision-language model for multimodal reasoning and prompt-driven segmentation," *arXiv preprint arXiv:2601.09879,* 2026.
[345] X. He *et al.*, "OmniMRI: A Unified Vision--Language Foundation Model for Generalist MRI Interpretation," *arXiv preprint arXiv:2508.17524,* 2025.
[346] X. Guo, W. Chai, S.-Y. Li, and G. Wang, "LLaVA-ultra: Large Chinese language and vision assistant for ultrasound," in *Proceedings of the 32nd ACM international conference on multimedia*, 2024, pp. 8845–8854.
[347] L. Berger *et al.*, "Context-Aware Vision Language Foundation Models for Ocular Disease Screening in Retinal Images," *arXiv preprint arXiv:2503.15212,* 2025.
[348] Z. Deng *et al.*, "OphGLM: An ophthalmology large language-and-vision assistant," *Artificial Intelligence in Medicine,* vol. 157, p. 103001, 2024/11/01/ 2024.
[349] Y. Sun *et al.*, "Cpath-omni: A unified multimodal foundation model for patch and whole slide image analysis in computational pathology," in *Proceedings of the Computer Vision and Pattern Recognition Conference*, 2025, pp. 10360–10371.
[350] J. Bai *et al.*, "Qwen technical report," *arXiv preprint arXiv:2309.16609,* 2023.
[351] T. Liu *et al.*, "Teampath: Building multimodal pathology experts with reasoning ai copilots," *arXiv preprint arXiv:2511.17652,* 2025.
[352] X. Liu *et al.*, "Vision-language foundation model for generalizable nasal disease diagnosis using unlabeled endoscopic records," *Pattern Recognition,* vol. 165, p. 111646, 2025.
[353] X. Han *et al.*, "Npc Model: A Multi-Task Solution for Nasopharyngeal Carcinoma Based on Pretraining and Large Language Models," *Available at SSRN 4882213.*
[354] Z. Wang, M. Fang, L. Tang, J. Tian, and D. Dong, "HiCur-NPC: Hierarchical Feature Fusion Curriculum Learning for Multi-Modal Foundation Model in Nasopharyngeal Carcinoma," *IEEE Transactions on Medical Imaging,* 2025.
[355] J. Rückert *et al.*, "Rocov2: Radiology objects in context version 2, an updated multimodal image dataset," *Scientific Data,* vol. 11, no. 1, p. 688, 2024.
[356] X. He *et al.*, "Towards Visual Question Answering on Pathology Images," Online, August 2021: Association for Computational Linguistics, pp. 708–718.

[357] Y. Ye, J. Zhang, Z. Chen, and Y. Xia, "Desd: Self-supervised learning with deep self-distillation for 3d medical image segmentation," in *International Conference on Medical Image Computing and Computer-Assisted Intervention*, 2022: Springer, pp. 545–555.
[358] D. MH Nguyen *et al.*, "Lvm-med: Learning large-scale self-supervised vision models for medical imaging via second-order graph matching," *Advances in Neural Information Processing Systems,* vol. 36, pp. 27922–27950, 2023.
[359] X. He *et al.*, "Matchanything: Universal cross-modality image matching with large-scale pre-training," *arXiv preprint arXiv:2501.07556,* 2025.
[360] Y. Xie, J. Zhang, Y. Xia, and Q. Wu, "Unimiss: Universal medical self-supervised learning via breaking dimensionality barrier," in *European Conference on Computer Vision*, 2022: Springer, pp. 558–575.
[361] W. Tian *et al.*, "MOSMOS: Multi-organ segmentation facilitated by medical report supervision," *Biomedical Signal Processing and Control,* vol. 106, p. 107743, 2025, doi: 10.1016/j.bspc.2025.107743.
[362] S. Graham *et al.*, "Hover-Net: Simultaneous segmentation and classification of nuclei in multi-tissue histology images," *Medical Image Analysis,* vol. 58, p. 101563, 2019/12/01/ 2019.
[363] H. Abdi and L. J. Williams, "Principal component analysis," *Wiley interdisciplinary reviews: computational statistics,* vol. 2, no. 4, pp. 433–459, 2010.
[364] G. C. Linderman and S. Steinerberger, "Clustering with t-SNE, provably," *SIAM journal on mathematics of data science,* vol. 1, no. 2, pp. 313–332, 2019.
[365] L. McInnes, J. Healy, N. Saul, and L. Großberger, "UMAP: Uniform Manifold Approximation and Projection," *Journal of Open Source Software,* vol. 3, no. 29, p. 861, 2018 2018, doi: 10.21105/joss.00861.
[366] S. Woerner and C. F. Baumgartner, "Navigating data scarcity using foundation models: A benchmark of few-shot and zero-shot learning approaches in medical imaging," in *International Workshop on Foundation Models for General Medical AI*, 2024: Springer, pp. 30–39.
[367] A. Sekuboyina *et al.*, "VerSe: a vertebrae labelling and segmentation benchmark for multi-detector CT images," *Medical image analysis,* vol. 73, p. 102166, 2021.
[368] J. G. de Almeida *et al.*, "Foundation models for radiology—the position of the AI for Health Imaging (AI4HI) network," *Insights into Imaging,* vol. 16, no. 1, p. 168, 2025.
[369] N. Sourlos *et al.*, "Recommendations for the creation of benchmark datasets for reproducible artificial intelligence in radiology," *Insights into Imaging,* vol. 15, no. 1, p. 248, 2024.
[370] Q. Deng *et al.*, "Grounded Knowledge-Enhanced Medical VLP for Chest X-Ray," *CoRR,* vol. abs/2404.14750, 2024.
[371] X. Zhang, C. Wu, Y. Zhang, W. Xie, and Y. Wang, "Knowledge-enhanced visual-language pre-training on chest radiology images," *Nature Communications,* vol. 14, no. 1, p. 4542, 2023.
[372] L. Blankemeier *et al.*, "Merlin: A vision language foundation model for 3d computed tomography," *Research Square,* pp. rs. 3. rs–4546309, 2024.
[373] R. R. Selvaraju, M. Cogswell, A. Das, R. Vedantam, D. Parikh, and D. Batra, "Grad-cam: Visual explanations from deep networks via gradient-based localization," in *Proceedings of the IEEE international conference on computer vision*, 2017, pp. 618–626.
[374] J. Wu *et al.*, "Vision-language foundation model for 3D medical imaging," *npj Artificial Intelligence,* vol. 1, no. 1, p. 17, 2025.
[375] T. Tu *et al.*, "Towards generalist biomedical AI," *Nejm Ai,* vol. 1, no. 3, p. AIoa2300138, 2024.
[376] X. Wang *et al.*, "Openmedlab: An open-source platform for multi-modality foundation models in medicine," *arXiv preprint arXiv:2402.18028,* 2024.

[377] J. Gu *et al.*, "A survey on LLM-as-a-Judge," *The Innovation*, doi: 10.1016/j.xinn.2025.101253.

[378] H. Lovenia, W. Dai, S. Cahyawijaya, Z. Ji, and P. Fung, "Negative object presence evaluation (nope) to measure object hallucination in vision-language models," in *Proceedings of the 3rd Workshop on Advances in Language and Vision Research (ALVR)*, 2024, pp. 37–58.

[379] M. O. Khan, M. M. Afzal, S. Mirza, and Y. Fang, "How fair are medical imaging foundation models?," in *Machine Learning for Health (ML4H)*, 2023: PMLR, pp. 217–231.

[380] Q. Li *et al.*, "An empirical study on the fairness of foundation models for multi-organ image segmentation," in *International Conference on Medical Image Computing and Computer-Assisted Intervention*, 2024: Springer, pp. 432–442.

[381] B. Glocker, C. Jones, M. Roschewitz, and S. Winzeck, "Risk of bias in chest radiography deep learning foundation models," *Radiology: Artificial Intelligence,* vol. 5, no. 6, p. e230060, 2023.

[382] R. J. Chen *et al.*, "Algorithmic fairness in artificial intelligence for medicine and healthcare," *Nature biomedical engineering,* vol. 7, no. 6, pp. 719–742, 2023.

[383] Z. Xu, J. Li, Q. Yao, H. Li, M. Zhao, and S. K. Zhou, "Addressing fairness issues in deep learning-based medical image analysis: a systematic review," *npj Digital Medicine,* vol. 7, no. 1, p. 286, 2024.

[384] R. Geirhos *et al.*, "Shortcut learning in deep neural networks," *Nature Machine Intelligence,* vol. 2, no. 11, pp. 665–673, 2020.

[385] R. Jin *et al.*, "Fairmedfm: fairness benchmarking for medical imaging foundation models," *Advances in Neural Information Processing Systems,* vol. 37, pp. 111318–111357, 2024.

[386] A. Ade-Ibijola and C. Okonkwo, "Artificial intelligence in Africa: Emerging challenges," in *Responsible AI in Africa: Challenges and opportunities*: Springer International Publishing Cham, 2023, pp. 101–117.

[387] I. Y. Chen, E. Pierson, S. Rose, S. Joshi, K. Ferryman, and M. Ghassemi, "Ethical machine learning in healthcare," *Annual review of biomedical data science,* vol. 4, no. 1, pp. 123–144, 2021.

[388] F. EUROPEAN COMMISSION, "Regulation of the European parliament and of the council laying down harmonised rules on artificial intelligence (Artificial Intelligence Act) and amending certain union legislative acts," ed, 2021.

[389] B. Vasey *et al.*, "Reporting guideline for the early stage clinical evaluation of decision support systems driven by artificial intelligence: DECIDE-AI," *bmj,* vol. 377, 2022.

[390] G. S. Collins *et al.*, "TRIPOD+ AI statement: updated guidance for reporting clinical prediction models that use regression or machine learning methods," *bmj,* vol. 385, 2024.

[391] J. Mongan, L. Moy, and C. E. Kahn Jr, "Checklist for artificial intelligence in medical imaging (CLAIM): a guide for authors and reviewers," vol. 2, ed: Radiological Society of North America, 2020, p. e200029.

**Table 1.** Vision Foundation Models (VFMs) for Medical Imaging.

*Abbreviations: Seg = Segmentation, Cls = Classification, Det = Detection, Reg = Registration, Gen = Generation, Pred = Prediction, Enh = Enhancement, Ret = Retrieval, Track = Tracking, Loc = Localization, VQA = Visual Question Answering, Dim. = Dimensionality.*

| Study | Model | Backbone | Modality | Tasks | Dim. | Repository |
|---|---|---|---|---|---|---|
| ***Foundational SAM Adaptations*** | | | | | | |
| Gong et al., 2023 | 3DSAM-Adapter | ViT-B (SAM) | CT | Seg | 3D | https://github.com/med-air/3DSAM-adapter |
| Yan et al., 2024a | AFTer-SAM | ViT-B (SAM) | CT | Seg | 3D | — |
| Huang et al., 2026 | ECT-3DMedSAM | Hiera (SAM 2) | Multi-modal | Seg | 3D | — |
| Chen et al., 2024a | MA-SAM | ViT-B (SAM) | Multi-modal | Seg | 3D | https://github.com/cchen-cc/MA-SAM |
| Wu et al., 2025 | Med-SA | ViT-B (SAM) | Multi-modal | Seg | 2D/3D | https://github.com/KidsWithTokens/Medical-SAM-Adapter |
| Lei et al., 2025 | MedLSAM | ViT-B (SAM) | CT | Det, Seg | 3D | https://github.com/openmedlab/MedLSAM |
| Ma et al., 2024 | MedSAM | ViT-B (SAM) | Multi-modal | Seg | 2D | https://github.com/bowang-lab/MedSAM |
| Zhu et al., 2024 | MedSAM-2 | Hiera (SAM 2) | Multi-modal | Seg | 2D/3D | https://github.com/supermedintel/Medical-SAM2 |
| Sahay et al., 2025 | MoPEFT | ViT-B (SAM) | Multi-modal | Seg | 2D | — |
| Shen et al., 2025 | ProtoSAM-3D | ViT-B (SAM) | Multi-modal | Seg | 3D | — |
| Cheng et al., 2023 | SAM-Med2D | ViT-B (SAM) | Multi-modal | Seg | 2D | https://github.com/uni-medical/SAM-Med2D |
| Wang et al., 2025b | SAM-Med3D | ViT-B (SAM) | Multi-modal | Seg | 3D | https://github.com/uni-medical/SAM-Med3D |
| Zhang et al., 2023 | SAMed | ViT-B (SAM) | CT | Seg | 2D | https://github.com/hitachinsk/SAMed |
| ***Prompt Automation & Optimization SAM*** | | | | | | |
| Li et al., 2025a | AutoProSAM | ViT-B (SAM) | CT | Seg | 3D | — |
| Shaharabany et al., 2023 | AutoSAM | ViT-B (SAM) | Multi-modal | Seg | 2D | — |
| Guo et al., 2024 | ClickSAM | ViT-B (SAM) | Ultrasound | Seg | 2D | — |
| Xu et al., 2024 | ESP-MedSAM | ViT-B (SAM) | Multi-modal | Seg | 2D | https://github.com/xq141839/ESP-MedSAM |
| Xu et al., 2024 | EviPrompt | ViT-B (SAM) | Multi-modal | Seg | 2D | https://github.com/SPIresearch/EviPrompt |
| Ayzenberg et al., 2025 | ProtoSAM | DINOv2 + ViT-B (SAM) | Multi-modal | Seg | 2D | https://github.com/levayz/ProtoSAM |
| Xie et al., 2025 | RFMedSAM 2 | Hiera (SAM 2) | CT | Seg | 3D | — |
| Wang et al., 2025a | RRL-MedSAM | ViT-B (SAM) | Multi-modal | Reg, Seg | 3D | — |
| Deng et al., 2023 | SAM-U | ViT-B (SAM) | Ophthalmology | Seg | 2D | — |
| Xing et al., 2025 | SAM2-SGP | Hiera (SAM 2) | Multi-modal | Seg | 2D/3D | https://github.com/astlian9/SAM_Support |
| Wahd et al., 2025 | Sam2Rad | Hiera (SAM 2) | Ultrasound | Det, Seg | 2D | https://github.com/aswahd/SamRadiology |
| Towle et al., 2024 | SimSAM | ViT-B (SAM) | Multi-modal | Seg | 2D | https://github.com/BenjaminTowle/SimSAM |
| Zhu et al., 2025a | SSS | Hiera (SAM 2) | Multi-modal | Seg | 2D/3D | https://github.com/AIGeeksGroup/SSS |
| Pandey et al., 2023 | YOLOv8 + SAM | YOLOv8 + ViT-B (SAM) | Multi-modal | Det, Seg | 2D | — |
| Dai et al., 2025 | Zeus | ViT-B (MedCLIP + SAM) | Multi-modal | Seg | 2D | — |
| ***Hybrid SAM Architectures*** | | | | | | |
| Anand et al., 2023 | DINOv2 + SAM | DINOv2 + ViT-B (SAM) | Multi-modal | Det, Seg | 2D | — |
| Tian et al., 2025 | MedSAM-CA | ViT-B (SAM) + CBR-Net | Multi-modal | Seg | 2D | https://github.com/Tianpeiting/MedSAM-CA |
| Li et al., 2024 | ProMISe | ViT-B (SAM) + CNN | CT | Seg | 3D | https://github.com/MedICL-VU/ProMISe |
| Chai et al., 2024 | SAM-LST | ViT-B (SAM) + ResNet-18 | CT | Seg | 2D | https://github.com/11yxk/SAM-LST |
| Yang et al., 2024 | SAM-UNet-50 | ViT-B (SAM) + ResNet-50 | Multi-modal | Seg | 2D | https://github.com/Hhankyangg/sam-unet |
| Chen et al., 2025 | SAM2Med3D | Hiera (SAM 2) + nnU-Net | MRI | Seg | 3D | — |
| Wang et al., 2025 | SAMDA | ViT-B (MedSAM) + U-Net | Multi-modal | Seg | 2D | — |
| Chen et al., 2025 | SLM-SAM 2 | Hiera (SAM 2) + LSTM | Multi-modal | Seg | 3D | https://github.com/mazurowski-lab/SLM-SAM2 |
| Li et al., 2025b | TAGS | ViT-B (SAM) + CLIP | CT | Seg | 3D | https://github.com/sirileeee/TAGS |
| ***Efficient & Lightweight SAM*** | | | | | | |
| Pfefferle et al., 2024 | DAFT | EfficientViT | Multi-modal | Seg | 2D/3D | https://github.com/automl/CVPR24-MedSAM-on-Laptop |
| Xu et al., 2025 | De-LightSAM | DC-Encoder | Multi-modal | Seg | 2D | https://github.com/xq141839/De-LightSAM |
| Shen et al., 2024 | FastSAM3D | ViT-Tiny | Multi-modal | Seg | 3D | https://github.com/arcadelab/FastSAM3D |
| Liu et al., 2024 | LiteMedSAM + LoRA | TinyViT | Multi-modal | Seg | 2D/3D | https://github.com/lseventeen/SAMIL |
| Lyu et al., 2024 | MCP-MedSAM | TinyViT | Multi-modal | Seg | 2D | https://github.com/dong845/MCP-MedSAM |
| Luo et al., 2024 | Med-FastSAM | ViT-B (SAM) | Multi-modal | Seg | 2D | https://github.com/GalacticHogrider/Med-FastSAM |
| Le et al., 2024 | MedficientSAM | EfficientViT | Multi-modal | Seg | 2D/3D | https://github.com/hieplpvip/medficientsam |
| Zhang et al., 2024 | RepMedSAM | RepViT | Multi-modal | Seg | 2D/3D | — |
| Ali et al., 2024 | RepViT-MedSAM | RepViT | Multi-modal | Seg | 2D/3D | https://github.com/icecap360/TurboMedSAM |
| Kong et al., 2024 | SwiftMedSAM | TinyViT | Multi-modal | Seg | 2D/3D | — |
| Gao et al., 2024 | Swin-LiteMedSAM | Swin Transformer | Multi-modal | Seg | 2D/3D | https://github.com/RuochenGao/Swin_LiteMedSAM |
| ***Domain-Specific SAM*** | | | | | | |
| Paranjape et al., 2024 | AdaptiveSAM | ViT-B (SAM) | Multi-modal | Seg | 2D | https://github.com/JayParanjape/biastuning |
| Gowda et al., 2025 | CC-SAM | ViT-B (SAM) + ResNet-50 | Ultrasound | Seg | 2D | — |
| Marks et al., 2025 | CellSAM | ViT-B (SAM) | Pathology | Seg | 2D | https://github.com/vanvalenlab/cellsam |
| Deng et al., 2024 | MemSAM | ViT-B (SAM) | Ultrasound | Det, Seg | 2D | https://github.com/dengxl0520/MemSAM |
| Zhang et al., 2023 | SAM-Path | ViT-B (SAM) | Pathology | Seg | 2D | https://github.com/cvlab-stonybrook/SAMPath |
| Matasyoh et al., 2024 | SAMSurg | ViT-B (SAM) | Endoscopy | Seg | 2D | — |
| Lin et al., 2024 | SAMUS | ViT-B (SAM) + CNN | Ultrasound | Seg | 2D | https://github.com/xianlin7/SAMUS |
| Wang et al., 2024 | SegAnyPath | ViT-B (SAM) | Pathology | Seg | 2D | https://github.com/wagnchogn/SegAnyPath |
| Ravishankar et al., 2023 | SonoSAM | ViT-B (SAM) | Ultrasound | Seg | 2D/3D | — |
| Sheng et al., 2024 | Surgical-DeSAM | Swin-B + DETR | Endoscopy | Seg | 2D | https://github.com/YuyangSheng/Surgical-DeSAM |
| Yue et al., 2023 | SurgicalPart-SAM | ViT-B (SAM) | Endoscopy | Seg | 2D | https://github.com/wenxi-yue/SurgicalPart-SAM |
| Yue et al., 2024 | SurgicalSAM | ViT-B (SAM) | Endoscopy | Seg | 2D | https://github.com/wenxi-yue/SurgicalSAM |
| Kamtam et al., 2025 | SurgiSAM 2 | Hiera (SAM 2) | Endoscopy | Seg, Track | 2D | https://github.com/Devanish31/SurgiSAM2 |
| Sengupta et al., 2025 | SynthFM | ViT-B (SAM) | Multi-modal | Seg | 2D | — |
| Chen et al., 2024 | UN-SAM | ViT-B (SAM) | Pathology | Seg | 2D | https://github.com/CUHK-AIM-Group/UN-SAM |
| Liu et al., 2024 | WSI-SAM | TinyViT | Pathology | Seg | 2D | https://github.com/HongLiuuuuu/WSI-SAM |
| Archit et al., 2025 | μSAM | ViT-B (SAM) | Multi-modal | Seg, Track | 2D/3D | https://github.com/computational-cell-analytics/micro-sam |
| ***Task-Agnostic VFMs*** | | | | | | |
| Zhang et al., 2025 | BDFM | ViT-B + Swin-T | MRI | Cls, Seg | 3D | https://github.com/zzzjjj98/BDFM |
| Yang et al., 2025 | BEPH | ViT-B | Pathology | Cls, Pred | 2D | https://github.com/Zhcyoung/BEPH |
| Sun et al., 2025 | BME-X | DU-Net | MRI | Cls, Enh, Reg, Seg | 3D | https://github.com/DBC-Lab/Brain_MRI_Enhancement |
| Yang et al., 2025 | CheXFound | ViT-L | X-ray | Cls, Det, Pred, Seg | 2D | https://github.com/RPIDIAL/CheXFound |
| Wang et al., 2024 | CHIEF | CTransPath | Pathology | Cls, Det, Pred | 2D | https://github.com/hms-dbmi/CHIEF |
| Fu et al., 2025 | CineMA | ConvMAE | MRI | Cls, Pred, Seg | 2D/3D | https://github.com/mathpluscode/CineMA |
| Pai et al., 2025 | CT-FM | SegResNet | CT | Cls, Ret, Seg | 3D | https://aim.hms.harvard.edu/ct-fm |
| Dancette et al., 2025 | Curia-L | DINOv2 | Multi-modal | Cls, Det, Pred, Reg | 2D | https://huggingface.co/raidium/curia |
| Xu et al., 2024 | CXRBase | ViT-L | X-ray | Cls, Det | 2D | https://github.com/MedHK23/MAE_DR |
| Yao et al., 2025 | EVA-X | ViT-B | X-ray | Cls, Det, Seg | 2D | https://github.com/hustvl/EVA-X |
| Shi et al., 2024 | EyeFound | ViT-L | Multi-modal | Cls, Pred, VQA | 2D | — |
| Gao et al., 2025 | LCTfound | U-Net + Transformer | CT | Cls, Enh, Pred, Seg | 2D | https://github.com/gingerbread000/LCTfound |
| Jun et al., 2023 | Medical Transformer | Transformer | MRI | Cls, Reg, Seg | 2.5D | https://github.com/ku-milab/MedicalTransformer |
| Zhou et al., 2025 | MerMED-FM | ViT-B | Multi-modal | Cls | 2D | — |
| Dong et al., 2025 | MRI-CORE | ViT-B | MRI | Cls, Seg | 2D | https://github.com/mazurowski-lab/mri_foundation |
| Zheng et al., 2025 | OpenUS | VMamba-S | Ultrasound | Cls, Det, Seg | 2D | https://github.com/XZheng0427/OpenUS |
| Juyal et al., 2024 | PLUTO | ViT-S | Pathology | Cls, Seg, Pred | 2D | — |
| Liu et al., 2025 | RadLAS | ViT-B | X-ray | Cls, Loc | 2D | — |
| Moutakanni et al., 2024 | RayDINO | ViT-L | X-ray | Cls, Gen, Seg | 2D | https://github.com/facebookresearch/dinov2 |
| Ding et al., 2025b | TITAN | ViT-B | Pathology | Cls, Gen, Pred, Ret | 2D | https://github.com/mahmoodlab/TITAN |
| Jiao et al., 2024 | USFM | ViT-B | Ultrasound | Cls, Enh, Seg | 2D | — |
| Qiu et al., 2024 | VisionFM | Transformer | Ophthalmology | Cls, Det, Pred, Seg | 2D/3D | — |
| Yan et al., 2025 | PathOrchestra | ViT-L | Pathology | Cls, Seg, Det, Pred | 2D | https://github.com/yanfang-research/PathOrchestra |
| Zhou et al., 2023 | RETFound | ViT-L | Multi-modal | Cls, Pred | 2D | https://github.com/rmaphoh/RETFound_MAE |
| Pohjonen et al., 2024 | HistoEncoder | XCiT-S12/M24 | Pathology | Cls, Pred | 2D | https://github.com/jopo666/HistoEncoder |
| ***Segmentation VFMs (Non-SAM)*** | | | | | | |
| Juwita et al., 2025 | 3D-SCUMamba | 3D CNN + Mamba | CT | Seg | 3D | https://github.com/juwita-sj/3DSCU-Mamba |
| Cox et al., 2024 | BrainSegFounder | Swin-T | MRI | Seg | 3D | https://github.com/lab-smile/BrainSegFounder |
| Gao et al., 2025 | Dino U-Net | DINOv2 | Multi-modal | Seg | 2D | https://github.com/yifangao112/DinoUNet |
| Taheri Otaghsara et al., 2025 | F3-Net | 3D CNN / nnU-Net | MRI | Seg | 3D | — |
| Yan et al., 2024b | iMOS | ResNet-like | Multi-modal | Seg, Track | 2D/3D | https://github.com/yzhn16/iMOS |
| Rokuss et al., 2025 | LesionLocator | 3D ResNet U-Net | Multi-modal | Seg, Track | 3D | https://github.com/MIC-DKFZ/LesionLocator |
| Jiang et al., 2025a | M4oE | Swin-T + $M^4$oE | Multi-modal | Seg | 2D | https://github.com/JefferyJiang-YF/M4oE |
| Li et al., 2025b | MedDINOv2 | ViT-B | CT | Seg | 2D | https://github.com/ricklisz/MedDINOv3 |
| Zhang et al., 2025 | MedSegX | ViT-B (SAM) | Multi-modal | Seg | 2D/3D | https://github.com/MedSegX/MedSegX-code |
| Wang et al., 2023 | MIS-FM | CNN + 3D Swin-T | CT | Seg | 3D | https://github.com/openmedlab/MIS-FM |
| Chen et al., 2024 | MOFO | CSWin-T | Ultrasound | Seg | 2D | https://github.com/Git-HB-CHEN/MOFO |
| Zhang et al., 2024 | MoME | nnU-Net | MRI | Seg | 3D | https://github.com/ZhangxinruBIT/MoME |
| Chen et al., 2025 | MPUM | Custom 3D CNN | Multi-modal | Seg | 3D | https://github.com/YixinChen-AI/MPUM |
| Deng et al., 2024 | PrPSeg | ResU-Net | Pathology | Seg | 2D | — |
| Li et al., 2025a | RoMedFormer | 3D ViT | CT | Seg | 3D | — |
| Peiris et al., 2025 | SAT3D | Swin-T | Multi-modal | Seg | 3D | https://github.com/himashi92/SAT3D |
| Huang et al., 2023 | STU-Net | 3D U-Net | Multi-modal | Seg | 3D | https://github.com/Ziyan-Huang/STU-Net |
| Xie et al., 2024 | TSFM | ResBlock + ViT-B | Multi-modal | Seg | 3D | — |
| Wu et al., 2023 | ULS4US | U-Net | Ultrasound | Seg | 2D | https://github.com/cakuba/ULS4US |
| Zhang et al., 2024 | UniMRISegNet | 3D CNN + CLIP | MRI | Seg | 3D | — |
| Butoi et al., 2023 | UniverSeg | CNN U-Net | Multi-modal | Seg | 2D | https://universeg.csail.mit.edu |
| Wen et al., 2025 | URVSM | U-Net | Ophthalmology | Seg | 2D | https://github.com/JRC-VPLab/URVSM |
| Liu et al., 2025 | VIS-MAE-Modality | Swin-T | Multi-modal | Seg | 2D | https://github.com/lzl199704/VIS-MAE |
| ***Classification VFMs*** | | | | | | |
| Ma et al., 2025 | Ark+ | Swin-L | X-ray | Cls | 2D | https://github.com/jlianglab/Ark |
| Tak et al., 2024 | BrainIAC | ResNet-50 | MRI | Cls, Pred | 3D | https://huggingface.co/Divytak/brainiac |
| Yoo et al., 2025 | DeepCNTD-Net | 3D Dense U-Net | CT | Cls | 3D | — |
| Ding et al., 2025a | DenseFormer-MoE | DenseFormer | MRI | Cls | 3D | — |
| Zhu et al., 2025b | FM-HCT | ViT-B (3D) | CT | Det | 3D | https://github.com/NYUMedML/headCT_foundation |
| Lazard et al., 2023 | Giga-SSL | ResNet-18 | Pathology | Cls | 2D | — |
| Shao et al., 2025 | MRI-PTPCa | MobileNetV3 + ViT | MRI | Cls, Pred | 3D | https://github.com/StandWisdom/MRI-based-Predicted-Transformer-for-Prostate-cancer |
| Zafar et al., 2025 | OpenSlideFM | ConvNeXt-Tiny | Pathology | Cls, Det | 2D | https://github.com/Sjtu-Fuxilab/OpenSlideFM |
| Hua et al., 2024 | PathoDuet | ViT-B/16 | Pathology | Cls | 2D | https://github.com/openmedlab/PathoDuet |
| Wilson et al., 2024 | ProstNFound+ | ViT-B (MedSAM) | Ultrasound | Cls, Det | 2D | https://github.com/pfrwilson/prostNfound |
| Xu et al., 2024 | Prov-GigaPath | ViT + LongNet | Pathology | Cls | 2D | https://github.com/prov-gigapath/prov-gigapath |
| Zedda et al., 2025 | Radio-DINO | ViT-B | Multi-modal | Cls | 2D | https://github.com/Snarci/Radio-DINO |
| Chen et al., 2024b | UNI | ViT-L | Pathology | Cls, Det | 2D | https://github.com/mahmoodlab/UNI |
| Kang et al., 2025 | URFM | ViT-B | Ultrasound | Cls | 2D | — |
| Vorontsov et al., 2024 | Virchow | ViT-H | Pathology | Cls, Pred | 2D | https://huggingface.co/paige-ai/Virchow |
| Jiang et al., 2025b | XFM | ViT-B | Fundus | Cls | 2D | https://github.com/neugmd/Foundation_XAI |
| ***Image Registration VFMs*** | | | | | | |
| Yang et al., 2026 | Det-SAMReg | ViT (DINO) + ViT (SAM 2) | Multi-modal | Det, Reg, Seg | 3D | https://github.com/ymttt111/Det-SAMReg |
| Song et al., 2025 | DINO-Reg | ViT-L (DINOv2) | Multi-modal | Reg | 3D | https://github.com/RPIDIAL/DINO-Reg |
| Demir et al., 2024 | multiGradICON | U-Net | Multi-modal | Reg | 3D | https://github.com/uncbiag/uniGradICON |
| Tian et al., 2024 | uniGradICON | U-Net | Mono-modal | Reg | 3D | https://github.com/uncbiag/uniGradICON |
| Li et al., 2025d | UniReg | ViT-B (SAM) | CT | Reg | 3D | — |
| ***Image Synthesis & Generation VFMs*** | | | | | | |
| Guo et al., 2025 | MAISI | 3D CNN VAE | Multi-modal | Gen, Seg | 3D | — |
| Yu et al., 2025 | MedDiff-FM | U-Net | CT | Det, Gen, Enh | 3D | — |
| Li et al., 2025c | ToPoFM | DINO ViT-S + U-Net | Pathology | Cls, Gen, Seg | 2D | — |

**Table 2.** Vision-Language Foundation Models (VLFMs) for Medical Imaging.

*Abbreviations: Seg = Segmentation, Cls = Classification, Det = Detection, Ret = Retrieval, VQA = Visual Question Answering, RepGen = Report Generation, Cap = Captioning, Loc = Localization, Dx = Diagnosis, Synth = Image Synthesis, Sum = Summarization, Dim. = Dimensionality.*

| Study | Model | Image Encoder | Text Encoder | Modality | Tasks | Dim. | Repository |
|---|---|---|---|---|---|---|---|
| ***Contrastive VLFMs*** | | | | | | | |
| Yang et al., 2026 | AFLoc | ResNet-50 | BioClinicalBERT | Multi-modal | Loc, Cls | 2D | https://github.com/YH0517/AFLoc |
| Xiao et al., 2024 | ASIMSA | ResNet-50 | BioClinicalBERT | X-ray | Cls, Seg | 2D | — |
| Zhang et al., 2025 | BiomedCLIP | ViT-B | PubMedBERT | Multi-modal | Ret, Cls, VQA | 2D | https://aka.ms/biomedclip |
| Bannur et al., 2023 | BioViL-T | CNN-Transformer | CXR-BERT | X-ray | Cls, RepGen, Loc, Ret | 2D | https://aka.ms/biovil-t-code |
| Hu et al., 2025 | Cardiac-CLIP | ViT-B | PubMedBERT | CT | Cls, Ret, Dx | 3D | https://github.com/Sliver-g/Cardiac-CLIP |
| Tiu et al., 2022 | CheXzero | ViT-B | Transformer | X-ray | Cls | 2D | https://github.com/rajpurkarlab/CheXzero |
| Han et al., 2026 | CLEAR | ViT-B (DINOv2) | GPT-2 | X-ray | Cls | 2D | https://github.com/peterhan91/CLEAR |
| Ma et al., 2025 | CLISC | ViT-B / ResNet-50 | CLIP | MRI | Seg | 3D | — |
| Nie et al., 2025 | ConceptCLIP | SigLIP | PubMedBERT | Multi-modal | Cls, VQA, Ret, RepGen | 2D | https://github.com/JerrryNie/ConceptCLIP |
| Javed et al., 2024 | CPLIP | ViT-B | CLIP Text Encoder | Pathology | Cls, Seg, Ret | 2D | https://cplip.github.io |
| Yuan et al., 2025 | CRNS-Net | TransNeXt | CLIP Text Encoder | Pathology | Seg | 2D | https://github.com/rsy980/CRNS-Net |
| Hamamci et al., 2024 | CT-CLIP | ViT-B | CXR-BERT | CT | Cls, Det, Ret, RepGen, VQA | 3D | https://github.com/ibrahimethemhamamci/CT-CLIP |
| Lin et al., 2024 | CT-GLIP | nnU-Net | ClinicalBERT | Multi-modal | Cls, Seg, Det | 3D | — |
| Yang et al., 2025 | Decipher-MR | ViT-B | PubMedBERT | MRI | Cls, Seg, Ret, Loc | 3D | — |
| Luo et al., 2024 | DeViDe | ViT-B | Med-KEBERT | Multi-modal | Cls, Seg, Det | 2D | — |
| Wang et al., 2025 | ECAMP | ViT-B | BERT | Multi-modal | Cls, Seg, Det | 2D | https://github.com/ToniChopp/ECAMP |
| Christensen et al., 2024 | EchoCLIP | ConvNeXt-Base | Transformer (Decoder) | Echocardiography | Cls, Dx, Ret | 2D | https://github.com/echonet/echo_CLIP |
| Xu et al., 2023 | ELIXR | EfficientNet-L2 | PaLM 2-S | X-ray | Cls, Ret, VQA, RepGen | 2D | — |
| Maani et al., 2025 | FetalCLIP | ViT-L | Transformer | Ultrasound | Cls, Seg, Det, Dx | 2D | https://github.com/BioMedIA-MBZUAI/FetalCLIP |
| Silva-Rodríguez et al., 2025 | FLAIR | ResNet-50 | BioClinicalBERT | Fundus | Cls | 2D | https://github.com/jusiro/FLAIR |
| Killeen et al., 2025 | FluoroSAM | Swin (SAM) | CLIP Text Encoder | X-ray | Seg | 2D | https://github.com/arcadelab/fluorosam |
| Mirhosseini et al., 2026 | FT-PLIP | ViT-B | CLIP Text Encoder | Pathology | Cls | 2D | https://github.com/erfunm/ipath-ipclip |
| Deng et al., 2026 | GK-MVLP | ViT-B | SciBERT | X-ray | Cls, Loc, RepGen, VQA | 2D | — |
| Huang et al., 2021 | GLoRIA | ResNet-50 | BioClinicalBERT | X-ray | Cls, Seg, Ret | 2D | https://github.com/marshuang80/gloria |
| Liu et al., 2025 | IMITATE | ResNet-50 | BioClinicalBERT | Multi-modal | Cls, Seg, Det, Ret | 2D | https://github.com/cheliu-computation/IMITATE-TMI2024 |
| Dawidowicz et al., 2023 | LIMITR | ResNet-50 | BioClinicalBERT | X-ray | Ret, Loc, Cls | 2D | https://github.com/gefend/LIMITR |
| Liu et al., 2023 | M-FLAG | ResNet-50 | CXR-BERT | X-ray | Cls, Seg, Det | 2D | https://github.com/cheliu-computation/M-FLAG-MICCAI2023 |
| Ghosh et al., 2024 | Mammo-CLIP | EfficientNet | BioClinicalBERT | Mammography | Cls, Loc | 2D | https://github.com/batmanlab/Mammo-CLIP |
| Wan et al., 2023 | Med-UniC | ResNet-50 / ViT-B / ViT-L | CXR-BERT | X-ray | Cls, Seg, Det | 2D | https://github.com/SUSTechBruce/Med-UniC |
| Gao et al., 2024 | MEDBind | Swin-T | BioBERT | Multi-modal | Cls, Ret | 2D | — |
| Wang et al., 2022 | MedCLIP | Swin-T | BioClinicalBERT | X-ray | Cls, Ret | 2D | https://github.com/RyanWangZf/MedCLIP |
| Koleilat et al., 2024 | MedCLIP-SAM | ViT-B (SAM) + Mask Decoder | PubMedBERT | Multi-modal | Seg | 2D | https://github.com/HealthX-Lab/MedCLIP-SAM |
| Koleilat et al., 2025 | MedCLIP-SAMv2 | ViT-B (SAM) | PubMedBERT | Multi-modal | Seg | 2D | https://github.com/HealthX-Lab/MedCLIP-SAMv2 |
| Codella et al., 2024 | MedImageInsight | DaViT | Transformer | Multi-modal | Cls, Ret, RepGen | 2D/3D | — |
| Wu et al., 2023 | MedKLIP | ResNet-50 | ClinicalBERT | Multi-modal | Cls, Seg, Dx, Loc | 2D | https://github.com/chaoyi-wu/MedKLIP |
| Blankemeier et al., 2026 | Merlin | ResNet-152 | Clinical Longformer | CT | Cls, Seg, RepGen, Ret, Pred | 3D | https://github.com/StanfordMIMI/Merlin |
| Lu et al., 2023 | MI-Zero | CTransPath | HistPathGPT | Multi-modal | Cls, Ret | 2D | https://github.com/mahmoodlab/MI-Zero |
| Li et al., 2024 | MLIP | ResNet-50 / ViT-B | BioClinicalBERT | X-ray | Cls, Seg, Det | 2D | https://github.com/gentlefress/MLIP |
| Wang et al., 2025 | MMKD-CLIP | MetaCLIP ViT-B/16 | BioMed-BERT | Multi-modal | Cls, VQA, Ret | 2D | https://github.com/wangshansong1/MMKD-CLIP |
| Kim et al., 2024 | MONET | ViT-L | CLIP Text Encoder | Multi-modal | Cls | 2D | https://github.com/suinleelab/MONET |
| Huang et al., 2025 | NeoCLIP | ResNet-50 | BERT | X-ray | Cls | 2D | — |
| Li et al., 2025 | OpenVocabCT | STU-Net-Large | BIOLORD | CT | Seg | 3D | https://github.com/ricklisz/OpenVocabCT |
| Huang et al., 2023 | PLIP | ViT-B/32 | Transformer | Pathology | Cls, Ret | 2D | https://tinyurl.com/webplip |
| Lin et al., 2023 | PMC-CLIP | ResNet-50 | PubMedBERT | Multi-modal | Cls, Ret, VQA | 2D | — |
| Cheng et al., 2023 | PRIOR | ResNet-50 | ClinicalBERT | X-ray | Cls, Seg, Det, Ret | 2D | https://github.com/QtacierP/PRIOR |
| Chen et al., 2023 | PTUnifier | CLIP ViT-B | RoBERTa | Multi-modal | Cls, Ret, RepGen, VQA | 2D | https://github.com/zhjohnchan/PTUnifier |
| Lu et al., 2025 | RadCLIP | ViT-L | CLIP Text Encoder | Multi-modal | Cls, Ret | 2D/3D | https://github.com/luzhixiu/RadCLIP |
| Du et al., 2024 | RET-CLIP | ViT-B | RoBERTa | Fundus | Cls | 2D | https://github.com/sStonemason/RET-CLIP |
| Zhao et al., 2023 | SAT-Pro-Ft | 3D U-Net | MedCPT | Multi-modal | Seg | 3D | https://github.com/zhaoziheng/SAT |
| Yamamoto et al., 2026 | TotalFM | 3D ViT | Q-Former + RadLLaMA-7B | CT | Cls, RepGen | 3D | — |
| Dai et al., 2024 | UniChest | ResNet-50 | PubMedBERT | X-ray | Cls, Loc | 2D | https://github.com/Elfenreigen/UniChest |
| Khattak et al., 2024 | UniMed-CLIP | MetaCLIP | BioMed-BERT | Multi-modal | Cls | 2D | https://github.com/mbzuai-oryx/UniMed-CLIP |
| Liu et al., 2023 | Universal Model | U-Net | CLIP Text Encoder | CT | Seg, Det | 3D | https://github.com/ljwztc/CLIP-Driven-Universal-Model |
| Yu et al., 2024 | UrFound | ViT-B | BERT-Base | Multi-modal | Cls | 2D | https://github.com/yukkai/UrFound |
| Liang et al., 2026 | VISTA-PATH | PLIP | PLIP | Pathology | Seg | 2D | https://github.com/zhihuanglab/VISTA-PATH |
| Wan et al., 2025 | VOILA | Custom CNN | CLIP Text Encoder | Multi-modal | Seg | 3D | https://github.com/ZishuoWan/VOILA |
| ***Generative VLFMs*** | | | | | | | |
| Li et al., 2025 | RadDiagSeg-M | LISA | PaliGemma 2 | Multi-modal | Cls, Seg, VQA, Dx, Pred | 2D | https://github.com/RadDiagSeg |
| Zhao et al., 2025 | BiomedParse | FocalNet | PubMedBERT | Multi-modal | Seg, Det | 2D/3D | https://aka.ms/biomedparse-release |
| Zhao et al., 2025 | BiomedParse-V | FocalNet | UniCL | Multi-modal | Seg, Det | 3D | https://github.com/microsoft/BiomedParse |
| Liu et al., 2025 | NasVLM | ViT-B | BioClinicalBERT | Endoscopy | Cls, Seg | 2D | — |
| Molino et al., 2026 | XGeM | Stable Diffusion | Transformer | X-ray | Gen, Synth | 2D | — |
| Xiao et al., 2025 | MedDAM | Various VLMs | LLaMA 3.2 / Qwen2.5 | Multi-modal | Cap | 2D | — |
| Berger et al., 2025 | Context-Aware VLF | ResNet-50 | BioClinicalBERT | Fundus | Dx | 2D | — |
| Niu et al., 2025 | M3FM-Large | CTViT | RoBERTa | Multi-modal | Dx | 3D | https://github.com/niuchuangnn/M3FM |
| Zhang et al., 2025 | EditGRPO | ViT | Qwen2.5-VL-3B | X-ray | RepGen | 2D | — |
| Srivastav et al., 2024 | MAIRA-RRG24 | RAD-DINO | Vicuna-7B v1.5 | X-ray | RepGen | 2D | — |
| Chen et al., 2024 | CheXagent | ViT | Mistral-7B | X-ray | Cls, VQA, RepGen, Sum | 2D | https://github.com/stanford-aimi/chexagent |
| Mao et al., 2025 | MedSegFactory | Stable Diffusion | — | Multi-modal | Synth, Seg | 2D | https://github.com/jwmao1/MedSegFactory |
| Wang et al., 2025 | UniSyn | NeuroCLIP | NeuroCLIP | Multi-modal | Synth | 3D | — |
| Wang et al., 2025 | MINIM | Stable Diffusion | CLIP Text Encoder | Multi-modal | Synth | 2D | https://github.com/WithStomach/MINIM |
| Wang et al., 2025 | TUMSyn | CLIP Image Encoder | CLIP Text Encoder | MRI | Dx | 3D | https://github.com/Wangyulin-user/TUMSyn |
| Fan et al., 2025 | ChestX-Reasoner | ViT | Qwen2 | X-ray | Cls, Det | 2D | — |
| Sun et al., 2025 | CPath-Omni | CPath-CLIP | Qwen2.5-14B | Pathology | Cls, VQA, Cap | 2D | — |
| Lee et al., 2025 | CXR-LLaVA | ViT-L/16 | LLaMA 2 | X-ray | Cls, RepGen | 2D | https://github.com/ECOFRI/CXR_LLaVA |
| Lin et al., 2025 | HealthGPT | CLIP ViT-L | Phi-4 | Multi-modal | VQA, Gen, RepGen | 2D | https://github.com/DCDmllm/HealthGPT |
| Wang et al., 2025 | HiCur-NPC | ConvNeXt + ViT | LLaMA 3 Instruct | Multi-modal | Dx, Seg, Pred, VQA, RepGen | 2D | https://github.com/displaywz/HiCur-NPC-preview |
| Xu et al., 2025 | Lingshu | Qwen2.5-VL | Qwen2.5-VL | Multi-modal | Cls, RepGen, VQA, Cap | 2D | https://alibaba-damo-academy.github.io/lingshu/ |
| Li et al., 2023 | LLaVA-Med | CLIP ViT-L | Vicuna | Multi-modal | VQA | 2D | https://aka.ms/llava-med |
| Guo et al., 2024 | LLaVA-Ultra | CLIP ViT-L + ViT-B (SAM) | LLaMA | Ultrasound | VQA | 2D | — |
| Shi et al., 2025 | Med-2E3 | M3D-CLIP + SigLIP | Phi-3 Mini | CT | RepGen, VQA, Seg | 3D | https://github.com/MSIIP/Med-2E3 |
| Moor et al., 2023 | Med-Flamingo | CLIP ViT-L/14 | LLaMA-7B | Multi-modal | VQA | 2D | https://github.com/snap-stanford/med-flamingo |
| Xin et al., 2025 | Med3DVLM | DCFormer-S | ClinicalBERT | CT | VQA, RepGen, Ret | 3D | https://github.com/mirthAI/Med3DVLM |
| Fallahpour et al., 2025 | MedRAX | GPT-4o Vision | GPT-4o | X-ray | Cls, Det, Loc, Dx | 2D | https://github.com/bowang-lab/MedRAX |
| Zhou et al., 2024 | MedVersa | Swin-T / 3D U-Net | LLaMA 2 Chat | Multi-modal | Cap, Cls, Det, Seg, VQA | 2D/3D | https://tinyurl.com/fhnv3be4 |
| Xing et al., 2026 | MedVL-SAM2 | M3D-CLIP | InternVL-2.5-4B | CT | Seg, VQA, RepGen | 3D | — |
| Chen et al., 2025 | MIMO | CLIP ViT-H/14 | Vicuna | Multi-modal | Seg, VQA | 2D | https://github.com/pkusixspace/MIMO |
| He et al., 2025 | OmniMRI | Swin-T | Qwen2.5 | MRI | RepGen, Det, Dx, Seg | 2D/3D | — |
| Jiang et al., 2025 | OmniV-Med | SigLIP | Qwen2.5 | Multi-modal | RepGen, Ret, Loc | 2D/3D | — |
| Deng et al., 2024 | OphGLM | ConvNeXt | BERT + ChatGLM-6B | Fundus | Cls, Seg, VQA | 2D | https://github.com/ML-AILab/OphGLM |
| Lu et al., 2024 | PathChat | UNI | LLaMA 2 | Pathology | Cls, Cap, VQA, Dx | 2D | https://github.com/fedshyvana/pathology_mllm_training |
| Zhang et al., 2025 | Patho-AgenticRAG | ColQwen2 | Qwen3-4B | Pathology | VQA | 2D | https://github.com/Wenchuan-Zhang/Patho-AgenticRAG |
| Jiang et al., 2026 | PathReasoner-R1 | ViT | SlideChat-7B | Pathology | Cls, Ret, VQA | 2D | https://github.com/cyclexfy/PathReasoner-R1 |
| Wu et al., 2025 | RadFM | 3D ViT | MedLLaMA-13B | Multi-modal | Cls, VQA, RepGen | 2D/3D | https://github.com/chaoyi-wu/RadFM |
| Pellegrini et al., 2025 | RaDialog | BioViL-T | Vicuna | X-ray | RepGen, VQA, Sum, Cls | 2D | https://github.com/ChantalMP/RaDialog |
| Liu et al., 2025 | TeamPath-7B | Patho-R1 ViT | Qwen2.5-VL | Pathology | Dx, Cap, VQA | 2D | https://github.com/HelloWorldLTY/TeamPath |
| Zhu et al., 2024 | Uni-Med | EVA-CLIP | LLaMA 2 Chat | Multi-modal | VQA, RepGen, Loc, Cap, Cls | 2D | https://github.com/MSIIP/Uni-Med |
| Wu et al., 2025 | UniBiomed | Hiera (SAM 2) | InternVL2.5 | Multi-modal | Seg, Cls, RepGen, Dx, VQA | 2D | https://github.com/Luffy03/UniBiomed |
| Li et al., 2025 | VisionUnite | EVA02-CLIP | LLaMA-7B | Multi-modal | Cls, VQA | 2D | https://github.com/HUANGLIZI/VisionUnite |
| Luo et al., 2025 | VividMed | ViT | Vicuna v1.5 | Multi-modal | Seg, Det, RepGen, VQA | 2D/3D | https://github.com/function2-llx/MMMM |
| Thawakar et al., 2024 | XrayGPT | MedCLIP | Vicuna | X-ray | Sum, VQA | 2D | https://github.com/mbzuai-oryx/XrayGPT |